\documentclass[fleqn,twoside]{article}
\usepackage{espcrc1}

\newcommand{\ttbs}{\char'134}
\newcommand{\AmS}{{\protect\the\textfont2
  A\kern-.1667em\lower.5ex\hbox{M}\kern-.125emS}}

\title{Improvement/Extension of Modular Systems
 as Combinatorial Reengineering (Survey)}

\author{Mark Sh. Levin
\thanks{
 Mark Sh. Levin:~
 Inst. for Inform. Transmission Problems,
 Russian Academy of Sciences;
  http://www.mslevin.iitp.ru;
 email: mslevin@acm.org
  }
  }

\begin{document}

\maketitle

\begin{abstract}
 The paper describes
 development
 (improvement/extension) approaches
 for composite (modular) systems
 (as combinatorial reengineering).
%
%
 The following system improvement/extension actions are considered:
 (a) improvement of systems component(s)
 (e.g.,
 improvement of a system component,
 replacement of a system component);
 (b) improvement of system component
 interconnection (compatibility);
 (c) joint improvement improvement
 of system components(s) and their interconnection;
 (d) improvement of  system structure
 (replacement of system part(s),
 addition of a system part,
  deletion of a system part, modification of system structure).
  The study of system improvement approaches involve
 some crucial issues:
 (i) scales for evaluation of system components  and
 component compatibility
 (quantitative scale, ordinal scale, poset-like scale,
 scale based on interval multiset estimate),
 (ii) evaluation of integrated system quality,
 (iii) integration methods to obtain the
 integrated system quality.

 The system improvement/extension strategies
 can be examined as seleciton/combination of the improvement
 action(s) above and as modification of system structure.
 The strategies are based on combinatorial optimization problems
 (e.g.,
 multicriteria selection, knapsack problem, multiple choice problem,
 combinatorial synthesis based on morphological clique problem,
 assignment/reassignment problem, graph recoloring problem, spanning problems, hotlink assignment).
 Here, heuristics are used.
 Various system improvement/extension strategies are presented
  including illustrative numerical examples.

~~

{\it Keywords:}~
                   modular systems,
                   system design,
                   systems engineering,
                   engineering frameworks,
                   combinatorial optimization,
                   combinatorial reengineering,
                   networked systems,
                   heuristics

\vspace{1pc}
\end{abstract}


\tableofcontents

\newcounter{cms}
\setlength{\unitlength}{1mm}

\section{Introduction}


 In recent two decades, the significance of system reengineering
 (i.e., issues of systems redesign, rebuilt,
 improvement, upgrade, extension)
 has been increased
 (e.g.,
  \cite{lev96b,lev98,lev06,levsib10,levdan05,nolt99}).
 This paper addresses
  systems development schemes (i.e., improvement/upgrade, extension)
  for composite (modular) systems (as combinatorial reengineering).
 Generally,
 the system improvement process  is the following:

~~

 {\it Initial system(s)}
 ~ \(\Longrightarrow\) ~
 {\it Improvement process}
 ~ \(\Longrightarrow\) ~
 {\it Resultant (improved) system(s)}.

~~

%
  The systems approaches can be considered as follows:
 (a) improvement of systems components and/or their
 interconnections,
 (b) improvement/extension of a system structure,
 (c) ``space'' (e.g., geographical) extension of a system as
 designing an additional system part, and
 (d) combined system improvement/extension.

 A general approach to system development
 consists of the following (Fig. 1):

 {\it 1.} system improvement or modification
 (e.g., by components,
 by component interconnection,
 by system structure);

 {\it 2.} system extension as designing an additional system part
 and its integration with the basic system.

\begin{center}
\begin{picture}(93,22)
\put(15,0){\makebox(0,0)[bl]{Fig. 1. System and additional part
 \cite{levsib10}}}

\put(24,13){\oval(48,14)} \put(24,13){\oval(47,13.5)}

\put(12,15.5){\makebox(0,8)[bl]{Basic system:}}
\put(05,11.5){\makebox(0,8)[bl]{structure, components,}}
\put(03.5,07.5){\makebox(0,8)[bl]{component interconnection}}


\put(27,06){\line(1,0){66}} \put(27,20){\line(1,0){66}}

\put(93,06){\line(0,1){14}}


\put(50,15.5){\makebox(0,8)[bl]{Additional system part:}}
\put(50,11.5){\makebox(0,8)[bl]{structure, components, }}
\put(50,07.5){\makebox(0,8)[bl]{component interconnection}}

\end{picture}
\end{center}

 Fig. 2 depicts a hierarchy of general system improvement/development activities:
 (i) improvement of basic system,
 (ii) extension as designing of an additional system part,
 and
 (iii) coordinated improvement of basic system and
 designing an additional system part.


 The set of basic system improvement/extension actions is the following:
 (a) improvement of systems component(s)
 (e.g.,
 improvement of a system component,
 replacement of a system component);
 (b) improvement of system component interconnection (compatibility),
 (c) joint improvement improvement
 of system components(s) and their interconnection;
 (d) improvement of  system structure
 (replacement of system part(s),
 addition of a system part,
  deletion of a system part, modification of system structure).
%


 The system improvement/extension strategies
 are based on seleciton/combination of the improvement
 action(s) above (including modification of system structure).
 The strategies consist of combinatorial optimization problems
 (e.g.,
 multicriteria selection, knapsack problem, multiple choice problem,
 combinatorial synthesis based on morphological clique problem,
 assignment/reassignment problem, graph recoloring problem,
 spanning problems, hotlink assignment)
 (e.g., \cite{bar08,bilo08,gar79,kee76,kra01,lev09,roy96}).
 Here, various algorithms (including heuristics) are used.
%


 Note, system improvement/extension approaches involve
 some system evaluation crucial issues:
 (i) scales for evaluation of system components  and
 component compatibility
 (quantitative scale, ordinal scale, poset-like scale,
 scale based on interval multiset estimate)
 (ii) evaluation of integrated system quality
 (i.e., scale/space of system total quality),
 (iii) integration methods to obtain the integrated system quality.

  This paper is research survey.
  Various system improvement/extension strategies are presented.
 Special attention is targeted to networked systems.
  Numerical examples
 illustrate the described approaches.

\begin{center}
\begin{picture}(98,76)
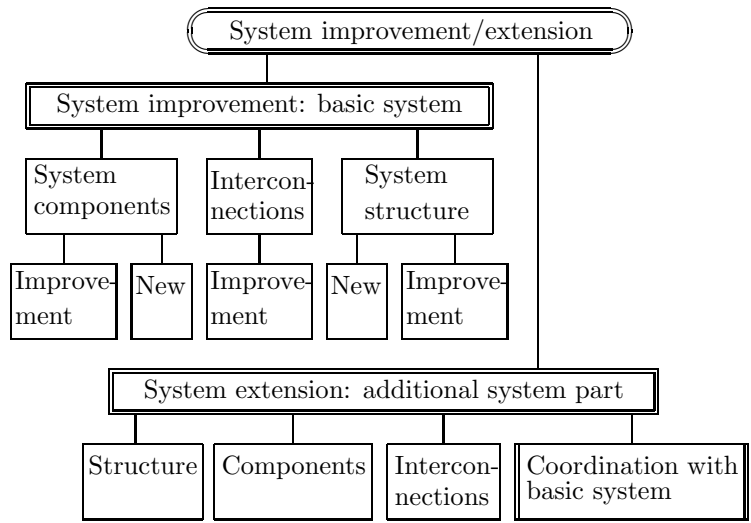
\put(07,0){\makebox(0,0)[bl]{Fig. 2. Hierarchy of system
 development  actions \cite{levsib10}}}

\put(53,71){\oval(59,06)} \put(53,71){\oval(58,05)}

\put(29,69){\makebox(0,0)[bl]{System improvement/extension}}

\put(34,64){\line(0,1){4}}

\put(70,26){\line(0,1){42}}


\put(2,58){\line(1,0){62}} \put(2,64){\line(1,0){62}}
\put(2,58){\line(0,1){6}} \put(64,58){\line(0,1){6}}

\put(2.5,58.5){\line(1,0){61}} \put(2.5,63.5){\line(1,0){61}}
\put(2.5,58.5){\line(0,1){5}} \put(63.5,58.5){\line(0,1){5}}

\put(06,59.5){\makebox(0,8)[bl]{System improvement: basic system}}


\put(12,54){\line(0,1){4}}

\put(02,44){\line(1,0){20}} \put(02,54){\line(1,0){20}}
\put(02,44){\line(0,1){10}} \put(22,44){\line(0,1){10}}

\put(03,50){\makebox(0,0)[bl]{System}}
\put(03,46){\makebox(0,0)[bl]{components}}


\put(33,54){\line(0,1){4}}

\put(26,44){\line(1,0){14}} \put(26,54){\line(1,0){14}}
\put(26,44){\line(0,1){10}} \put(40,44){\line(0,1){10}}

\put(26.5,50){\makebox(0,8)[bl]{Intercon-}}
\put(26.5,46){\makebox(0,8)[bl]{nections}}


\put(54,54){\line(0,1){4}}

\put(44,44){\line(1,0){20}} \put(44,54){\line(1,0){20}}
\put(44,44){\line(0,1){10}} \put(64,44){\line(0,1){10}}

\put(47,50){\makebox(0,0)[bl]{System}}
\put(47,46){\makebox(0,0)[bl]{structure}}

\put(07,40){\line(0,1){4}}

\put(00,30){\line(1,0){14}} \put(00,40){\line(1,0){14}}
\put(00,30){\line(0,1){10}} \put(14,30){\line(0,1){10}}

\put(0.5,36){\makebox(0,0)[bl]{Improve-}}
\put(0.5,32){\makebox(0,0)[bl]{ment}}


\put(20,40){\line(0,1){4}}

\put(16,30){\line(1,0){08}} \put(16,40){\line(1,0){08}}
\put(16,30){\line(0,1){10}} \put(24,30){\line(0,1){10}}

\put(16.5,36){\makebox(0,0)[bl]{New}}


\put(33,40){\line(0,1){4}}

\put(26,30){\line(1,0){14}} \put(26,40){\line(1,0){14}}
\put(26,30){\line(0,1){10}} \put(40,30){\line(0,1){10}}

\put(26.5,36){\makebox(0,8)[bl]{Improve-}}
\put(26.5,32){\makebox(0,8)[bl]{ment}}


\put(46,40){\line(0,1){4}}

\put(42,30){\line(1,0){08}} \put(42,40){\line(1,0){08}}
\put(42,30){\line(0,1){10}} \put(50,30){\line(0,1){10}}

\put(42.5,36){\makebox(0,0)[bl]{New}}


\put(59,40){\line(0,1){4}}

\put(52,30){\line(1,0){14}} \put(52,40){\line(1,0){14}}
\put(52,30){\line(0,1){10}} \put(66,30){\line(0,1){10}}

\put(52.5,36){\makebox(0,0)[bl]{Improve-}}
\put(52.5,32){\makebox(0,0)[bl]{ment}}


\put(13,20){\line(1,0){73}} \put(13,26){\line(1,0){73}}
\put(13,20){\line(0,1){6}} \put(86,20){\line(0,1){6}}

\put(13.5,20.5){\line(1,0){72}} \put(13.5,25.5){\line(1,0){72}}
\put(13.5,20.5){\line(0,1){5}} \put(85.5,20.5){\line(0,1){5}}

\put(17.6,21.5){\makebox(0,0)[bl]{System extension: additional
system part}}


\put(82.5,16){\line(0,1){4}}

\put(67,06){\line(1,0){31}} \put(67,16){\line(1,0){31}}
\put(67,06){\line(0,1){10}} \put(98,6){\line(0,1){10}}
\put(67.5,06){\line(0,1){10}} \put(97.5,6){\line(0,1){10}}

\put(68.5,12){\makebox(0,0)[bl]{Coordination with}}
\put(68.5,08){\makebox(0,0)[bl]{basic system }}


\put(16.5,16){\line(0,1){04}}

\put(09.5,06){\line(1,0){16}} \put(09.5,16){\line(1,0){16}}
\put(09.5,06){\line(0,1){10}} \put(25.5,6){\line(0,1){10}}

\put(10.3,12){\makebox(0,0)[bl]{Structure}}


\put(37.5,16){\line(0,1){04}}

\put(27,06){\line(1,0){21}} \put(27,16){\line(1,0){21}}
\put(27,06){\line(0,1){10}} \put(48,6){\line(0,1){10}}

\put(28,11.4){\makebox(0,0)[bl]{Components}}


\put(57.5,16){\line(0,1){04}}

\put(50,06){\line(1,0){15}} \put(50,16){\line(1,0){15}}
\put(50,06){\line(0,1){10}} \put(65,6){\line(0,1){10}}

\put(51,12){\makebox(0,0)[bl]{Intercon-}}
\put(51,08){\makebox(0,0)[bl]{nections}}

\end{picture}
\end{center}

\section{Four System Improvement Situations}

\subsection{Basic System Improvement Situation}

 The basic system improvement situation is depicted in Fig. 3
 (one initial system and one resultant improved system).

\begin{center}
\begin{picture}(71,42)
\put(001,00){\makebox(0,0)[bl]{Fig. 3. Basic system improvement
 \cite{lev98,lev06,levsib10}}}


\put(00,24){\line(1,0){18}}

\put(00,24){\line(0,1){08}} \put(18,24){\line(0,1){08}}

\put(00,32){\line(1,1){09}} \put(18,32){\line(-1,1){09}}

\put(3.5,30){\makebox(0,0)[bl]{Initial}}
\put(3,26){\makebox(0,0)[bl]{system}}

\put(18,31){\vector(1,0){6}}


\put(24,24){\line(1,0){24}} \put(24,38){\line(1,0){24}}
\put(24,24){\line(0,1){14}} \put(48,24){\line(0,1){14}}

\put(25,34){\makebox(0,0)[bl]{Improvement}}
\put(25,30){\makebox(0,0)[bl]{(combinatorial }}
\put(25,26){\makebox(0,0)[bl]{(synthesis)}}

\put(48,31){\vector(1,0){5.5}}


\put(24,06){\line(1,0){24}} \put(24,20){\line(1,0){24}}
\put(24,06){\line(0,1){14}} \put(48,06){\line(0,1){14}}

\put(26,16){\makebox(0,0)[bl]{Improvement}}
\put(27,12){\makebox(0,0)[bl]{operations}}
\put(30,08){\makebox(0,0)[bl]{(items)}}

\put(36,20){\vector(0,1){4}}


\put(00,06){\line(1,0){18}} \put(00,20){\line(1,0){18}}
\put(00,06){\line(0,1){14}} \put(18,06){\line(0,1){14}}

\put(0.5,16){\makebox(0,0)[bl]{Detection}}
\put(01,11.5){\makebox(0,0)[bl]{of system}}
\put(0.5,08){\makebox(0,0)[bl]{bottlenecks}}

\put(09,24){\vector(0,-1){4}}

\put(18,13){\vector(1,0){6}}


\put(54,24.5){\line(1,0){16}}

\put(54,24.5){\line(0,1){11}} \put(70,24.5){\line(0,1){11}}
\put(54,35.5){\line(2,1){08}} \put(70,35.5){\line(-2,1){08}}

\put(54.7,32.5){\makebox(0,0)[bl]{Resultant}}
\put(55,29){\makebox(0,0)[bl]{improved}}
\put(57,26){\makebox(0,0)[bl]{system}}

\put(53.5,24){\line(1,0){17}}

\put(53.5,24){\line(0,1){12}} \put(70.5,24){\line(0,1){12}}
\put(53.5,36){\line(2,1){08.5}} \put(70.5,36){\line(-2,1){08.5}}

\end{picture}
\end{center}

 The following special cases can be considered for the above-mentioned basic system improvement situation
 (Fig. 3):

 {\it Special case 1.} Improvement of systems components:
 (1.1) improvement of system elements,
 (1.2) improvement of system elements interconnection (i.e., compatibility),
 and
 (1.3) joint improvement of system elements and their
 compatibility.

 {\it Special case 2.} Improvement as modification of system structure:
 (2.1) extension of  system structure (i.e., addition of a system structure part),
 (2.2) modification system structure as deletion a system structure part,
 (2.3) modification of system structure (i.e., transformation:
 addition/deletion of elements,
 addition/deletion of element interconnections),
 and
 (2.4) joint case (i.e., deletion of a system part structure
 and addition of a system structure part, modification of system structure).

 {\it Special case 3.}
 Extension of a system as
 designing an additional system part
 (i.e., additional system structure part and additional system elements).

 {\it Special case 4.} Combination of the above-mentioned cases.

\subsubsection{System Improvement/Modification}

 In general, system improvement/modification processes
 are based on the following three action kinds
 (e.g., \cite{lev98,lev06,levsib10}):
 (i) improvement (modification, upgrade, addition) of a system
 component,
 (ii) improvement of system components compatibility,
 (iii) change of a system structure.
 Some applied examples of system improvements for modular systems
 were  presented in the following sources:
 (a) building  (e.g., \cite{lev06,levdan05}),
 (b) information system  (e.g., \cite{lev98}),
 (c) human-computer systems (e.g., \cite{lev06,lev02}),
 (d) communication protocols and standards
 (e.g., \cite{lev09had,levand12}),
 (e) management system for smart homes
 (e.g., \cite{lev13home,levand11}),
  and
 (f) communication networks (e.g., \cite{levsib10,levsaf10a}).

 Evidently, detection of {\it system bottlenecks}
 may be considered as a preliminary phase.
 Fig. 4 depicts an illustrative example
 for a component-based improvement process:
 ~\(S^{a} \Rightarrow S^{b} \)
 ~(\(X_{1} \Rightarrow  X_{2}\),~
 \(Z_{1} \Rightarrow  Z_{3}\)).

\begin{center}
\begin{picture}(40,39)
\put(16.5,00){\makebox(0,0)[bl]{Fig. 4. System improvements (by
components)}}

\put(07,34){\makebox(0,8)[bl]{System}}
\put(03.8,30){\makebox(0,8)[bl]{morphology}}

\put(3,24){\line(1,0){20}} \put(13,24){\line(0,1){04}}
\put(13,28){\circle*{2.8}}

\put(3,24){\line(0,-1){04}} \put(13,24){\line(0,-1){04}}
\put(23,24){\line(0,-1){04}}

\put(0,21){\makebox(0,8)[bl]{\(X\)}}
\put(09.5,21){\makebox(0,8)[bl]{\(Y\)}}
\put(19.5,21){\makebox(0,8)[bl]{\(Z\)}}

\put(3,19){\circle*{2}} \put(13,19){\circle*{2}}
\put(23,19){\circle*{2}}


\put(04,13){\makebox(0,8)[bl]{\(X_{1}\)}}
\put(04,09){\makebox(0,8)[bl]{\(X_{2}\)}}
\put(04,05){\makebox(0,8)[bl]{\(X_{3}\)}}

\put(0,14){\line(1,0){02}} \put(2,14){\circle*{1.5}}
\put(0,09){\line(1,0){02}} \put(2,09){\circle*{1.5}}
\put(0,05){\line(1,0){02}} \put(2,05){\circle*{1.5}}

\put(0,19){\line(0,-1){14}} \put(0,19){\line(1,0){02}}

\put(14,13){\makebox(0,8)[bl]{\(Y_{1}\)}}
\put(14,09){\makebox(0,8)[bl]{\(Y_{2}\)}}
\put(14,05){\makebox(0,8)[bl]{\(Y_{3}\)}}

\put(10,14){\line(1,0){02}} \put(12,14){\circle*{1.5}}
\put(10,09){\line(1,0){02}} \put(12,09){\circle*{1.5}}
\put(10,05){\line(1,0){02}} \put(12,05){\circle*{1.5}}

\put(10,19){\line(0,-1){14}} \put(10,19){\line(1,0){02}}


\put(24,13){\makebox(0,8)[bl]{\(Z_{1}\)}}
\put(24,09){\makebox(0,8)[bl]{\(Z_{1}\)}}
\put(24,05){\makebox(0,8)[bl]{\(Z_{1}\)}}

\put(20,14){\line(1,0){02}} \put(22,14){\circle*{1.5}}
\put(20,09){\line(1,0){02}} \put(22,09){\circle*{1.5}}
\put(20,05){\line(1,0){02}} \put(22,05){\circle*{1.5}}

\put(20,19){\line(0,-1){14}} \put(20,19){\line(1,0){02}}


\end{picture}
%
\begin{picture}(70,39)

\put(03,34){\makebox(0,8)[bl]{Initial system}}

\put(00,30){\makebox(0,8)[bl]{\(S^{a} = X_{1} \star Y_{1} \star
 Z_{1}\)}}

\put(3,24){\line(1,0){20}} \put(13,24){\line(0,1){04}}
\put(13,28){\circle*{2.8}}

\put(3,24){\line(0,-1){04}} \put(13,24){\line(0,-1){04}}
\put(23,24){\line(0,-1){04}}

\put(0,21){\makebox(0,8)[bl]{\(X\)}}
\put(09.5,21){\makebox(0,8)[bl]{\(Y\)}}
\put(19.5,21){\makebox(0,8)[bl]{\(Z\)}}

\put(3,19){\circle*{2}} \put(13,19){\circle*{2}}
\put(23,19){\circle*{2}}


\put(04,13){\makebox(0,8)[bl]{\(X_{1}\)}}

\put(0,14){\line(1,0){02}} \put(2,14){\circle*{1.5}}

\put(0,19){\line(0,-1){5}} \put(0,19){\line(1,0){02}}

\put(14,13){\makebox(0,8)[bl]{\(Y_{1}\)}}

\put(10,14){\line(1,0){02}} \put(12,14){\circle*{1.5}}

\put(10,19){\line(0,-1){5}} \put(10,19){\line(1,0){02}}


\put(24,13){\makebox(0,8)[bl]{\(Z_{1}\)}}

\put(20,14){\line(1,0){02}}

\put(22,14){\circle*{1.5}}

\put(20,19){\line(0,-1){5}} \put(20,19){\line(1,0){02}}


\put(30,19){\makebox(0,8)[bl]{\(\Longrightarrow \)}}


\put(40.4,34){\makebox(0,8)[bl]{Improved system}}

\put(40,30){\makebox(0,8)[bl]{\(S^{b} = X_{2} \star Y_{1} \star
 Z_{3}\)}}

\put(43,24){\line(1,0){20}} \put(53,24){\line(0,1){04}}
\put(53,28){\circle*{2.8}}

\put(43,24){\line(0,-1){04}} \put(53,24){\line(0,-1){04}}
\put(63,24){\line(0,-1){04}}

\put(40,21){\makebox(0,8)[bl]{\(X\)}}
\put(49.5,21){\makebox(0,8)[bl]{\(Y\)}}
\put(59.5,21){\makebox(0,8)[bl]{\(Z\)}}

\put(43,19){\circle*{2}} \put(53,19){\circle*{2}}
\put(63,19){\circle*{2}}


\put(44,09){\makebox(0,8)[bl]{\(X_{2}\)}}

\put(40,10){\line(1,0){02}}  \put(42,10){\circle{1.5}}

\put(42,10){\circle*{0.7}}

\put(40,19){\line(0,-1){9}} \put(40,19){\line(1,0){02}}


\put(54,13){\makebox(0,8)[bl]{\(Y_{1}\)}}

\put(50,14){\line(1,0){02}} \put(52,14){\circle*{1.5}}

\put(50,19){\line(0,-1){5}} \put(50,19){\line(1,0){02}}

\put(64,5){\makebox(0,8)[bl]{\(Z_{3}\)}}

\put(60,06){\line(1,0){02}}

\put(62,06){\circle{1.5}} \put(62,06){\circle*{0.7}}

\put(60,19){\line(0,-1){13}} \put(60,19){\line(1,0){02}}

\end{picture}
\end{center}

\begin{center}
\begin{picture}(117,150)

\put(00,146){\makebox(0,0)[bl]{Table 1. Basic system improvement
 situation, underlying problems/models}}

\put(00,0){\line(1,0){117}} \put(00,134){\line(1,0){117}}
\put(00,144){\line(1,0){117}}

\put(00,0){\line(0,1){144}} \put(40,0){\line(0,1){144}}
\put(62,0){\line(0,1){144}} \put(101,0){\line(0,1){144}}
\put(117,0){\line(0,1){144}}


\put(01,140){\makebox(0,0)[bl]{Type of system }}
\put(01,136){\makebox(0,0)[bl]{improvement}}

\put(41,140){\makebox(0,0)[bl]{Approaches}}

\put(63,139.5){\makebox(0,0)[bl]{Underlying problems/}}
\put(63,136){\makebox(0,0)[bl]{models}}

\put(102,140){\makebox(0,0)[bl]{Some }}
\put(102,136){\makebox(0,0)[bl]{sources}}


\put(0.5,129){\makebox(0,0)[bl]{Case 1  (improvement of}}

\put(2.5,125){\makebox(0,0)[bl]{systems
 components):}}


\put(1.5,121){\makebox(0,0)[bl]{(1.1) improvement of }}
\put(4.5,117){\makebox(0,0)[bl]{system elements}}

\put(41,121){\makebox(0,0)[bl]{Replacement,}}
\put(41,117){\makebox(0,0)[bl]{redesign}}

\put(63,121){\makebox(0,0)[bl]{Knapsack-like problems,}}
\put(63,117){\makebox(0,0)[bl]{HMMD,}}
\put(63,113){\makebox(0,0)[bl]{assigment/allocation}}
\put(63,109){\makebox(0,0)[bl]{graph recoloring,}}

\put(102,121){\makebox(0,0)[bl]{\cite{lev96b,lev98,lev02}}}
\put(102,117){\makebox(0,0)[bl]{\cite{lev06,lev09,lev12morph}}}
\put(102,113){\makebox(0,0)[bl]{\cite{lev13home,levsaf10a}}}


\put(1.5,105){\makebox(0,0)[bl]{(1.2) improvement of }}
\put(4.5,101){\makebox(0,0)[bl]{element compatibility}}

\put(41,105){\makebox(0,0)[bl]{Replacement, }}
\put(41,101){\makebox(0,0)[bl]{redesign}}

\put(63,105){\makebox(0,0)[bl]{Knapsack-like problems,}}
\put(63,101){\makebox(0,0)[bl]{HMMD,}}
\put(63,97){\makebox(0,0)[bl]{assigment/allocation }}

\put(102,105){\makebox(0,0)[bl]{\cite{lev98,lev06}}}


\put(1.5,93){\makebox(0,0)[bl]{(1.3) joint improvement}}
\put(4.5,89){\makebox(0,0)[bl]{of system element(s) }}
\put(4.5,85){\makebox(0,0)[bl]{\& compatibility}}

\put(41,93){\makebox(0,0)[bl]{Combined }}
\put(41,89){\makebox(0,0)[bl]{approaches}}

\put(63,93){\makebox(0,0)[bl]{Composite frameworks,}}
\put(63,89){\makebox(0,0)[bl]{graph recoloring,}}
\put(63,85){\makebox(0,0)[bl]{HMMD}}

\put(102,93){\makebox(0,0)[bl]{\cite{lev98,lev06,lev09}}}
\put(102,89){\makebox(0,0)[bl]{\cite{lev12morph}}}


\put(0.5,80){\makebox(0,0)[bl]{Case 2 (modification of }}
\put(2.5,76){\makebox(0,0)[bl]{system structure):}}



\put(1.5,72){\makebox(0,0)[bl]{(2.1) addition of system}}
\put(4.5,68){\makebox(0,0)[bl]{structure part}}

\put(41,72){\makebox(0,0)[bl]{Addition}}

\put(63,72){\makebox(0,0)[bl]{Knapsack-like problems,}}
\put(63,68){\makebox(0,0)[bl]{HMMD,}}
\put(63,64){\makebox(0,0)[bl]{hotlink assignment}}

\put(102,72){\makebox(0,0)[bl]{\cite{lev12hier,levsaf10a}}}


\put(1.5,60){\makebox(0,0)[bl]{(2.2) deletion of system }}
\put(4.5,56){\makebox(0,0)[bl]{structure part}}

\put(41,60){\makebox(0,0)[bl]{Deletion}}

\put(63,60){\makebox(0,0)[bl]{Knapsack-like problems,}}
\put(63,56){\makebox(0,0)[bl]{HMMD}}

\put(102,60){\makebox(0,0)[bl]{\cite{lev12hier}}}


\put(1.5,52){\makebox(0,0)[bl]{(2.3) modification of }}
\put(4.5,48){\makebox(0,0)[bl]{system structure}}

\put(41,52){\makebox(0,0)[bl]{Addition,}}
\put(41,48){\makebox(0,0)[bl]{deletion,}}
\put(41,44){\makebox(0,0)[bl]{aggregation/}}
\put(41,40){\makebox(0,0)[bl]{integration,}}
\put(41,36){\makebox(0,0)[bl]{restructuring}}

\put(63,52){\makebox(0,0)[bl]{Spanning problems, }}
\put(63,48){\makebox(0,0)[bl]{knapsack-like problems, }}
\put(63,44){\makebox(0,0)[bl]{HMMD, }}
\put(63,40){\makebox(0,0)[bl]{recovering problems,}}
\put(63,36){\makebox(0,0)[bl]{augmentation problems,}}
\put(63,32){\makebox(0,0)[bl]{reoptimization}}

\put(102,52){\makebox(0,0)[bl]{\cite{lev81,lev98,lev11restr}}}
\put(102,48){\makebox(0,0)[bl]{\cite{lev12hier,levnur11,levzam11}}}


\put(1.5,28){\makebox(0,0)[bl]{(2.4) joint case}}

\put(41,28.5){\makebox(0,0)[bl]{Combined}}
\put(41,24){\makebox(0,0)[bl]{approaches}}

\put(63,28){\makebox(0,0)[bl]{Composite frameworks}}

\put(102,28){\makebox(0,0)[bl]{\cite{lev12hier}}}


\put(0.5,19){\makebox(0,0)[bl]{Case 3 (addition of}}
\put(2.5,15){\makebox(0,0)[bl]{system part)}}

\put(41,19){\makebox(0,0)[bl]{Design }}

\put(63,19){\makebox(0,0)[bl]{Design frameworks,}}
\put(63,15){\makebox(0,0)[bl]{Knapsack-like problems,}}
\put(63,11){\makebox(0,0)[bl]{HMMD}}

\put(102,19){\makebox(0,0)[bl]{\cite{levsib10,lev12hier}}}


\put(0.5,06){\makebox(0,0)[bl]{Case 4 (combination}}
\put(2.5,02){\makebox(0,0)[bl]{of previous  cases)}}

\put(41,06){\makebox(0,0)[bl]{Combined }}
\put(41,02){\makebox(0,0)[bl]{approaches}}

\put(63,06){\makebox(0,0)[bl]{Problem frameworks}}

\put(102,06){\makebox(0,0)[bl]{\cite{levsib10,lev12hier}}}

\end{picture}
\end{center}

\subsubsection{System Extension}

 Generally, the system extension problem consists in
 designing an additional system part.
 Here, three basic extension strategies may be considered:

 {\it I. Independent (separated) design.} Designing the additional system part.
 As a result, the new system will include two system parts (i.e.,
 initial and additional).

 {\it II. Generalized new design.}
 Designing a new ``generalized'' system, which involves
 the initial system part and additional system part (integrated design).

 {\it III. Separated design with coordination.}
 Designing the additional system part, modification of the initial
 system part and coordination of initial system part and the new
 additional system part.

\subsection{Basic System Improvement Situation and Combinatorial Problems}

 Table 1 contains a list of basic approaches and corresponding
 combinatorial optimization problems for the considered basic
 system improvement situation (Fig. 3).

\subsection{Additional System Improvement Situations}

 In general, it is reasonable
 to consider the following additional system improvement situations:

 (a) aggregation:
 to obtain an improved system as aggregation of several initial systems
 (Fig. 5),

 (b) designing a set of improved systems (Fig. 6),

 (c) joint aggregation and designing the set of improved systems
 (Fig. 7).

 In the case of multi-objective design approaches,
 designing the set of improved systems may be based on
 obtaining the Pareto-efficient solutions, which can be considered
 as the system set.

\begin{center}
\begin{picture}(75,44)

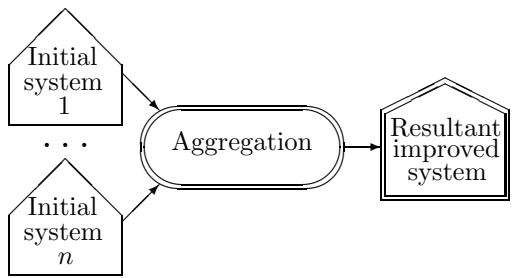
\put(00,00){\makebox(0,0)[bl]{Fig. 5. System improvement as
 aggregation}}


\put(00,26){\line(1,0){15}}

\put(00,26){\line(0,1){08}} \put(15,26){\line(0,1){08}}

\put(00,34){\line(1,1){07.5}} \put(15,34){\line(-1,1){07.5}}

\put(2.5,34){\makebox(0,0)[bl]{Initial}}
\put(2,30){\makebox(0,0)[bl]{system}}
\put(06.5,27.5){\makebox(0,0)[bl]{\(1\)}}

\put(15,33){\vector(1,-1){5}}

\put(04.5,23){\makebox(0,0)[bl]{{\bf .~.~.}}}

\put(00,06){\line(1,0){15}}

\put(00,06){\line(0,1){08}} \put(15,06){\line(0,1){08}}

\put(00,14){\line(1,1){07.5}} \put(15,14){\line(-1,1){07.5}}

\put(2.5,14){\makebox(0,0)[bl]{Initial}}
\put(2,10){\makebox(0,0)[bl]{system}}
\put(06.5,07.5){\makebox(0,0)[bl]{\(n\)}}

\put(15,13){\vector(1,1){5}}


\put(31,23){\oval(26,10)} \put(31,23){\oval(27,11)}
\put(21.6,22){\makebox(0,0)[bl]{Aggregation}}

\put(44.5,23){\vector(1,0){5}}


\put(50,16.5){\line(1,0){16}}

\put(50,16.5){\line(0,1){11}} \put(66,16.5){\line(0,1){11}}
\put(50,27.5){\line(2,1){08}} \put(66,27.5){\line(-2,1){08}}

\put(50.7,24.5){\makebox(0,0)[bl]{Resultant}}
\put(51,21){\makebox(0,0)[bl]{improved}}
\put(53,18){\makebox(0,0)[bl]{system}}

\put(49.5,16){\line(1,0){17}}

\put(49.5,16){\line(0,1){12}} \put(66.5,16){\line(0,1){12}}
\put(49.5,28){\line(2,1){08.5}} \put(66.5,28){\line(-2,1){08.5}}

\end{picture}
%
\begin{picture}(70,44)
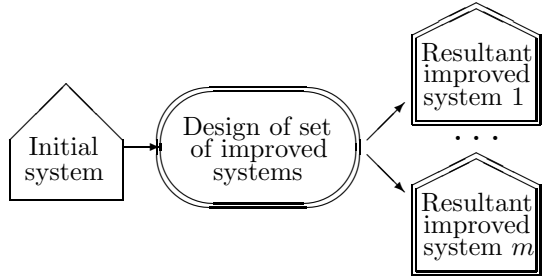
\put(01,00){\makebox(0,0)[bl]{Fig. 6. Improvement to
 design a system set}}


\put(00,16){\line(1,0){15}}

\put(00,16){\line(0,1){08}} \put(15,16){\line(0,1){08}}

\put(00,24){\line(1,1){07.5}} \put(15,24){\line(-1,1){07.5}}

\put(2.5,22){\makebox(0,0)[bl]{Initial}}
\put(2,18){\makebox(0,0)[bl]{system}}

\put(15,23){\vector(1,0){5}}



\put(33,23){\oval(26,15)} \put(33,23){\oval(27,16)}

\put(23,24){\makebox(0,0)[bl]{Design of set}}
\put(24,21){\makebox(0,0)[bl]{of improved}}
\put(27,18){\makebox(0,0)[bl]{systems}}

\put(47.5,22){\vector(1,-1){5}} \put(47.5,24){\vector(1,1){5}}


\put(54,26.5){\line(1,0){16}}

\put(54,26.5){\line(0,1){11}} \put(70,26.5){\line(0,1){11}}
\put(54,37.5){\line(2,1){08}} \put(70,37.5){\line(-2,1){08}}

\put(54.7,34.5){\makebox(0,0)[bl]{Resultant}}
\put(55,31){\makebox(0,0)[bl]{improved}}
\put(55,28){\makebox(0,0)[bl]{system \(1\)}}

\put(53.5,26){\line(1,0){17}}

\put(53.5,26){\line(0,1){12}} \put(70.5,26){\line(0,1){12}}
\put(53.5,38){\line(2,1){08.5}} \put(70.5,38){\line(-2,1){08.5}}

\put(59,24){\makebox(0,0)[bl]{{\bf .~.~.}}}

\put(54,6.5){\line(1,0){16}}

\put(54,6.5){\line(0,1){11}} \put(70,6.5){\line(0,1){11}}
\put(54,17.5){\line(2,1){08}} \put(70,17.5){\line(-2,1){08}}

\put(54.7,14.5){\makebox(0,0)[bl]{Resultant}}
\put(55,11){\makebox(0,0)[bl]{improved}}
\put(55,8){\makebox(0,0)[bl]{system \(m\)}}

\put(53.5,6){\line(1,0){17}}

\put(53.5,6){\line(0,1){12}} \put(70.5,6){\line(0,1){12}}
\put(53.5,18){\line(2,1){08.5}} \put(70.5,18){\line(-2,1){08.5}}

\end{picture}
\end{center}

\begin{center}
\begin{picture}(70,44)
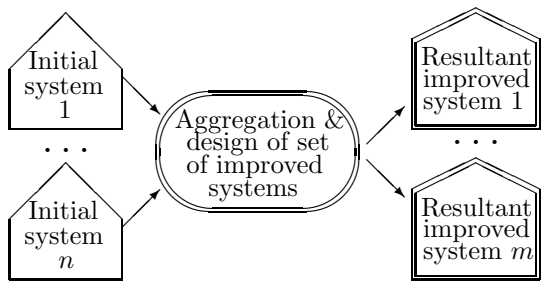
\put(02,00){\makebox(0,0)[bl]{Fig. 7. Aggregation \&
 design of system set}}


\put(00,26){\line(1,0){15}}

\put(00,26){\line(0,1){08}} \put(15,26){\line(0,1){08}}

\put(00,34){\line(1,1){07.5}} \put(15,34){\line(-1,1){07.5}}

\put(2.5,34){\makebox(0,0)[bl]{Initial}}
\put(2,30){\makebox(0,0)[bl]{system}}
\put(06.5,27.5){\makebox(0,0)[bl]{\(1\)}}

\put(15,33){\vector(1,-1){5}}

\put(04.5,23){\makebox(0,0)[bl]{{\bf .~.~.}}}

\put(00,06){\line(1,0){15}}

\put(00,06){\line(0,1){08}} \put(15,06){\line(0,1){08}}

\put(00,14){\line(1,1){07.5}} \put(15,14){\line(-1,1){07.5}}

\put(2.5,14){\makebox(0,0)[bl]{Initial}}
\put(2,10){\makebox(0,0)[bl]{system}}
\put(06.5,07.5){\makebox(0,0)[bl]{\(n\)}}

\put(15,13){\vector(1,1){5}}


\put(33,23){\oval(26,15)} \put(33,23){\oval(27,16)}

\put(22,25.5){\makebox(0,0)[bl]{Aggregation \&}}
\put(23.5,22.5){\makebox(0,0)[bl]{design of set}}
\put(23.5,19.5){\makebox(0,0)[bl]{of improved}}
\put(26.5,16.5){\makebox(0,0)[bl]{systems}}

\put(47.5,22){\vector(1,-1){5}} \put(47.5,24){\vector(1,1){5}}


\put(54,26.5){\line(1,0){16}}

\put(54,26.5){\line(0,1){11}} \put(70,26.5){\line(0,1){11}}
\put(54,37.5){\line(2,1){08}} \put(70,37.5){\line(-2,1){08}}

\put(54.7,34.5){\makebox(0,0)[bl]{Resultant}}
\put(55,31){\makebox(0,0)[bl]{improved}}
\put(55,28){\makebox(0,0)[bl]{system \(1\)}}

\put(53.5,26){\line(1,0){17}}

\put(53.5,26){\line(0,1){12}} \put(70.5,26){\line(0,1){12}}
\put(53.5,38){\line(2,1){08.5}} \put(70.5,38){\line(-2,1){08.5}}

\put(59,24){\makebox(0,0)[bl]{{\bf .~.~.}}}

\put(54,6.5){\line(1,0){16}}

\put(54,6.5){\line(0,1){11}} \put(70,6.5){\line(0,1){11}}
\put(54,17.5){\line(2,1){08}} \put(70,17.5){\line(-2,1){08}}

\put(54.7,14.5){\makebox(0,0)[bl]{Resultant}}
\put(55,11){\makebox(0,0)[bl]{improved}}
\put(55,8){\makebox(0,0)[bl]{system \(m\)}}

\put(53.5,6){\line(1,0){17}}

\put(53.5,6){\line(0,1){12}} \put(70.5,6){\line(0,1){12}}
\put(53.5,18){\line(2,1){08.5}} \put(70.5,18){\line(-2,1){08.5}}

\end{picture}
\end{center}

 A recent survey of aggregation approaches to  hierarchical modular systems
 is presented in \cite{lev11agg}.
  In the case of multi-objective design approaches,
 designing the set of improved systems may be based on
 obtaining the Pareto-efficient solutions,
 which can be considered as the required system set
 (e.g., \cite{lev98,lev06,lev09,lev11agg}).


\section{Approaches to System Evaluation and Improvement}

\subsection{Spaces of System Quality and Improvement}

  Generally, system improvement process
 for composite (modular) systems consists in
 selection of system improvement actions (operations) to obtain increasing
 a generalized system ``utility'' (quality, excellence)
  while taking into account a
 total constraint(s) for costs of the improvement actions.
 Thus, it is necessary to consider the following issues:

 (i) assessment of the  system ``utility''
 (a ``space'' as a scale, multidimensional space, a poset/lattice),

 (ii) cost of the improvement  actions,

 (iii) combinatorial improvement problem
 (e.g., multicriteria selection of the improvement actions,
 knapsack-like problem, combinatorial synthesis
 as selection and composition of the improvement actions).

 Evidently, special multi-stage system improvement procedures
 can be considered as well.

 System evaluation approaches include
 a method to integrate estimates for system components and
 estimates for system component interconnection (compatibility).
 Table 2 contains the basic approaches to evaluation of the total
 system ``utility'' (quality, excellence).

\begin{center}
\begin{picture}(102,79)

\put(17,75){\makebox(0,0)[bl]{Table 2. Evaluation of total system
quality}}

\put(00,0){\line(1,0){102}} \put(00,66){\line(1,0){102}}
\put(00,73){\line(1,0){102}}

\put(00,00){\line(0,1){73}} \put(81,00){\line(0,1){73}}
\put(102,00){\line(0,1){73}}


\put(01,68){\makebox(0,0)[bl]{Method }}

\put(82,68){\makebox(0,0)[bl]{Sources}}


\put(01,61){\makebox(0,0)[bl]{1.``Utility'' function for system
 quality}}

\put(04,57){\makebox(0,0)[bl]{(system components are assessed
 by}}

\put(04,53){\makebox(0,0)[bl]{quantitative/ordinal scales)}}


\put(82,61){\makebox(0,0)[bl]{\cite{fis70,kee76,ste86}
 }}


\put(01,48){\makebox(0,0)[bl]{2.Multicriteria description for
 system quality}}


\put(82,48){\makebox(0,0)[bl]{\cite{kee76,pareto71,roy96,ste86} }}


\put(01,43){\makebox(0,0)[bl]{3.Poset-like scale for system
 quality:}}

\put(82,43){\makebox(0,0)[bl]{\cite{lev98,lev06,lev12morph}}}


\put(02,39){\makebox(0,0)[bl]{3.1.Ordinal scales for system
 components}}

\put(05,35){\makebox(0,0)[bl]{and for components compatibility}}


\put(02,31){\makebox(0,0)[bl]{3.2.Ordinal scales for system
 components,}}

\put(05,27){\makebox(0,0)[bl]{poset-like scale for components
 compatibility}}


\put(01,22){\makebox(0,0)[bl]{4. Poset-like scale for system
 quality}}

\put(04,18.5){\makebox(0,0)[bl]{based on interval multiset
 estimates:}}

\put(82,22){\makebox(0,0)[bl]{\cite{lev12a}}}


\put(02,14){\makebox(0,0)[bl]{4.1.Multiset estimates for system
 components and}}

\put(05,10){\makebox(0,0)[bl]{ordinal estimates for components
 compatibility}}


\put(02,06){\makebox(0,0)[bl]{4.2.Multiset estimates for system
 components and}}

\put(05,02){\makebox(0,0)[bl]{multiset estimates for components
 compatibility}}


\end{picture}
\end{center}

 Illustrations for the above-mentioned methods  are the following:

 {\bf Method 1:}~ scale of system  ``utility'' (Fig. 8),
  initial system \(S^{a}\) is transformed into improved system
 \(S^{b}\) where system ``utility'' is designated as \(N(S^{a})\)
 (\(N(S^{b})\)).

 {\bf Method 2:}~ multicriteria case (two criteria), ideal point
 \(S^{I}\),
 four Pareto-efficient solutions (\(S^{p}_{1}\), \(S^{p}_{2}\), \(S^{p}_{3}\), \(S^{p}_{4}\))
 (Fig. 9),
 three improvement processes:

  (i) initial system \(S'\) is transformed into Pareto-efficient
  solution \(S_{2}^{p}\),

 (ii) initial system \(S'\) is transformed into ideal solution
  \(S^{I}\).

 (iii) Pareto-efficient solution \(S_{3}^{p}\)
 is transformed into ideal solution \(S^{I}\).

\begin{center}
\begin{picture}(70,43)

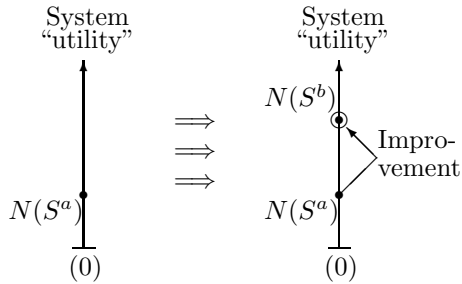
\put(00,00){\makebox(0,0)[bl]{Fig. 8. Scale of
  ``utility'', improvement}}

\put(08.5,10){\line(1,0){3}}

\put(08,05.5){\makebox(0,0)[bl]{(\(0\))}}

\put(10,10){\vector(0,1){25}}

\put(05,39){\makebox(0,0)[bl]{System}}
\put(04,36){\makebox(0,0)[bl]{``utility''}}

\put(10,17){\circle*{1.2}}

\put(00,13){\makebox(0,0)[bl]{\(N(S^{a})\)}}


\put(22,26){\makebox(0,0)[bl]{\(\Longrightarrow\)}}
\put(22,22){\makebox(0,0)[bl]{\(\Longrightarrow\)}}
\put(22,18){\makebox(0,0)[bl]{\(\Longrightarrow\)}}


\put(42.5,10){\line(1,0){3}}

\put(39,39){\makebox(0,0)[bl]{System}}
\put(38,36){\makebox(0,0)[bl]{``utility''}}

\put(44,10){\vector(0,1){25}}
\put(42,05.5){\makebox(0,0)[bl]{(\(0\))}}

\put(44,17){\circle*{1.2}}
\put(34,13){\makebox(0,0)[bl]{\(N(S^{a})\)}}

\put(44,27){\circle*{1}} \put(44,27){\circle{2}}

\put(34,28){\makebox(0,0)[bl]{\(N(S^{b})\)}}

\put(44,17){\line(1,1){5}}

\put(49,22){\vector(-1,1){4}}

\put(49.1,22.6){\makebox(0,0)[bl]{Impro-}}
\put(49.1,19.6){\makebox(0,0)[bl]{vement}}

\end{picture}
%
\begin{picture}(80,41)

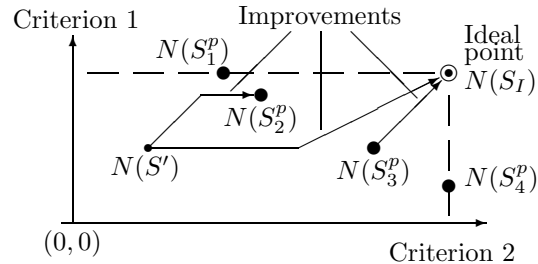
\put(03,00){\makebox(0,0)[bl]{Fig. 9. Multicriteria
 ``utility'', improvement}}

\put(08.5,10){\line(1,0){3}}

\put(06,05.5){\makebox(0,0)[bl]{\((0,0)\)}}

\put(10,10){\vector(0,1){25}} \put(10,10){\vector(1,0){55}}

\put(52,05){\makebox(0,0)[bl]{Criterion 2}}
\put(02,36){\makebox(0,0)[bl]{Criterion 1}}

\put(20,20){\circle*{1.2}}
\put(15,15.5){\makebox(0,0)[bl]{\(N(S')\)}}


\put(32,35.8){\makebox(0,0)[bl]{Improvements}}

\put(39,35.5){\line(-1,-1){8}}

\put(43,35.5){\line(0,-1){13}}

\put(47,35.5){\line(1,-1){8.8}}

\put(20,20){\line(1,1){7}}

\put(27,27){\vector(1,0){7}}

\put(20,20){\line(1,0){20}} \put(40,20){\vector(2,1){18.7}}

\put(50,20){\vector(1,1){09}}



\put(12,30){\line(1,0){4}} \put(18,30){\line(1,0){4}}
\put(24,30){\line(1,0){4}} \put(30,30){\line(1,0){4}}
\put(36,30){\line(1,0){4}} \put(42,30){\line(1,0){4}}
\put(48,30){\line(1,0){4}} \put(54,30){\line(1,0){4}}



\put(60,11){\line(0,1){4}} \put(60,17){\line(0,1){4}}
\put(60,23){\line(0,1){4}}



\put(60,30){\circle*{1}}\put(60,30){\circle{2}}

\put(62,34){\makebox(0,0)[bl]{Ideal}}
\put(62,31){\makebox(0,0)[bl]{point}}
\put(62,27){\makebox(0,0)[bl]{\(N(S_{I})\)}}


\put(30,30){\circle*{1.7}}
\put(21,31){\makebox(0,0)[bl]{\(N(S_{1}^{p})\)}}

\put(35,27){\circle*{1.7}}
\put(30,22){\makebox(0,0)[bl]{\(N(S_{2}^{p})\)}}

\put(50,20){\circle*{1.7}}
\put(45,15){\makebox(0,0)[bl]{\(N(S_{3}^{p})\)}}

\put(60,15){\circle*{1.7}}
\put(62,14){\makebox(0,0)[bl]{\(N(S_{4}^{p})\)}}

\end{picture}
\end{center}

 {\bf Method 3:}~ poset-like scales (or lattices) for system
 quality (case 3.1)
 (e.g., \cite{lev98,levf01,lev06,lev12morph}).

 Here,
 Hierarchical Morphological Multicriteria Design (HMMD) approach
 is used \cite{lev98,lev06,lev12morph},
 which  is based on the morphological clique problem.
 The composite (modular, decomposable) system under examination consists
 of the components and their interconnections or compatibilities.
%
 The designations are:
  ~(1) design alternatives (DAs) for
  leaf nodes of the tree-like model;
  ~(2) priorities of DAs (\(r=\overline{1,k}\);
      \(1\) corresponds to the best level);
  ~(3) ordinal compatibility estimates for each pair of DAs
  (\(w=\overline{0,l}\); \(l\) corresponds to the best level).
 The system consists of design alternatives (DAs) for system parts
 (\(P(1),...,P(i),...,P(m)\)):
 ~ \(S=S(1)\star ...\star S(i)\star ...\star S(m)\)~~
 {\it of}~ DAs
 (one representative design alternative \(S(i)\)),
  with non-zero
  pair interconnection (pair compatibility
  IC) between the selected DAs.
  A discrete space (poset, lattice) of the system excellence on the basis of the
 following vector is used:
 ~~\(N(S)=(w(S);n(S))\),
 ~where \(w(S)\) is the minimum of pairwise compatibility
 between DAs which correspond to different system components
 (i.e.,
 \(~\forall ~P_{j_{1}}\) and \( P_{j_{2}}\),
 \(1 \leq j_{1} \neq j_{2} \leq m\))
 in \(S\),
 ~\(n(S)=(n_{1},...,n_{r},...n_{k})\),
 ~where ~\(n_{r}\) is the number of DAs of the \(r\)th quality in ~\(S\)
 ~(\(\sum^{k}_{r=1} n_{r} = m \)).

 A three-component system
 ~\(S = X \star Y \star Z\)
 is presented  as an illustrative example.
 Ordinal scale for elements (priorities) is \([1,2,3]\), ordinal
 scale for compatibility is \([1,2,3]\).
 For this case,
 Fig. 10 depicts the poset of system quality by components and
 Fig. 11 depicts an integrated poset with compatibility
 (each triangle corresponds to poset from Fig. 10).
 This is {\it case 3.1.}

 Fig. 10 and Fig. 11 illustrate the improvement processes:

 {\it Improvement A:}~
  poset-like scale for total quality of system elements,
  (Fig. 10),
  initial system \(S^{a}\) (\(n(S^{a}) = (0,2,1)\)) is transformed into improved system
  (by components)
 \(S^{b}\) (\(n(S^{b}) = (2,0,1)\)).

 {\it Improvement B:}~
 integrated poset-like scale for total quality of system elements and their compatibility
 (ordinal scale is used for estimates of compatibility ),
 ideal point \(S^{I}\),
 three Pareto-efficient solutions ( \(S^{p}_{1}\), \(S^{p}_{2}\), \(S^{p}_{3}\))
 (Fig. 11),
 three improvement processes:

  (a) initial system \(S'\) is transformed into Pareto-efficient
  solution \(S_{1}^{p}\),

  (b) initial system \(S'\) is transformed into ideal solution
  \(S^{I}\),
  and

  (c) Pareto-efficient solution \(S_{3}^{p}\)
 is transformed into ideal solution \(S^{I}\).

    Generally, the following layers of system excellence can be considered:
  ~(i) ideal point;
  ~(ii) Pareto-efficient points; and
  ~(iii) a neighborhood of Pareto-efficient DAs
 (e.g., a composite decision of this set can be
 transformed into a Pareto-efficient point on the basis of a
 simple
 improvement action(s) as modification of the only one  element).
 The compatibility component of vector ~\(N(S)\)
 can be considered on the basis of a poset-like scale too
 (as \(n(S)\))
  (\cite{levf01,lev06}).
 In this case, the discrete space of
 system excellence will be an analogical lattice.

\begin{center}
\begin{picture}(74,77)

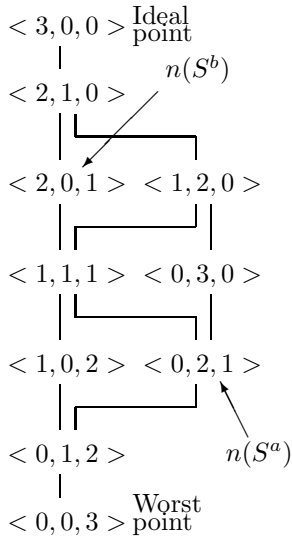
\put(00,00){\makebox(0,0)[bl] {Fig. 10. System quality by elements
 \(n(S)\)}}

\put(05,71){\makebox(0,0)[bl]{\(<3,0,0>\) }}

\put(21.6,73){\makebox(0,0)[bl]{Ideal}}
\put(21.6,70){\makebox(0,0)[bl]{point}}

\put(12,67){\line(0,1){3}}
\put(05,62){\makebox(0,0)[bl]{\(<2,1,0>\)}}

\put(26,65){\makebox(0,0)[bl]{\(n(S^{b})\)}}
\put(25,64){\vector(-1,-1){10}}

\put(12,55){\line(0,1){6}}
\put(05,50){\makebox(0,0)[bl]{\(<2,0,1>\) }}

\put(12,43){\line(0,1){6}}
\put(05,38){\makebox(0,0)[bl]{\(<1,1,1>\) }}

\put(12,31){\line(0,1){6}}
\put(05,26){\makebox(0,0)[bl]{\(<1,0,2>\) }}


\put(12,19){\line(0,1){6}}
\put(05,14){\makebox(0,0)[bl]{\(<0,1,2>\) }}

\put(34,14){\makebox(0,0)[bl]{\(n(S^{a})\)}}
\put(37,18){\vector(-1,2){3.7}}

\put(12,10){\line(0,1){3}}

\put(05,05){\makebox(0,0)[bl]{\(<0,0,3>\) }}

\put(21.6,08){\makebox(0,0)[bl]{Worst}}
\put(21.6,05){\makebox(0,0)[bl]{point}}

\put(14,58){\line(0,1){3}} \put(30,58){\line(-1,0){16}}
\put(30,55){\line(0,1){3}}

\put(23,50){\makebox(0,0)[bl]{\(<1,2,0>\) }}

\put(30,49){\line(0,-1){3}} \put(30,46){\line(-1,0){16}}
\put(14,46){\line(0,-1){3}}
\put(32,43){\line(0,1){6}}
\put(23,38){\makebox(0,0)[bl]{\(<0,3,0>\) }}

\put(14,34){\line(0,1){3}} \put(30,34){\line(-1,0){16}}
\put(30,31){\line(0,1){3}}

\put(32,31){\line(0,1){6}}
\put(23,26){\makebox(0,0)[bl]{\(<0,2,1>\) }}

\put(30,25){\line(0,-1){3}} \put(30,22){\line(-1,0){16}}
\put(14,22){\line(0,-1){3}}

\end{picture}
%
\begin{picture}(75,59)

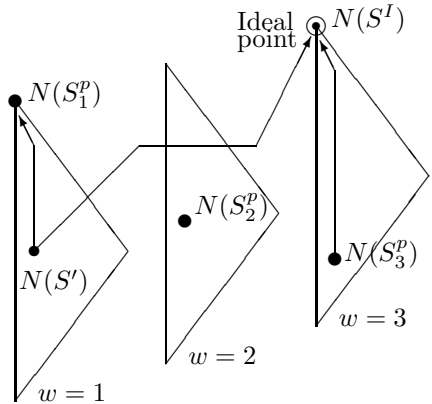
\put(00,00){\makebox(0,0)[bl]{Fig. 11. System quality with
 compatibility \(N(S)\)}}

\put(00,06){\line(0,1){40}} \put(00,06){\line(3,4){15}}
\put(00,046){\line(3,-4){15}}

\put(20,011){\line(0,1){40}} \put(20,011){\line(3,4){15}}
\put(20,051){\line(3,-4){15}}

\put(40,016){\line(0,1){40}} \put(40,016){\line(3,4){15}}
\put(40,056){\line(3,-4){15}}

\put(00,46){\circle*{1.7}}
\put(01.6,45){\makebox(0,0)[bl]{\(N(S^{p}_{1})\)}}

\put(22.5,30){\circle*{1.7}}
\put(23.8,30){\makebox(0,0)[bl]{\(N(S^{p}_{2})\)}}

\put(42.5,25){\circle*{1.7}}
\put(43.7,23.8){\makebox(0,0)[bl]{\(N(S^{p}_{3})\)}}

\put(02.5,26){\circle*{1.3}}
\put(00.5,20){\makebox(0,0)[bl]{\(N(S')\)}}


\put(02.5,26){\line(0,1){14}} \put(02.5,40){\vector(-1,2){2}}


\put(02.5,26){\line(1,1){14}} \put(16.5,40){\line(1,0){15.5}}
\put(32,40){\vector(1,2){7.3}}


\put(42.5,25){\line(0,1){25}} \put(42.5,50){\vector(-1,2){2}}


\put(40,56){\circle*{1}} \put(40,56){\circle{2.5}}

\put(29.5,55.5){\makebox(0,0)[bl]{Ideal}}
\put(29.5,52.5){\makebox(0,0)[bl]{point}}

\put(42,55){\makebox(0,0)[bl]{\(N(S^{I})\)}}

\put(03,06){\makebox(0,0)[bl]{\(w=1\)}}
\put(23,11){\makebox(0,0)[bl]{\(w=2\)}}
\put(43,16){\makebox(0,0)[bl]{\(w=3\)}}

\end{picture}
\end{center}

 {\bf Method 4:}~
 In \cite{lev12a},
 analogical poset-like system quality domains have been suggested
 in the case of interval multi-set estimates for
 DAs (or/and for system compatibility).
 Fig. 12 depicts the poset-like scale for
 the interval multiset estimate (\(3\) position, \(3\) assessment element).

 This system evaluation case can be used for the previous method 3.
 On the other hand,
 this system evaluation approach can be very useful for system improvement by components
 and system extension.
 Here, the total system estimate (i.e., estimate of system
 quality) is considered as the following approaches to aggregation
 of interval estimates of system components \cite{lev12a}:
 (a) an integrated interval multiset estimate,
 (b) median-like interval multiset estimate.

 Fig. 13 depicts an example of this kind of system transformation (reconfiguration):
 ~(a)  replacement of component: \(X_{1} \Longrightarrow X_{2}\),
 ~(b) deletion of component: \(Z_{1}\),
 ~(c) addition of two-component part: \(U \star V\).
 Interval multiset estimates for system components and for system
 quality are depicted in parentheses (Fig. 13).
 Median-like interval multiset estimate is used for evaluation of the system quality.

\begin{center}
\begin{picture}(80,89)
\put(17,00){\makebox(0,0)[bl] {Fig. 12. Scale, estimates
 \cite{lev12a}
 }}

\put(20,85){\makebox(0,0)[bl] {Ideal point}}


\put(25,78.7){\makebox(0,0)[bl]{\(e^{3,3}_{1}\) }}

\put(28,81){\oval(16,5)} \put(28,81){\oval(16.5,5.5)}


\put(42,79){\makebox(0,0)[bl]{\(\{1,1,1\}\) or \((3,0,0)\) }}

\put(00,80.5){\line(0,1){07.5}} \put(04,80.5){\line(0,1){07.5}}

\put(00,83){\line(1,0){04}} \put(00,85.5){\line(1,0){04}}
\put(00,88){\line(1,0){4}}

\put(00,80.5){\line(1,0){12}}

\put(00,79){\line(0,1){3}} \put(04,79){\line(0,1){3}}
\put(08,79){\line(0,1){3}} \put(12,79){\line(0,1){3}}

\put(01.5,76.5){\makebox(0,0)[bl]{\(1\)}}
\put(05.5,76.5){\makebox(0,0)[bl]{\(2\)}}
\put(09.5,76.5){\makebox(0,0)[bl]{\(3\)}}


\put(28,72){\line(0,1){6}}


\put(25,66.7){\makebox(0,0)[bl]{\(e^{3,3}_{2}\) }}

\put(28,69){\oval(16,5)}


\put(42,67){\makebox(0,0)[bl]{\(\{1,1,2\}\) or \((2,1,0)\) }}

\put(00,70.5){\line(0,1){05}} \put(04,70.5){\line(0,1){05}}
\put(8,70.5){\line(0,1){02.5}}

\put(00,73){\line(1,0){8}} \put(00,75.5){\line(1,0){4}}

\put(00,70.5){\line(1,0){12}}

\put(00,69){\line(0,1){3}} \put(04,69){\line(0,1){3}}
\put(08,69){\line(0,1){3}} \put(12,69){\line(0,1){3}}

\put(01.5,66.5){\makebox(0,0)[bl]{\(1\)}}
\put(05.5,66.5){\makebox(0,0)[bl]{\(2\)}}
\put(09.5,66.5){\makebox(0,0)[bl]{\(3\)}}


\put(28,60){\line(0,1){6}}

\put(25,54.7){\makebox(0,0)[bl]{\(e^{3,3}_{3}\) }}

\put(28,57){\oval(16,5)}


\put(42,55){\makebox(0,0)[bl]{\(\{1,2,2\}\) or \((1,2,0)\) }}

\put(00,60.5){\line(0,1){02.5}} \put(04,60.5){\line(0,1){05}}
\put(8,60.5){\line(0,1){05}}

\put(00,63){\line(1,0){8}} \put(04,65.5){\line(1,0){4}}

\put(00,60.5){\line(1,0){12}}

\put(00,59){\line(0,1){3}} \put(04,59){\line(0,1){3}}
\put(08,59){\line(0,1){3}} \put(12,59){\line(0,1){3}}

\put(01.5,56.5){\makebox(0,0)[bl]{\(1\)}}
\put(05.5,56.5){\makebox(0,0)[bl]{\(2\)}}
\put(09.5,56.5){\makebox(0,0)[bl]{\(3\)}}


\put(28,51){\line(0,1){3}}

\put(25,45.7){\makebox(0,0)[bl]{\(e^{3,3}_{4}\) }}

\put(28,48){\oval(16,5)}


\put(45,47){\makebox(0,0)[bl]{\(\{2,2,2\}\) or \((0,3,0)\) }}

\put(04,49.5){\line(0,1){06}} \put(08,49.5){\line(0,1){06}}

\put(04,51.5){\line(1,0){04}} \put(04,53.5){\line(1,0){04}}
\put(04,55.5){\line(1,0){04}}

\put(00,49.5){\line(1,0){12}}

\put(00,48){\line(0,1){3}} \put(04,48){\line(0,1){3}}
\put(08,48){\line(0,1){3}} \put(12,48){\line(0,1){3}}

\put(01.5,45.5){\makebox(0,0)[bl]{\(1\)}}
\put(05.5,45.5){\makebox(0,0)[bl]{\(2\)}}
\put(09.5,45.5){\makebox(0,0)[bl]{\(3\)}}


\put(28,36){\line(0,1){9}}

\put(25,30.7){\makebox(0,0)[bl]{\(e^{3,3}_{6}\) }}

\put(28,33){\oval(16,5)}


\put(42,31){\makebox(0,0)[bl]{\(\{2,2,3\}\) or \((0,2,1)\) }}

\put(04,31.5){\line(0,1){05}} \put(08,31.5){\line(0,1){05}}
\put(12,31.5){\line(0,1){02.5}}

\put(04,34){\line(1,0){8}} \put(04,36.5){\line(1,0){4}}

\put(00,31.5){\line(1,0){12}}

\put(00,30){\line(0,1){3}} \put(04,30){\line(0,1){3}}
\put(08,30){\line(0,1){3}} \put(12,30){\line(0,1){3}}

\put(01.5,27.5){\makebox(0,0)[bl]{\(1\)}}
\put(05.5,27.5){\makebox(0,0)[bl]{\(2\)}}
\put(09.5,27.5){\makebox(0,0)[bl]{\(3\)}}


\put(28,24){\line(0,1){6}}

\put(25,18.7){\makebox(0,0)[bl]{\(e^{3,3}_{7}\) }}

\put(28,21){\oval(16,5)}


\put(42,19){\makebox(0,0)[bl]{\(\{2,3,3\}\) or \((0,1,2)\) }}

\put(04,21.5){\line(0,1){02.5}} \put(08,21.5){\line(0,1){05}}
\put(12,21.5){\line(0,1){05}}

\put(04,24){\line(1,0){8}} \put(08,26.5){\line(1,0){4}}

\put(00,21.5){\line(1,0){12}}

\put(00,20){\line(0,1){3}} \put(04,20){\line(0,1){3}}
\put(08,20){\line(0,1){3}} \put(12,20){\line(0,1){3}}

\put(01.5,17.5){\makebox(0,0)[bl]{\(1\)}}
\put(05.5,17.5){\makebox(0,0)[bl]{\(2\)}}
\put(09.5,17.5){\makebox(0,0)[bl]{\(3\)}}


\put(28,12){\line(0,1){6}}


\put(25,06.7){\makebox(0,0)[bl]{\(e^{3,3}_{8}\) }}

\put(28,09){\oval(16,5)}


\put(42,7){\makebox(0,0)[bl]{\(\{3,3,3\}\) or \((0,0,3)\) }}

\put(08,9){\line(0,1){07.5}} \put(12,9){\line(0,1){07.5}}
\put(08,11.5){\line(1,0){4}} \put(08,14){\line(1,0){4}}
\put(08,16.5){\line(1,0){4}}

\put(00,9){\line(1,0){12}}

\put(00,07.5){\line(0,1){3}} \put(04,07.5){\line(0,1){3}}
\put(08,07.5){\line(0,1){3}} \put(12,07.5){\line(0,1){3}}

\put(01.5,5){\makebox(0,0)[bl]{\(1\)}}
\put(05.5,5){\makebox(0,0)[bl]{\(2\)}}
\put(09.5,5){\makebox(0,0)[bl]{\(3\)}}


\put(45.5,45.5){\line(-1,1){09.5}}

\put(45.5,38.5){\line(-3,-1){10}}


\put(45,39.7){\makebox(0,0)[bl]{\(e^{3,3}_{5}\) }}

\put(48,42){\oval(16,5)}


\put(58,40.5){\makebox(0,0)[bl]{\(\{1,2,3\}\) or \((1,1,1)\) }}

\put(00,41.5){\line(0,1){02.5}} \put(04,41.5){\line(0,1){02.5}}
\put(8,41.5){\line(0,1){02.5}} \put(12,41.5){\line(0,1){02.5}}

\put(00,44){\line(1,0){12}}

\put(00,41.5){\line(1,0){12}}

\put(00,40){\line(0,1){3}} \put(04,40){\line(0,1){3}}
\put(08,40){\line(0,1){3}} \put(12,40){\line(0,1){3}}

\put(01.5,37.5){\makebox(0,0)[bl]{\(1\)}}
\put(05.5,37.5){\makebox(0,0)[bl]{\(2\)}}
\put(09.5,37.5){\makebox(0,0)[bl]{\(3\)}}


\end{picture}
\end{center}

\begin{center}
\begin{picture}(117,35)
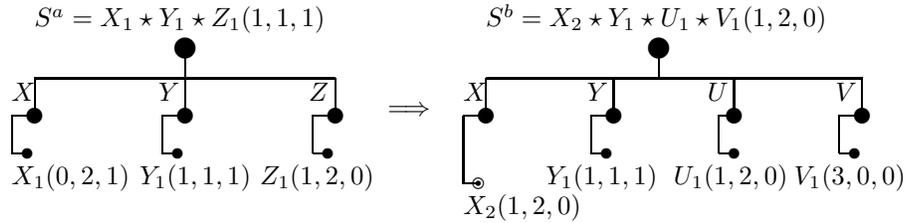
\put(06.5,00){\makebox(0,0)[bl]{Fig. 13. System improvements (by
components and by extension)}}

\put(03,30){\makebox(0,8)[bl]{\(S^{a} = X_{1} \star Y_{1} \star
 Z_{1} (1,1,1)\)}}

\put(3,24){\line(1,0){40}} \put(23,24){\line(0,1){04}}
\put(23,28){\circle*{2.8}}

\put(3,24){\line(0,-1){04}} \put(23,24){\line(0,-1){04}}
\put(43,24){\line(0,-1){04}}

\put(0,21){\makebox(0,8)[bl]{\(X\)}}
\put(19.5,21){\makebox(0,8)[bl]{\(Y\)}}
\put(39.5,21){\makebox(0,8)[bl]{\(Z\)}}

\put(3,19){\circle*{2}} \put(23,19){\circle*{2}}
\put(43,19){\circle*{2}}


\put(00,09){\makebox(0,8)[bl]{\(X_{1}(0,2,1)\)}}


\put(0,14){\line(1,0){02}} \put(2,14){\circle*{1.5}}

\put(0,19){\line(0,-1){5}} \put(0,19){\line(1,0){02}}

\put(17,09){\makebox(0,8)[bl]{\(Y_{1}(1,1,1)\)}}

\put(20,14){\line(1,0){02}} \put(22,14){\circle*{1.5}}

\put(20,19){\line(0,-1){5}} \put(20,19){\line(1,0){02}}


\put(33,09){\makebox(0,8)[bl]{\(Z_{1}(1,2,0)\)}}

\put(40,14){\line(1,0){02}}

\put(42,14){\circle*{1.5}}

\put(40,19){\line(0,-1){5}} \put(40,19){\line(1,0){02}}


\put(50,19){\makebox(0,8)[bl]{\(\Longrightarrow \)}}


\put(63,30){\makebox(0,8)[bl]{\(S^{b} = X_{2} \star Y_{1}
 \star U_{1} \star V_{1} (1,2,0)\)}}

\put(63,24){\line(1,0){50}} \put(86,24){\line(0,1){04}}
\put(86,28){\circle*{2.8}}

\put(63,24){\line(0,-1){04}} \put(80,24){\line(0,-1){04}}
\put(96,24){\line(0,-1){04}} \put(113,24){\line(0,-1){04}}

\put(60,21){\makebox(0,8)[bl]{\(X\)}}
\put(76.5,21){\makebox(0,8)[bl]{\(Y\)}}
\put(92.5,21){\makebox(0,8)[bl]{\(U\)}}
\put(109.5,21){\makebox(0,8)[bl]{\(V\)}}

\put(63,19){\circle*{2}} \put(80,19){\circle*{2}}
\put(96,19){\circle*{2}} \put(113,19){\circle*{2}}


\put(60,05){\makebox(0,8)[bl]{\(X_{2}(1,2,0)\)}}

\put(60,10){\line(1,0){02}}  \put(62,10){\circle{1.5}}

\put(62,10){\circle*{0.7}}

\put(60,19){\line(0,-1){9}} \put(60,19){\line(1,0){02}}


\put(71,09){\makebox(0,8)[bl]{\(Y_{1}(1,1,1)\)}}

\put(77,14){\line(1,0){02}} \put(79,14){\circle*{1.5}}

\put(77,19){\line(0,-1){5}} \put(77,19){\line(1,0){02}}


\put(88,09){\makebox(0,8)[bl]{\(U_{1}(1,2,0)\)}}

\put(94,14){\line(1,0){02}} \put(96,14){\circle*{1.5}}

\put(94,19){\line(0,-1){5}} \put(94,19){\line(1,0){02}}


\put(104,09){\makebox(0,8)[bl]{\(V_{1}(3,0,0)\)}}

\put(110,14){\line(1,0){02}} \put(112,14){\circle*{1.5}}

\put(110,19){\line(0,-1){5}} \put(110,19){\line(1,0){02}}

\end{picture}
\end{center}

\subsection{Towards Reoptimization \cite{lev11restr}}

 In recent several years,
 a special class of combinatorial optimization
 problems as  ``reoptimization'' has been
 studied for several well-known problems (Table 3).
 In general, the reoptimization problem is formulated as follows:

~~

 {\it Given}:
 (i) an instance of the combinatorial problem over a graph
 and corresponding optimal solution,
 (ii) some ``small'' perturbations
 (i.e., modifications) on this instance
 (e.g., node-insertion, node-deletion).

 {\it Question}:~~
 {\it Is it possible to compute a new good (optimal or near-optimal)
 solution
 subject to minor modifications?}

~~

\begin{center}
\begin{picture}(83,35)

\put(13,31){\makebox(0,0)[bl]{Table 3. Studies of
 reoptimization}}

\put(00,00){\line(1,0){83}} \put(00,23){\line(1,0){83}}
\put(00,29){\line(1,0){83}}

\put(00,00){\line(0,1){29}} \put(66,00){\line(0,1){29}}
\put(83,00){\line(0,1){29}}


\put(01,25){\makebox(0,0)[bl]{Combinatorial optimization problem}}
\put(67,25){\makebox(0,0)[bl]{Sources }}


\put(01,18){\makebox(0,0)[bl]{1. Minimum spanning tree problem}}
\put(67,18){\makebox(0,0)[bl]{\cite{boria10}}}


\put(01,14){\makebox(0,0)[bl]{2. Traveling salesman problems}}
\put(67,14){\makebox(0,0)[bl]{\cite{archetti03,aus09}}}


\put(01,10){\makebox(0,0)[bl]{3. Steiner tree problems}}
\put(67,10){\makebox(0,0)[bl]{\cite{bilo08,esc09}}}


\put(01,06){\makebox(0,0)[bl]{4. Covering problems}}
\put(67,06){\makebox(0,0)[bl]{\cite{bilo08a}}}


\put(01,02){\makebox(0,0)[bl]{5. Shortest common superstring
problem}}

\put(67,02){\makebox(0,0)[bl]{\cite{bilo11}}}


\end{picture}
\end{center}

 A survey of complexity issues for reoptimization problems
 is presented in \cite{bok08}.
 Mainly, the problems belong to class of NP-hard problems and
 various approximation algorithms have been suggested.

 Another approach to modification in
 combinatorial optimization problems as ``restructuring''
 has been suggested in \cite{lev11restr}.
 The approach corresponds to many applied
 reengineering (redesign) problems in  existing modular systems.
 The restructuring process is illustrated in Fig. 14 \cite{lev11restr}.
 Here, modifications are based on insertion/deletion of elements
 (i.e., elements, nodes, arcs)
 and changes of a structure as well.
 Two main features of the restructuring process are examined:
 (i) a cost of the initial problem solution restructuring
 (i.e., cost of the selected modifications),
 (ii) a closeness the obtained restructured solution to a goal solution.

\begin{center}
\begin{picture}(80,52)

\put(01.5,00){\makebox(0,0)[bl]{Fig. 14. Illustration for
restructuring process \cite{lev11restr}}}

\put(00,09){\vector(1,0){80}}

\put(00,7.5){\line(0,1){3}} \put(11,7.5){\line(0,1){3}}
\put(69,7.5){\line(0,1){3}}

\put(00,05){\makebox(0,0)[bl]{\(0\)}}
\put(11,05){\makebox(0,0)[bl]{\(\tau_{1}\)}}
\put(67,05){\makebox(0,0)[bl]{\(\tau_{2}\)}}

\put(79,05.3){\makebox(0,0)[bl]{\(t\)}}


\put(00,41){\line(1,0){22}} \put(00,51){\line(1,0){22}}
\put(00,41){\line(0,1){10}} \put(22,41){\line(0,1){10}}

\put(0.5,46){\makebox(0,0)[bl]{Requirements}}
\put(0.5,42){\makebox(0,0)[bl]{(for \(\tau_{1}\))}}


\put(11,41){\vector(0,-1){4}}

\put(00,23){\line(1,0){22}} \put(00,37){\line(1,0){22}}
\put(00,23){\line(0,1){14}} \put(22,23){\line(0,1){14}}

\put(0.5,32){\makebox(0,0)[bl]{Optimization}}
\put(0.5,28){\makebox(0,0)[bl]{problem}}
\put(0.5,24){\makebox(0,0)[bl]{(for \(\tau_{1})\)}}


\put(11,23){\vector(0,-1){4}}

\put(11,16){\oval(22,06)}

\put(02,15){\makebox(0,0)[bl]{Solution \(S^{1}\)}}


\put(26,14){\line(1,0){28}} \put(26,49){\line(1,0){28}}
\put(26,14){\line(0,1){35}} \put(54,14){\line(0,1){35}}

\put(26.5,14.5){\line(1,0){27}} \put(26.5,48.5){\line(1,0){27}}
\put(26.5,14.5){\line(0,1){34}} \put(53.5,14.5){\line(0,1){34}}

\put(28,44){\makebox(0,0)[bl]{Restructuring:}}
\put(28,40){\makebox(0,0)[bl]{\(S^{1} \Rightarrow S^{*}  \)}}
\put(28,35.5){\makebox(0,0)[bl]{while taking}}
\put(28,32.5){\makebox(0,0)[bl]{into account:}}
\put(28,28){\makebox(0,0)[bl]{(i) \(S^{*}\) is close}}
\put(28,24.4){\makebox(0,0)[bl]{to \(S^{2}\),}}
\put(28,20){\makebox(0,0)[bl]{(ii) change of \(S^{1}\)}}
\put(28,16){\makebox(0,0)[bl]{into \(S^{*}\) is cheap.}}



\put(58,41){\line(1,0){22}} \put(58,51){\line(1,0){22}}
\put(58,41){\line(0,1){10}} \put(80,41){\line(0,1){10}}

\put(58.5,46){\makebox(0,0)[bl]{Requirements}}
\put(58.5,42){\makebox(0,0)[bl]{(for \(\tau_{2}\))}}


\put(69,41){\vector(0,-1){4}}

\put(58,23){\line(1,0){22}} \put(58,37){\line(1,0){22}}
\put(58,23){\line(0,1){14}} \put(80,23){\line(0,1){14}}

\put(58.5,32){\makebox(0,0)[bl]{Optimization}}
\put(58.5,28){\makebox(0,0)[bl]{problem }}
\put(58.5,24){\makebox(0,0)[bl]{(for \(\tau_{2})\)}}


\put(69,22){\vector(0,-1){4}}

\put(69,16){\oval(22,06)}

\put(59,15){\makebox(0,0)[bl]{Solution \(S^{2}\)}}


\end{picture}
\end{center}

  The optimization problem is solved for two time moments:
 \(\tau_{1}\) and  \(\tau_{2}\) to obtain corresponding solutions
 \(S^{1}\) and \(S^{2}\).
 The examined restructuring problem consists in
 a ``cheap'' transformation (change) of solution \(S^{1}\) to a solution \(S^{*}\) that
 is very close to \(S^{2}\).
 In \cite{lev11restr},
 this restructuring approach is described and illustrated for the following combinatorial
 optimization problems:
 knapsack problem,
 multiple choice problem,
 assignment problem,
 spanning tree problems.

 Fig. 15 depicts  the restructuring problem \cite{lev11restr}.

 Let \(P\) be a combinatorial optimization problem with a solution as
 structure
 \(S\)
 (i.e., subset, graph),
 \(\Omega\) be initial data (elements, element parameters, etc.),
 \(f(P)\) be objective function(s).
 Thus, \(S(\Omega)\) be a solution for initial data \(\Omega\),
 \(f(S(\Omega))\) be the corresponding objective function.
 Let \(\Omega^{1}\) be initial data at an initial stage,
  \(f(S(\Omega^{1}))\) be the corresponding objective function.
 \(\Omega^{2}\) be initial data at next stage,
  \(f(S(\Omega^{2}))\) be the corresponding objective function.
 As a result,
 the following solutions can be considered:
 ~(a) \( S^{1}=S(\Omega^{1})\) with \(f(S(\Omega^{1}))\) and
 ~(b) \( S^{2}=S(\Omega^{2})\) with \(f(S(\Omega^{2}))\).
 In addition it is reasonable to examine a cost of changing
 a solution into another one:~
 \( H(S^{\alpha} \rightarrow  S^{\beta})\).
 Let \(\rho ( S^{\alpha}, S^{\beta} )\) be a proximity between solutions
  \( S^{\alpha}\) and \( S^{\beta}\),
  for example,
 \(\rho ( S^{\alpha}, S^{\beta} ) = | f(S^{\alpha}) -  f(S^{\beta}) |\).
 Note, function \(f(S)\) is often a vector function.
 Finally, the restructuring problem is
 (a basic version):

~~

 {\it Find solution} \( S^{*}\)
 {\it while taking into account the following}:

  (i) \( H(S^{1} \rightarrow  S^{*}) \rightarrow \min \),
%
  ~(ii) \(\rho ( S^{*}, S^{2} )  \rightarrow \min  \) ~({\it or constraint}).

~~

 Thus, the basic optimization model can be considered as the following:

%
   \[\min \rho ( S^{*}, S^{2} ) ~~~s.t.
 ~ H(S^{1} \rightarrow  S^{*})  \leq \widehat{h}, \]
 where \(\widehat{h}\) is a constraint for cost of the solution
 change.

 Proximity function  ~\(\rho (S^{*},S^{2}) \)~
  can be considered as a vector function
 (analogically for the solution change cost).
 The situation will lead to a multicriteria restructuring problem
 (i.e., searching for a Pareto-efficient solutions).

\begin{center}
\begin{picture}(73,55)
\put(00,00){\makebox(0,0)[bl]{Fig. 15. Illustration for
restructuring problem \cite{lev11restr}}}

\put(00,05){\vector(0,1){46}} \put(00,05){\vector(1,0){70}}

\put(71,04.7){\makebox(0,0)[bl]{\(t\)}}

\put(00,51){\makebox(0,0)[bl]{``Quality''}}


\put(0.6,15){\makebox(0,0)[bl]{\(S^{1}\)}}
\put(5,15){\circle{1.7}}

\put(12,13){\makebox(0,0)[bl]{Initial}}
\put(12,10){\makebox(0,0)[bl]{solution}}
\put(12,06){\makebox(0,0)[bl]{(\(t=\tau^{1}\))}}

\put(11,10){\line(-4,3){5}}

\put(6,16){\vector(1,1){23}}


\put(40,35){\circle*{2.7}}

\put(54,37){\makebox(0,0)[bl]{Goal}}
\put(54,34){\makebox(0,0)[bl]{solution}}
\put(54,30){\makebox(0,0)[bl]{(\(t=\tau^{2}\)): \(S^{2}\)}}

\put(53,35){\line(-1,0){10.5}}

\put(40,35){\oval(12,10)} \put(40,35){\oval(17,17)}
\put(40,35){\oval(24,22)}


\put(40,35){\vector(-2,1){9}} \put(30,40){\vector(2,-1){9}}



\put(26,17){\makebox(0,0)[bl]{Proximity}}
\put(26,13){\makebox(0,0)[bl]{~\(\rho (S^{*},S^{2})\)}}

\put(36,20){\line(0,1){16}}


\put(46,19){\makebox(0,0)[bl]{Neighborhoods }}
\put(49,16){\makebox(0,0)[bl]{of ~\(S^{2}\)}}

\put(56,22){\line(-1,1){10}}

\put(53,22){\line(-2,1){8}}



\put(30,40){\circle{2}} \put(30,40){\circle*{1}}

\put(11,46){\makebox(0,0)[bl]{Obtained}}
\put(11,43){\makebox(0,0)[bl]{solution \(S^{*}\)}}

\put(20,42.8){\line(4,-1){7}}

\put(1,37.5){\makebox(0,0)[bl]{Solution }}
\put(1,33.5){\makebox(0,0)[bl]{change cost }}
\put(1,29.5){\makebox(0,0)[bl]{\(H(S^{1} \rightarrow S^{*})\)}}

\put(8,29){\line(1,-1){5}}


\end{picture}
\end{center}

\section{Improvement by System Components}

\subsection{Basic Framework}

 Generally, system improvement by components
 is based on improvement/replacement of system element.
  The basic framework to system improvement by system components
 can be considered as follows:

~~

 {\it Stage 1.} Detection of system bottlenecks set.

 {\it Stage 2.} Generation of system improvement actions
 (i.e., improvement of DA, improvement of interconnection between
 DAs) and their assessment.

 {\it Stage 3.} Formulation of the system improvement problem as
 combinatorial optimization problem
 as selection/combination of improvement actions
 (model: multiple choice problem or HMMD in the case of interconnection between the actions).

 {\it Stage 4.} Solving the system improvement problem.

~~

 An example of combinatorial synthesis of composite five-component
 system is presented in Fig. 16.
 Here, HMMD is used
  (e.g., \cite{lev98,lev06,lev09,lev12morph}).
 Ordinal quality of DAS are depicted in Fig. 16 (in parentheses, scale: \([1,2,3,4]\)).
 Table 4 contains ordinal compatibility estimates (scale:
 \([0,3]\)).

\begin{center}
\begin{picture}(80,38)

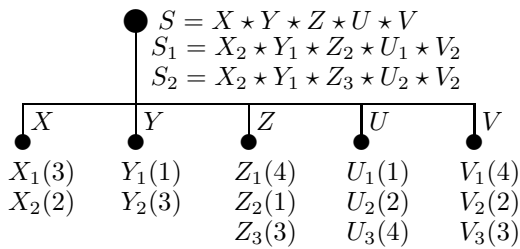
\put(04,00){\makebox(0,0)[bl] {Fig. 16. Example of system
structure}}

\put(1,13){\makebox(0,8)[bl]{\(X_{1}(3)\)}}
\put(1,09){\makebox(0,8)[bl]{\(X_{2}(2)\)}}

\put(16,13){\makebox(0,8)[bl]{\(Y_{1}(1)\)}}
\put(16,09){\makebox(0,8)[bl]{\(Y_{2}(3)\)}}

\put(31,13){\makebox(0,8)[bl]{\(Z_{1}(4)\)}}
\put(31,09){\makebox(0,8)[bl]{\(Z_{2}(1)\)}}
\put(31,05){\makebox(0,8)[bl]{\(Z_{3}(3)\)}}

\put(46,13){\makebox(0,8)[bl]{\(U_{1}(1)\)}}
\put(46,09){\makebox(0,8)[bl]{\(U_{2}(2)\)}}
\put(46,05){\makebox(0,8)[bl]{\(U_{3}(4)\)}}

\put(61,13){\makebox(0,8)[bl]{\(V_{1}(4)\)}}
\put(61,09){\makebox(0,8)[bl]{\(V_{2}(2)\)}}
\put(61,05){\makebox(0,8)[bl]{\(V_{3}(3)\)}}


\put(03,19){\circle*{2}} \put(18,19){\circle*{2}}
\put(33,19){\circle*{2}} \put(48,19){\circle*{2}}
\put(63,19){\circle*{2}}

\put(03,24){\line(0,-1){04}} \put(18,24){\line(0,-1){04}}
\put(33,24){\line(0,-1){04}} \put(48,24){\line(0,-1){04}}
\put(63,24){\line(0,-1){04}}


\put(03,24){\line(1,0){60}}

\put(04,20.5){\makebox(0,8)[bl]{\(X\) }}
\put(19,20.5){\makebox(0,8)[bl]{\(Y\) }}
\put(34,20.5){\makebox(0,8)[bl]{\(Z\) }}
\put(49,20.5){\makebox(0,8)[bl]{\(U\) }}
\put(64,20.5){\makebox(0,8)[bl]{\(V\) }}


\put(18,24){\line(0,1){10}} \put(18,35){\circle*{3}}

\put(21,34){\makebox(0,8)[bl] {\(S = X \star Y \star Z \star U
 \star V \)}}

\put(20,30){\makebox(0,8)[bl] {\(S_{1} = X_{2} \star Y_{1} \star
 Z_{2} \star U_{1} \star V_{2} \)}}

\put(20,26){\makebox(0,8)[bl] {\(S_{2} = X_{2} \star Y_{1} \star
 Z_{3} \star U_{2} \star V_{2} \)}}

\end{picture}
%
\begin{picture}(62,53)

\put(13,49){\makebox(0,0)[bl]{Table 4. Compatibility}}

\put(00,0){\line(1,0){62}} \put(00,42){\line(1,0){62}}
\put(00,48){\line(1,0){62}}

\put(00,0){\line(0,1){48}} \put(07,0){\line(0,1){48}}
\put(62,0){\line(0,1){48}}

\put(01,38){\makebox(0,0)[bl]{\(X_{1}\)}}
\put(01,34){\makebox(0,0)[bl]{\(X_{2}\)}}
\put(01,30){\makebox(0,0)[bl]{\(Y_{1}\)}}
\put(01,26){\makebox(0,0)[bl]{\(Y_{2}\)}}

\put(01,22){\makebox(0,0)[bl]{\(Z_{1}\)}}
\put(01,18){\makebox(0,0)[bl]{\(Z_{2}\)}}

\put(01,14){\makebox(0,0)[bl]{\(Z_{3}\)}}
\put(01,10){\makebox(0,0)[bl]{\(U_{1}\)}}
\put(01,06){\makebox(0,0)[bl]{\(U_{2}\)}}
\put(01,02){\makebox(0,0)[bl]{\(U_{3}\)}}

\put(12,42){\line(0,1){6}} \put(17,42){\line(0,1){6}}
\put(22,42){\line(0,1){6}} \put(27,42){\line(0,1){6}}
\put(32,42){\line(0,1){6}} \put(37,42){\line(0,1){6}}
\put(42,42){\line(0,1){6}} \put(47,42){\line(0,1){6}}
\put(52,42){\line(0,1){6}} \put(57,42){\line(0,1){6}}

\put(07.4,44){\makebox(0,0)[bl]{\(Y_{1}\)}}
\put(12.4,44){\makebox(0,0)[bl]{\(Y_{2}\)}}
\put(17.4,44){\makebox(0,0)[bl]{\(Z_{1}\)}}
\put(22.4,44){\makebox(0,0)[bl]{\(Z_{2}\)}}
\put(27.4,44){\makebox(0,0)[bl]{\(Z_{3}\)}}
\put(32.4,44){\makebox(0,0)[bl]{\(U_{1}\)}}
\put(37.4,44){\makebox(0,0)[bl]{\(U_{2}\)}}
\put(42.4,44){\makebox(0,0)[bl]{\(U_{3}\)}}
\put(47.4,44){\makebox(0,0)[bl]{\(V_{1}\)}}
\put(52.4,44){\makebox(0,0)[bl]{\(V_{2}\)}}
\put(57.4,44){\makebox(0,0)[bl]{\(V_{3}\)}}

\put(09,38){\makebox(0,0)[bl]{\(3\)}}
\put(14,38){\makebox(0,0)[bl]{\(3\)}}
\put(19,38){\makebox(0,0)[bl]{\(2\)}}
\put(24,38){\makebox(0,0)[bl]{\(3\)}}
\put(29,38){\makebox(0,0)[bl]{\(2\)}}
\put(34,38){\makebox(0,0)[bl]{\(3\)}}
\put(39,38){\makebox(0,0)[bl]{\(3\)}}
\put(44,38){\makebox(0,0)[bl]{\(0\)}}
\put(49,38){\makebox(0,0)[bl]{\(3\)}}
\put(54,38){\makebox(0,0)[bl]{\(3\)}}
\put(59,38){\makebox(0,0)[bl]{\(2\)}}

\put(09,34){\makebox(0,0)[bl]{\(3\)}}
\put(14,34){\makebox(0,0)[bl]{\(3\)}}
\put(19,34){\makebox(0,0)[bl]{\(3\)}}
\put(24,34){\makebox(0,0)[bl]{\(3\)}}
\put(29,34){\makebox(0,0)[bl]{\(3\)}}
\put(34,34){\makebox(0,0)[bl]{\(3\)}}
\put(39,34){\makebox(0,0)[bl]{\(3\)}}
\put(44,34){\makebox(0,0)[bl]{\(2\)}}
\put(49,34){\makebox(0,0)[bl]{\(3\)}}
\put(54,34){\makebox(0,0)[bl]{\(3\)}}
\put(59,34){\makebox(0,0)[bl]{\(1\)}}

\put(19,30){\makebox(0,0)[bl]{\(3\)}}
\put(24,30){\makebox(0,0)[bl]{\(3\)}}
\put(29,30){\makebox(0,0)[bl]{\(3\)}}
\put(34,30){\makebox(0,0)[bl]{\(3\)}}
\put(39,30){\makebox(0,0)[bl]{\(3\)}}
\put(44,30){\makebox(0,0)[bl]{\(2\)}}
\put(49,30){\makebox(0,0)[bl]{\(3\)}}
\put(54,30){\makebox(0,0)[bl]{\(3\)}}
\put(59,30){\makebox(0,0)[bl]{\(1\)}}

\put(19,26){\makebox(0,0)[bl]{\(3\)}}
\put(24,26){\makebox(0,0)[bl]{\(3\)}}
\put(29,26){\makebox(0,0)[bl]{\(3\)}}
\put(34,26){\makebox(0,0)[bl]{\(3\)}}
\put(39,26){\makebox(0,0)[bl]{\(2\)}}
\put(44,26){\makebox(0,0)[bl]{\(1\)}}
\put(49,26){\makebox(0,0)[bl]{\(3\)}}
\put(54,26){\makebox(0,0)[bl]{\(2\)}}
\put(59,26){\makebox(0,0)[bl]{\(2\)}}

\put(34,22){\makebox(0,0)[bl]{\(3\)}}
\put(39,22){\makebox(0,0)[bl]{\(1\)}}
\put(44,22){\makebox(0,0)[bl]{\(0\)}}
\put(49,22){\makebox(0,0)[bl]{\(3\)}}
\put(54,22){\makebox(0,0)[bl]{\(1\)}}
\put(59,22){\makebox(0,0)[bl]{\(1\)}}

\put(34,18){\makebox(0,0)[bl]{\(3\)}}
\put(39,18){\makebox(0,0)[bl]{\(0\)}}
\put(44,18){\makebox(0,0)[bl]{\(2\)}}
\put(49,18){\makebox(0,0)[bl]{\(3\)}}
\put(54,18){\makebox(0,0)[bl]{\(3\)}}
\put(59,18){\makebox(0,0)[bl]{\(1\)}}

\put(34,14){\makebox(0,0)[bl]{\(2\)}}
\put(39,14){\makebox(0,0)[bl]{\(3\)}}
\put(44,14){\makebox(0,0)[bl]{\(0\)}}
\put(49,14){\makebox(0,0)[bl]{\(3\)}}
\put(54,14){\makebox(0,0)[bl]{\(3\)}}
\put(59,14){\makebox(0,0)[bl]{\(1\)}}

\put(49,10){\makebox(0,0)[bl]{\(3\)}}
\put(54,10){\makebox(0,0)[bl]{\(1\)}}
\put(59,10){\makebox(0,0)[bl]{\(0\)}}

\put(49,06){\makebox(0,0)[bl]{\(2\)}}
\put(54,06){\makebox(0,0)[bl]{\(3\)}}
\put(59,06){\makebox(0,0)[bl]{\(1\)}}

\put(49,02){\makebox(0,0)[bl]{\(1\)}}
\put(54,02){\makebox(0,0)[bl]{\(3\)}}
\put(59,02){\makebox(0,0)[bl]{\(2\)}}

\end{picture}
\end{center}

 The resultant composite Pareto-efficient DAs are the following (Fig. 17):

 (a)
 \(S_{1} = X_{2} \star Y_{1} \star Z_{2} \star U_{1} \star V_{2}\),
 \(N(S_{1})= (1;3,2,0,0)\) and

 (b)
 \(S_{2} = X_{2} \star Y_{1} \star Z_{3} \star U_{2} \star V_{2}\),
 \(N(S_{2})= (3;2,2,1,0)\).

 Table 5 contains some bottlenecks and improvement actions.
 Further, it is possible to examine a combinatorial synthesis
 problem to design a system improvement plan
 based on the pointed out improvement actions
 (multiple choice problem or HMMD in the case of interconnection between the actions).

  The improvement procedure based on multiple choice problem
 for \(S_{2}\) is considered.
 It is assumed improvement actions
  are compatible.
 Table 6 contains improvement actions
 and their illustrative estimates (ordinal scales).

\begin{center}
\begin{picture}(56,59)
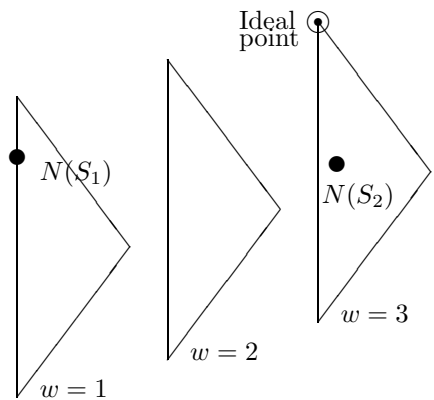
\put(00,00){\makebox(0,0)[bl]{Fig. 17. Poset of system quality}}

\put(00,06){\line(0,1){40}} \put(00,06){\line(3,4){15}}
\put(00,046){\line(3,-4){15}}

\put(20,011){\line(0,1){40}} \put(20,011){\line(3,4){15}}
\put(20,051){\line(3,-4){15}}

\put(40,016){\line(0,1){40}} \put(40,016){\line(3,4){15}}
\put(40,056){\line(3,-4){15}}

\put(00,38){\circle*{2}}
\put(03,34){\makebox(0,0)[bl]{\(N(S_{1})\)}}

\put(42.5,37){\circle*{2}}
\put(40.5,31){\makebox(0,0)[bl]{\(N(S_{2})\)}}

\put(40,56){\circle*{1}} \put(40,56){\circle{2.7}}

\put(29.5,55){\makebox(0,0)[bl]{Ideal}}
\put(29.5,52){\makebox(0,0)[bl]{point}}

\put(03,06){\makebox(0,0)[bl]{\(w=1\)}}
\put(23,11){\makebox(0,0)[bl]{\(w=2\)}}
\put(43,16){\makebox(0,0)[bl]{\(w=3\)}}

\end{picture}
\end{center}

\begin{center}
\begin{picture}(91,45)

\put(09,41){\makebox(0,0)[bl]{Table 5. Bottlenecks and improvement
actions}}

\put(00,00){\line(1,0){91}} \put(00,27){\line(1,0){91}}
\put(46,33){\line(1,0){30}} \put(00,39){\line(1,0){91}}

\put(00,0){\line(0,1){39}} \put(46,0){\line(0,1){39}}
\put(56,0){\line(0,1){33}} \put(76,0){\line(0,1){39}}
\put(91,0){\line(0,1){39}}

\put(01,34.4){\makebox(0,0)[bl]{Composite DAs}}
\put(52,35){\makebox(0,0)[bl]{Bottlenecks}}
\put(47,29){\makebox(0,0)[bl]{DAs}}
\put(64,29){\makebox(0,0)[bl]{IC}}
\put(77,35){\makebox(0,0)[bl]{Actions}}
\put(80,31){\makebox(0,0)[bl]{\(w/r\)}}


\put(01,22){\makebox(0,0)[bl]{\(S_{2}=
 X_{2}\star Y_{1}\star
 Z_{3}\star U_{2}\star V_{4} \)}}

\put(01,18){\makebox(0,0)[bl]{\(S_{2}=
 X_{2}\star Y_{1}\star
 Z_{3}\star U_{2}\star V_{4} \)}}

\put(01,14){\makebox(0,0)[bl]{\(S_{2}=
 X_{2}\star Y_{1}\star
 Z_{3}\star U_{2}\star V_{4} \)}}

\put(01,10){\makebox(0,0)[bl]{\(S_{2}=
 X_{2}\star Y_{1}\star
 Z_{3}\star U_{2}\star V_{4} \)}}

\put(01,06){\makebox(0,0)[bl]{\(S_{2}=
 X_{2}\star Y_{1}\star
 Z_{3}\star U_{2}\star V_{2}\)}}

\put(01,02){\makebox(0,0)[bl]{\(S_{1}=
 X_{2}\star Y_{1}\star
 Z_{2}\star U_{1}\star V_{2}\)}}

\put(79,22){\makebox(0,0)[bl]{\(2 \Rightarrow 1\)}}
\put(79,18){\makebox(0,0)[bl]{\(3 \Rightarrow 2\)}}
\put(79,14){\makebox(0,0)[bl]{\(3 \Rightarrow 1\)}}
\put(79,10){\makebox(0,0)[bl]{\(2 \Rightarrow 1\)}}
\put(79,06){\makebox(0,0)[bl]{\(2 \Rightarrow 1\)}}
\put(79,02){\makebox(0,0)[bl]{\(1 \Rightarrow 3\)}}

\put(48,22){\makebox(0,0)[bl]{\(X_{2}\)}}
\put(48,18){\makebox(0,0)[bl]{\(Z_{3}\)}}
\put(48,14){\makebox(0,0)[bl]{\(Z_{3}\)}}

\put(48,10){\makebox(0,0)[bl]{\(U_{2}\)}}
\put(48,06){\makebox(0,0)[bl]{\(V_{2}\)}}
\put(59,02){\makebox(0,0)[bl]{\((U_{1},V_{2})\)}}

\end{picture}
\end{center}

\begin{center}
\begin{picture}(72,55)

\put(08,51){\makebox(0,0)[bl] {Table 6. Improvement alternatives}}

\put(0,0){\line(1,0){72}} \put(0,39){\line(1,0){72}}
\put(0,49){\line(1,0){72}}


\put(0,0){\line(0,1){49}} \put(05,0){\line(0,1){49}}
\put(18,0){\line(0,1){49}}

\put(38,0){\line(0,1){49}}

 \put(52,0){\line(0,1){49}} \put(62,0){\line(0,1){49}}
\put(72,0){\line(0,1){49}}


\put(08.5,44.5){\makebox(0,0)[bl]{DA}}


\put(19,44.5){\makebox(0,0)[bl]{Improve-}}
\put(19,41){\makebox(0,0)[bl]{ment action}}

\put(39,44.5){\makebox(0,0)[bl]{Binary}}
\put(39,41){\makebox(0,0)[bl]{variable}}

\put(53,45){\makebox(0,0)[bl]{Cost}}
\put(55,40.5){\makebox(0,0)[bl]{\(a_{ij}\)}}

\put(62.6,45){\makebox(0,0)[bl]{Profit}}
\put(65,40.5){\makebox(0,0)[bl]{\(c_{ij}\)}}



\put(1.6,34){\makebox(0,0)[bl]{\(1\)}}

\put(9,33.5){\makebox(0,0)[bl]{\(X_{2}\) }}

\put(23,34){\makebox(0,0)[bl]{None}}

\put(43,34){\makebox(0,0)[bl]{\(x_{11}\)}}

\put(56,34){\makebox(0,0)[bl]{\(0\)}}
\put(66,34){\makebox(0,0)[bl]{\(0\)}}


\put(1.6,30){\makebox(0,0)[bl]{\(2\)}}

\put(9,29.5){\makebox(0,0)[bl]{\(X_{2}\) }}

\put(23,30){\makebox(0,0)[bl]{\(2 \Rightarrow 1\)}}

\put(43,30){\makebox(0,0)[bl]{\(x_{12}\)}}

\put(56,30){\makebox(0,0)[bl]{\(3\)}}
\put(66,30){\makebox(0,0)[bl]{\(4\)}}


\put(1.6,26){\makebox(0,0)[bl]{\(3\)}}

\put(9,25.5){\makebox(0,0)[bl]{\(Z_{3}\) }}

\put(23,26){\makebox(0,0)[bl]{None}}

\put(43,26){\makebox(0,0)[bl]{\(x_{21}\)}}

\put(56,26){\makebox(0,0)[bl]{\(0\)}}
\put(66,26){\makebox(0,0)[bl]{\(0\)}}


\put(1.6,22){\makebox(0,0)[bl]{\(4\)}}

\put(9,21.5){\makebox(0,0)[bl]{\(Z_{3}\) }}

\put(23,22){\makebox(0,0)[bl]{\(3 \Rightarrow 2\)}}

\put(43,22){\makebox(0,0)[bl]{\(x_{22}\)}}

\put(56,22){\makebox(0,0)[bl]{\(1\)}}
\put(66,22){\makebox(0,0)[bl]{\(2\)}}


\put(1.6,18){\makebox(0,0)[bl]{\(5\)}}

\put(9,17.5){\makebox(0,0)[bl]{\(Z_{3}\) }}

\put(23,18){\makebox(0,0)[bl]{\(3 \Rightarrow 1\)}}

\put(43,18){\makebox(0,0)[bl]{\(x_{23}\)}}

\put(56,18){\makebox(0,0)[bl]{\(5\)}}
\put(66,18){\makebox(0,0)[bl]{\(4\)}}


\put(1.6,14){\makebox(0,0)[bl]{\(6\)}}

\put(9,13.5){\makebox(0,0)[bl]{\(U_{2} \)}}

\put(23,14){\makebox(0,0)[bl]{None}}

\put(43,14){\makebox(0,0)[bl]{\(x_{31}\)}}

\put(56,14){\makebox(0,0)[bl]{\(0\)}}
\put(66,14){\makebox(0,0)[bl]{\(0\)}}


\put(1.6,10){\makebox(0,0)[bl]{\(7\)}}

\put(9,09.5){\makebox(0,0)[bl]{\(U_{2} \)}}

\put(23,10){\makebox(0,0)[bl]{\(2 \Rightarrow 1\)}}

\put(43,10){\makebox(0,0)[bl]{\(x_{32}\)}}

\put(56,10){\makebox(0,0)[bl]{\(4\)}}
\put(66,10){\makebox(0,0)[bl]{\(4\)}}


\put(1.6,6){\makebox(0,0)[bl]{\(8\)}}

\put(9,05.5){\makebox(0,0)[bl]{\(V_{2} \)}}

\put(23,06){\makebox(0,0)[bl]{None}}

\put(43,06){\makebox(0,0)[bl]{\(x_{41}\)}}

\put(56,06){\makebox(0,0)[bl]{\(0\)}}
\put(66,06){\makebox(0,0)[bl]{\(0\)}}


\put(1.6,2){\makebox(0,0)[bl]{\(9\)}}

\put(9,01.5){\makebox(0,0)[bl]{\(V_{2} \)}}

\put(23,02){\makebox(0,0)[bl]{\(2 \Rightarrow 1\)}}

\put(43,02){\makebox(0,0)[bl]{\(x_{42}\)}}

\put(56,02){\makebox(0,0)[bl]{\(2\)}}
\put(66,02){\makebox(0,0)[bl]{\(4\)}}

\end{picture}
\end{center}

 The multiple choice problem is:
 \[\max \sum_{i=1}^{4}  \sum_{j=1}^{q_{i}}   c_{ij} x_{ij}
 ~~~ s.t.~ \sum_{i=1}^{4}  \sum_{j=1}^{q_{i}}   a_{ij} x_{ij} \leq
 b,
%
 ~~ \sum_{j=1}^{q_{i}}   x_{ij} = 1 ~~  \forall i=\overline{1,4},
  ~~x_{ij} \in \{0,1\}.\]
%
%
 Clearly, \(q_{1} = 2\), \(q_{2} = 3\), \(q_{3} = 2\), \(q_{4} =
 2\).
 Some examples of the resultant improvement solutions  are
  (a simple greedy algorithm was used;
 the algorithm is based on ordering of elements by
  \(c_{i}/a_{i}\)):

  (1) \(b^{1}=1\):~
 (\(x_{11} = 1\), \(x_{22} = 1\), \(x_{31} = 1\), \(x_{41} = 1\)),
  ~ value of objective function
  ~\(c^{b_{1}} = 2\),
 ~\(N(S^{b_{1}}_{2}) = (3;2,3,0,0)\);

  (2) \(b^{2}=2\):~
 (\(x_{11} = 1\), \(x_{21} = 1\), \(x_{31} = 1\), \(x_{42} = 1\)),
%
  ~\(c^{b_{1}} = 4\),
 ~\(N(S^{b_{1}}_{2}) = (3;3,1,1,0)\);

%
%

  (3) \(b^{3}=7\):~
 (\(x_{11} = 1\), \(x_{23} = 1\), \(x_{31} = 1\), \(x_{42} = 1\)),
%
  ~\(c^{b_{3}} = 8\),
 ~\(N(S^{b_{3}}_{2}) = (3;3,2,0,0)\); and

  (4) \(b^{4}=14\):~
 (\(x_{12} = 1\), \(x_{23} = 1\), \(x_{32} = 1\), \(x_{42} = 1\)),
%
  ~\(c^{b_{4}} = 16\),
 ~\(N(S^{b_{4}}_{2}) = (3;5,0,0,0)\).

\subsection{Graph Vertex Recoloring Problem \cite{lev09}}


 Now, a simplified illustrative version of system improvement by components
 will be presented as graph vertex recoloring problem
 \cite{lev09}.
 In recent years,
 graph recoloring problems have been examined
 (e.g., \cite{bach06,bar08,hack08,moran07,moran08}).
 Here,
 the basic problem formulation is the following.
 There are the following:
 graph ~\( G=(A,E) \) (\(|A|=n\)),
 set of colors ~\(X = \{x_{1},...,x_{k} \}\),
 and
 initial {\it color configuration} for graph \(G\):
 ~\(C^{o}(G) = \{ C^{o}(a_{1}),...,C^{o}(a_{i}),...,C^{o}(a_{n}) \}\),
 ~\(C^{o}(a_{i})\in X\)
 ~\(\forall a_{i} \in A\)
 ~(\(i=\overline{1,n} \)).
 Let
 ~\( d_{a_{i}}( x_{\delta_{1}},x_{\delta_{2}}) \)
 (\(\delta_{1} =\overline{1,k}\),
  ~\(\delta_{1} =\overline{1,k}\))~
 be a nonnegative recoloring cost matrix for each vertex
 (i.e., individual recoloring cost matrix ~\(\forall a_{i} \in A\)).
 For graph \(G\),
 a goal {\it color configuration} ~\( C^{g}(G) \)
 or a set of goal {\it color configurations} ~\( \{ C^{g}(G) \} \)
 are used as well.
 Now, for each two {\it color combination}
 ~(\(C^{1}(G)\) and \(C^{2}(G)\))
 it is possible to consider an
 integrated cost (the cost of recoloring),
 e.g., as an additive function
 ~\( D ( C^{1}(G),C^{2}(G) )  = \sum_{i=1}^{n}  d_{a_{i}} ( C^{1}(a_{i}),C^{2}(a_{i}) )  \).
 In addition,
 it is necessary to consider
 a proximity of two {\it color configurations}:
 ~\( \rho ( C^{\beta_{1}} (G), C^{\beta_{2}} (G) ) \).
 Thus,
 the following transformation chain is examined:
 \[ C^{o}(G=(A,E)) \Rightarrow   C^{*}(G=(A,E)) \Rightarrow C^{g}(G=(A,E)) \]
 where ~\(C^{*}(G=(A,E))\) is a resultant
 {\it color configuration}.

 Generally, it is reasonable to examine two generalized objectives:
 ~(i) the cost of transformation
 ~\(C^{o}(G) \Rightarrow   C^{*}(G) \) and
 ~(ii) the quality of proximity between
 ~\( C^{*}(G)\) and  \( C^{g}(G) \) (or \( \{ C^{g}(G) \}  \)).
 Assessment of the cost and proximity above can be based on
 various approaches
 (e.g., additive function, minimization, 'min/max' function,
  vector function).

 Now, the basic problem of graph recoloring can be considered as follows:

~~

 {\it Find the new color configuration}
 ~\(C^{*}(G)\)~ for a given graph ~\(G=(A,E)\)~
%
  {\it to} {\it minimize}
  {\it the proximity of the resultant graph coloring
  configuration}
  ~ \(C^{*}(G(A,E))\)~
  {\it to} {\it the}  {\it goal coloring configuration of graph}
  ~\(C^{g}(G(A,E))\)~
 {\it while taking into account the limited integrated cost of the
 recoloring}
 ~(\( \leq  \overline{D}\)):
 \[  \min_{ \{C(G=(A,E)) \} }   \rho ( C^{*}(G), C^{g}(G) ) )\]
  \[ s.t.  ~~ D( C^{o}(G),C^{*}(G) ) \leq  \overline{D},
  ~ C^{*}(a_{i}) \neq  C^{*}(a_{j}) ~ \forall (a_{i},a_{j}) \in E, ~i \neq j.\]
 Fig. 18 illustrates the graph (vertex) recoloring problem:
 \[
 C^{o}(G) = (P_{2}\star Q_{3} \star U_{3} \star V_{2} \star W_{1})
 ~ \Rightarrow ~
  C^{*}(G) = (P_{1}\star Q_{2} \star U_{2} \star V_{1} \star
  W_{3}) ~.\]


\begin{center}
\begin{picture}(119,55)
\put(22,00){\makebox(0,0)[bl]{Fig. 18. Example of vertex
recoloring \cite{lev09}}}
%
\put(05,30){\circle*{1.5}}
\put(20,45){\circle*{1.5}}
\put(35,15){\circle*{1.5}}
\put(50,30){\circle*{1.5}}
\put(27.5,30){\circle*{1.5}}
\put(05,30){\line(1,0){45}} \put(05,30){\line(1,1){15}}
\put(20,45){\line(1,-2){15}} \put(35,15){\line(1,1){15}}
\put(20,45){\line(2,-1){30}}
\put(03,32){\makebox(0,0)[bl]{\(p\)}}

\put(02,25){\makebox(0,0)[bl]{\(P_{1}\)}}
\put(02,21){\makebox(0,0)[bl]{\(P_{2}\)}}
\put(02,17){\makebox(0,0)[bl]{\(P_{3}\)}}
\put(04.5,22.5){\oval(06,4)}
%
\put(50,25){\makebox(0,0)[bl]{\(V_{1}\)}}
\put(50,21){\makebox(0,0)[bl]{\(V_{2}\)}}
\put(50,17){\makebox(0,0)[bl]{\(V_{3}\)}}

\put(50,32){\makebox(0,0)[bl]{\(v\)}}
\put(52.5,22.5){\oval(06,4)}
%
\put(38,14){\makebox(0,0)[bl]{\(U_{1}\)}}
\put(38,10){\makebox(0,0)[bl]{\(U_{2}\)}}
\put(38,6){\makebox(0,0)[bl]{\(U_{3}\)}}

\put(34,18){\makebox(0,0)[bl]{\(u\)}}
\put(40.5,7.5){\oval(06,4)}
%
\put(27,50){\makebox(0,0)[bl]{\(Q_{1}\)}}
\put(27,46){\makebox(0,0)[bl]{\(Q_{2}\)}}
\put(27,42){\makebox(0,0)[bl]{\(Q_{3}\)}}

\put(16,45){\makebox(0,0)[bl]{\(q\)}}
\put(29.5,43.5){\oval(06,4)}
%
\put(22,25){\makebox(0,0)[bl]{\(W_{1}\)}}
\put(22,21){\makebox(0,0)[bl]{\(W_{2}\)}}
\put(22,17){\makebox(0,0)[bl]{\(W_{3}\)}}

\put(28,32){\makebox(0,0)[bl]{\(w\)}}
\put(24.5,26.5){\oval(06,4)}

\put(11,12){\makebox(0,0)[bl]{Initial}}
\put(11,8){\makebox(0,0)[bl]{coloring}}



\put(57,44){\makebox(0,0)[bl]{\(\Rightarrow \)}}

\put(57,40){\makebox(0,0)[bl]{\(\Rightarrow \)}}

\put(57,36){\makebox(0,0)[bl]{\(\Rightarrow \)}}

\put(65,30){\circle*{1.5}}
\put(80,45){\circle*{1.5}}
\put(95,15){\circle*{1.5}}
\put(110,30){\circle*{1.5}}
\put(87.5,30){\circle*{1.5}}
\put(65,30){\line(1,0){45}} \put(65,30){\line(1,1){15}}
\put(80,45){\line(1,-2){15}} \put(95,15){\line(1,1){15}}
\put(80,45){\line(2,-1){30}}
\put(63,32){\makebox(0,0)[bl]{\(p\)}}

\put(62,25){\makebox(0,0)[bl]{\(P_{1}\)}}
\put(62,21){\makebox(0,0)[bl]{\(P_{2}\)}}
\put(62,17){\makebox(0,0)[bl]{\(P_{3}\)}}
\put(64.5,26.5){\oval(06,4)}
%
\put(110,25){\makebox(0,0)[bl]{\(V_{1}\)}}
\put(110,21){\makebox(0,0)[bl]{\(V_{2}\)}}
\put(110,17){\makebox(0,0)[bl]{\(V_{3}\)}}

\put(110,32){\makebox(0,0)[bl]{\(v\)}}
\put(112.5,26.5){\oval(06,4)}
%
\put(98,14){\makebox(0,0)[bl]{\(U_{1}\)}}
\put(98,10){\makebox(0,0)[bl]{\(U_{2}\)}}
\put(98,6){\makebox(0,0)[bl]{\(U_{3}\)}}

\put(94,18){\makebox(0,0)[bl]{\(u\)}}
\put(100.5,11.5){\oval(06,4)}
%
\put(87,50){\makebox(0,0)[bl]{\(Q_{1}\)}}
\put(87,46){\makebox(0,0)[bl]{\(Q_{2}\)}}
\put(87,42){\makebox(0,0)[bl]{\(Q_{3}\)}}

\put(76,45){\makebox(0,0)[bl]{\(q\)}}
\put(89.5,47.5){\oval(06,4)}
%
\put(82,25){\makebox(0,0)[bl]{\(W_{1}\)}}
\put(82,21){\makebox(0,0)[bl]{\(W_{2}\)}}
\put(82,17){\makebox(0,0)[bl]{\(W_{3}\)}}

\put(88,32){\makebox(0,0)[bl]{\(w\)}}
\put(84.5,18.5){\oval(06,4)}
%
%

\put(71,12){\makebox(0,0)[bl]{Resultant}}
\put(71,8){\makebox(0,0)[bl]{coloring}}

\end{picture}
\end{center}

\section{Modification of System Structure}

 Modification of system structures
 is a crucial and complex combinatorial problem:

~~

 {\it Initial system structure} \(\Longrightarrow\)
 {\it Modification process}
 \(\Longrightarrow\)
 {\it Final system structure}

~~

 Table 7 contains a list if basic problems, which are targeted to
 modification of structures.

\begin{center}
\begin{picture}(145,78)

\put(43,74){\makebox(0,0)[bl]{Table 7. Modification of
 structures}}

\put(00,0){\line(1,0){145}} \put(00,66){\line(1,0){145}}
\put(00,72){\line(1,0){145}}

\put(00,0){\line(0,1){72}} \put(49,0){\line(0,1){72}}
\put(80,0){\line(0,1){72}} \put(128,0){\line(0,1){72}}
\put(145,0){\line(0,1){72}}

\put(01,68){\makebox(0,0)[bl]{Problem}}

\put(50,68){\makebox(0,0)[bl]{Initial structure}}

\put(81,68){\makebox(0,0)[bl]{Resultant structure}}

\put(129,68){\makebox(0,0)[bl]{Source}}


\put(01,61){\makebox(0,0)[bl]{1.Modification of tree-like}}
\put(03,57.5){\makebox(0,0)[bl]{structures:}}


\put(02,53){\makebox(0,0)[bl]{1.1.Hotlink assignment}}
\put(04,49){\makebox(0,0)[bl]{problems}}

\put(50,53){\makebox(0,0)[bl]{Tree-like structure }}

\put(81,53){\makebox(0,0)[bl]{Tree-like structure with }}
\put(81,49){\makebox(0,0)[bl]{additional link from root}}

\put(129,53){\makebox(0,0)[bl]{\cite{bose01,dou08,fuhr01} }}

\put(129,49){\makebox(0,0)[bl]{\cite{jacobs12,kra01,lev12hier}}}


\put(02,45){\makebox(0,0)[bl]{1.2.Modification of tree via}}
\put(04,41){\makebox(0,0)[bl]{condensing weighted vertices }}

\put(50,45){\makebox(0,0)[bl]{Weighted tree}}

\put(81,45){\makebox(0,0)[bl]{Weighted tree with}}
\put(81,41){\makebox(0,0)[bl]{aggregated vertices }}

\put(129,45){\makebox(0,0)[bl]{\cite{lev81,lev98,lev12hier}}}


\put(02,37){\makebox(0,0)[bl]{1.3.Transformation of tree}}
\put(04,33){\makebox(0,0)[bl]{into Steiner tree}}

\put(50,37){\makebox(0,0)[bl]{Tree}}

\put(81,37){\makebox(0,0)[bl]{Steiner tree}}

\put(129,37){\makebox(0,0)[bl]{\cite{lev12hier,levnur11,levzam11}}}


\put(01,28){\makebox(0,0)[bl]{2.Transformation of morpho-}}
\put(03,24){\makebox(0,0)[bl]{logical structure}}

\put(50,28){\makebox(0,0)[bl]{Morphological}}
\put(50,24.5){\makebox(0,0)[bl]{structure}}

\put(81,28){\makebox(0,0)[bl]{Morphological structure}}
\put(81,24){\makebox(0,0)[bl]{with required properties}}

\put(129,28){\makebox(0,0)[bl]{\cite{lev98,lev06,lev12hier}}}


\put(01,19){\makebox(0,0)[bl]{3.Transformation of layered}}
\put(03,15){\makebox(0,0)[bl]{structure}}

\put(50,19){\makebox(0,0)[bl]{Layered structure}}

\put(81,19){\makebox(0,0)[bl]{Layered structure with}}
\put(81,15){\makebox(0,0)[bl]{required properties}}



\put(01,10){\makebox(0,0)[bl]{4.Augmenation problem}}

\put(50,10){\makebox(0,0)[bl]{Connected graph}}

\put(81,10){\makebox(0,0)[bl]{Connected graph with}}
\put(81,06){\makebox(0,0)[bl]{required properties}}


\put(81,02){\makebox(0,0)[bl]{(e.g., increased connectivity)}}

\put(129,10){\makebox(0,0)[bl]{\cite{esw76,khu97}}}

\end{picture}
\end{center}

\subsection{Modification of Tree-like Structures}

\subsubsection{Hotlink Assignment Problems \cite{lev12hier}}

 In  general, ``hotlink assignment problem''
 is a network upgrade problem  (e.g., \cite{fuhr01}):

~~

 {\it Find additional new arc(s) to the initial graph
 in order to insert shortcuts and decrease the expected path
 length.}

~~

 Mainly, the problem is examined for trees.
 Let \(T = (A,E)\) be a directed tree with maximum degree \(d\),
 rooted at a node \(r_{0}\in A\)
 (elements of \(A\) correspond to Web sites, elements of \(E\) correspond to hyperlinks).
 A node weight equals its access (search) frequency (probability).
 It is assumed that required information is contained at
 leaf nodes (for simplicity).
 The length of the search for node \(v \in A\) equals
 the number of links in the path from \(r_{0}\) to \(v\).

 Let \(T_{u} = (A_{u},E_{u})\) be a subtree of \(T\)
 (\(A_{u} \subseteq A,  E_{u} \subseteq E \)), rooted at node \(u\in A\)
 (here \(u\) is not the son of \(r_{0}\)).
 Thus, additional direct link (``hotlink'') will be as follows:
 (\(r_{0},u\)).
 In this case, a path to all leaf nodes in \(T_{u}\) will be smaller.

 Fig. 19
 illustrates the simplest version of ``hotlink assignment'' problem.

\begin{center}
\begin{picture}(52,36)

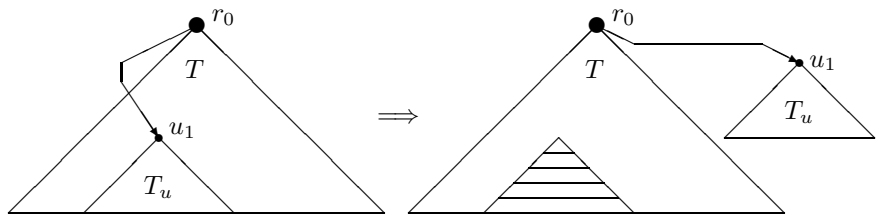
\put(17,00){\makebox(0,0)[bl]{Fig. 19. Hotlink assignment problem
(one hotlink) \cite{lev12hier}}}


\put(25,31){\circle*{2}}

\put(27,31){\makebox(0,0)[bl]{\(r_{0}\)}}

\put(23.5,24){\makebox(0,0)[bl]{\(T\)}}
\put(18,08){\makebox(0,0)[bl]{\(T_{u}\)}}

\put(00,06){\line(1,1){25}} \put(50,06){\line(-1,1){25}}

\put(00,06){\line(1,0){50}}


\put(25,31){\line(-2,-1){10}} \put(15,26){\line(0,-1){2.5}}
\put(15,23.5){\vector(2,-3){5}}


\put(20,16){\circle*{1.2}}

\put(21.3,16){\makebox(0,0)[bl]{\(u_{1}\)}}

\put(10,06){\line(1,1){10}} \put(30,06){\line(-1,1){10}}

\put(10,06){\line(1,0){20}}


\put(49,18){\makebox(0,0)[bl]{\(\Longrightarrow\)}}

\end{picture}
%
\begin{picture}(62,36)


\put(25,31){\circle*{2}}

\put(27,31){\makebox(0,0)[bl]{\(r_{0}\)}}

\put(00,06){\line(1,1){25}} \put(50,06){\line(-1,1){25}}

\put(00,06){\line(1,0){50}}

\put(23.5,24){\makebox(0,0)[bl]{\(T\)}}

\put(50,18){\makebox(0,0)[bl]{\(T_{u}\)}}


\put(10,06){\line(1,1){10}} \put(30,06){\line(-1,1){10}}


\put(10,06){\line(1,0){20}}


\put(18,14){\line(1,0){04}} \put(16,12){\line(1,0){08}}
\put(14,10){\line(1,0){12}} \put(12,08){\line(1,0){16}}


\put(25,31){\line(2,-1){5}} \put(30,28.5){\line(1,0){17}}
\put(47,28.5){\vector(2,-1){5}}


\put(52,26){\circle*{1.2}}

\put(53.3,25){\makebox(0,0)[bl]{\(u_{1}\)}}

\put(42,16){\line(1,1){10}} \put(62,16){\line(-1,1){10}}

\put(42,16){\line(1,0){20}}


\end{picture}
\end{center}

 Some versions of the  problem are presented in Table 8.
 Mainly, ``hotlink assignment'' problems belong to class of
 NP-hard problems
 (e.g.,
 \cite{jacobs12}).
 Many approximation algorithms have been suggested
 for the problems (including FPTAS)
(e.g.,
 \cite{kra01,mat04}).

\begin{center}
\begin{picture}(71,36)

\put(05.5,32){\makebox(0,0)[bl]{Table 8. Hotlink assignment
problems}}

\put(00,0){\line(1,0){71}} \put(00,23){\line(1,0){71}}
\put(00,30){\line(1,0){71}}

\put(00,0){\line(0,1){30}} \put(55,0){\line(0,1){30}}
\put(71,0){\line(0,1){30}}

\put(01,25){\makebox(0,0)[bl]{Problem}}
\put(56,25){\makebox(0,0)[bl]{Source}}


\put(01,18){\makebox(0,0)[bl]{1.Basic hotlink assignment}}
\put(56,18){\makebox(0,0)[bl]{\cite{bose01,perk99} }}


\put(01,14){\makebox(0,0)[bl]{2.Single  hotlink assignment}}
\put(56,14){\makebox(0,0)[bl]{\cite{dou10,kra01}}}


\put(01,10){\makebox(0,0)[bl]{3.Hotlinks only for leafs}}
\put(56,10){\makebox(0,0)[bl]{\cite{jacobs12}}}


\put(01,6){\makebox(0,0)[bl]{4.Multiple hotlink assignment}}
\put(56,6){\makebox(0,0)[bl]{\cite{dou10,fuhr01}}}

\put(01,02){\makebox(0,0)[bl]{5.Dynamic hotlink assignment}}
\put(56,02){\makebox(0,0)[bl]{\cite{dou08}}}

\end{picture}
\end{center}

\subsubsection{Modification of Tree via Condensing of Weighted
 Edges \cite{lev81,lev98,lev12hier}}

 This section
  describes briefly transformation of a tree
 (with weights of vertices and weights of edge/arcs)
 via integration (condensing)
 of some neighbor vertices
 while taking into account
 a constraint for a total weight of the maximum tree tail
 (i.e., length from root to a leaf vertex).
 The problem was firstly formulated
  for designing an overlay structure of
  a  modular software system in \cite{lev81}.
 The integration of software modules requires additional memory, but
 allows to decrease a time (i.e., frequency) of loading some
 corresponding modules.
 Other applications of the problem can be examined as well, e.g.,
  hierarchical structure of data,
  call problem,
  hierarchical information structure of Web-sites.
  This problem  is illustrated in
 Fig. 20 and Fig. 21 by an example for designing the over-lay structure
 on the basis of module integration, when different software or data modules
 can apply the same parts of RAM.
 A  new kind of FPTAS for the above-mentioned combinatorial optimization
 problem (a generalization of multiple choice problem over a
 tree-like structure and special constraints)
 was suggested in \cite{lev81}.

\begin{center}
\begin{picture}(95,41)

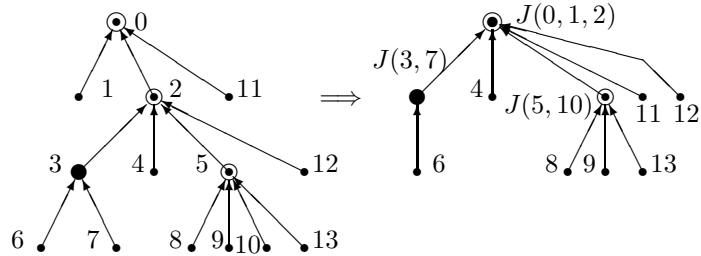
\put(00,00){\makebox(0,0)[bl] {Fig. 20. Integration of software
 modules (over-lay structure)
 \cite{lev12hier}
 }}

\put(5,07){\circle*{1}} \put(1,07){\makebox(0,0)[bl]{\(6\)}}

\put(15,07){\circle*{1}} \put(11,07){\makebox(0,0)[bl]{\(7\)}}

\put(05,07){\vector(1,2){4.5}} \put(15,07){\vector(-1,2){4.5}}

\put(25,07){\circle*{1}} \put(22,07){\makebox(0,0)[bl]{\(8\)}}

\put(35,07){\circle*{1}}
\put(30.7,06.3){\makebox(0,0)[bl]{\(10\)}}

\put(40,07){\circle*{1}} \put(41,07){\makebox(0,0)[bl]{\(13\)}}
\put(40,07){\vector(-1,1){9.5}}

\put(25,07){\vector(1,2){4.5}} \put(35,07){\vector(-1,2){4.5}}

\put(30,07){\circle*{1}} \put(27.6,07){\makebox(0,0)[bl]{\(9\)}}
\put(30,07){\vector(0,1){09}}

\put(10,17){\circle*{2}} \put(6,17){\makebox(0,0)[bl]{\(3\)}}
\put(20,17){\circle*{1}} \put(17,17){\makebox(0,0)[bl]{\(4\)}}
\put(30,17){\circle*{1}} \put(30,17){\circle{2}}
\put(25.5,17){\makebox(0,0)[bl]{\(5\)}}

\put(40,17){\circle*{1}} \put(41,17){\makebox(0,0)[bl]{\(12\)}}
\put(40,17){\vector(-2,1){19}}

\put(10,17){\vector(1,1){09}} \put(30,17){\vector(-1,1){09}}
\put(20,17){\vector(0,1){09}}

\put(20,27){\circle*{1}} \put(20,27){\circle{2.1}}
\put(22,27){\makebox(0,0)[bl]{\(2\)}}

\put(30,27){\circle*{1}} \put(31,27){\makebox(0,0)[bl]{\(11\)}}
\put(30,27){\vector(-3,2){14}}

\put(10,27){\circle*{1}} \put(13,27){\makebox(0,0)[bl]{\(1\)}}
\put(10,27){\vector(1,2){4.5}} \put(20,27){\vector(-1,2){4.5}}

\put(15,37){\circle*{1.2}} \put(15,37){\circle{2.3}}
\put(17.4,36){\makebox(0,0)[bl]{\(0\)}}


\put(65,37){\circle*{1.3}} \put(65,37){\circle{2.5}}
\put(68,36){\makebox(0,0)[bl]{\(J(0,1,2)\)}}

\put(55,27){\circle*{2}}
\put(49,30){\makebox(0,0)[bl]{\(J(3,7)\)}}

\put(65,27){\circle*{1}} \put(62,27){\makebox(0,0)[bl]{\(4\)}}

\put(80,27){\circle*{1}} \put(80,27){\circle{2}}
\put(80,27){\vector(-3,2){14}}
\put(66.5,24){\makebox(0,0)[bl]{\(J(5,10)\)}}

\put(85,27){\circle*{1}} \put(84,23.7){\makebox(0,0)[bl]{\(11\)}}
\put(85,27){\vector(-2,1){19}}

\put(90,27){\circle*{1}} \put(89,23.7){\makebox(0,0)[bl]{\(12\)}}
\put(90,27){\line(-1,1){5}} \put(85,32){\vector(-4,1){19}}

\put(55,27){\vector(1,1){09}} \put(65,27){\vector(0,1){09}}

\put(55,17){\circle*{1}} \put(57,17){\makebox(0,0)[bl]{\(6\)}}
\put(55,17){\vector(0,1){09}}

\put(75,17){\circle*{1}} \put(72,17){\makebox(0,0)[bl]{\(8\)}}
\put(75,17){\vector(1,2){4.5}}

\put(80,17){\circle*{1}} \put(77,17){\makebox(0,0)[bl]{\(9\)}}
\put(80,17){\vector(0,1){9}}

\put(85,17){\circle*{1}} \put(86,17){\makebox(0,0)[bl]{\(13\)}}
\put(85,17){\vector(-1,2){4.5}}


\put(42,26){\makebox(0,0)[bl]{\(\Longrightarrow\)}}

\end{picture}
\end{center}

\begin{center}
\begin{picture}(97,63)
\put(16,00){\makebox(0,0)[bl] {Fig. 21. Usage of memory (RAM)
 \cite{lev12hier}
}}


\put(58,07){\line(1,0){29}}

\put(085,07){\line(0,1){10}}
\put(086,12){\makebox(0,0)[bl]{\(9\)}}

\put(92,17){\line(0,-1){5}} \put(93,12){\makebox(0,0)[bl]{\(13\)}}
\put(085,17){\line(1,0){07}} \put(91,12){\line(1,0){02}}

\put(075,12){\line(0,1){5}} \put(077,12){\makebox(0,0)[bl]{\(8\)}}
\put(074,12){\line(1,0){2}}

\put(075,17){\line(1,0){10}}

\put(080,17){\line(0,1){3}}
\put(081,18){\makebox(0,0)[bl]{\(10\)}}
\put(079,20){\line(1,0){2}}

\put(080,20){\line(0,1){17}}
\put(081,32){\makebox(0,0)[bl]{\(5\)}}

\put(86,37){\line(0,-1){4}} \put(87,32){\makebox(0,0)[bl]{\(11\)}}
\put(80,37){\line(1,0){06}} \put(85,33){\line(1,0){02}}

\put(93,37){\line(0,-1){6}} \put(94,32){\makebox(0,0)[bl]{\(12\)}}
\put(86,37){\line(1,0){07}} \put(92,31){\line(1,0){02}}

\put(75,30){\line(0,1){7}} \put(76,32){\makebox(0,0)[bl]{\(4\)}}
\put(74,30){\line(1,0){2}}

\put(70,37){\line(1,0){10}}

\put(070,25){\line(0,1){12}}
\put(067,32){\makebox(0,0)[bl]{\(3\)}} \put(069,25){\line(1,0){2}}

\put(070,19){\line(0,1){6}} \put(071,21){\makebox(0,0)[bl]{\(7\)}}
\put(069,19){\line(1,0){2}}

\put(070,11){\line(0,1){8}} \put(071,13){\makebox(0,0)[bl]{\(6\)}}
\put(069,11){\line(1,0){2}}

\put(075,37){\line(0,1){7}} \put(076,40){\makebox(0,0)[bl]{\(2\)}}
\put(074,37){\line(1,0){2}}

\put(075,44){\line(0,1){11}}
\put(076,48){\makebox(0,0)[bl]{\(1\)}} \put(074,44){\line(1,0){2}}

\put(075,55){\line(0,1){6}} \put(076,57){\makebox(0,0)[bl]{\(0\)}}
\put(074,55){\line(1,0){2}}

\put(58,61){\line(1,0){19}}

\put(60,12){\vector(0,-1){5}} \put(60,12){\vector(0,1){49}}
\put(61,39){\makebox(0,0)[bl]{\(b(G^{'})\)}}


\put(50,39){\makebox(0,0)[bl]{\(\Longrightarrow\)}}


\put(0,61){\line(1,0){17}} \put(0,21){\line(1,0){36}}

\put(2,32){\vector(0,-1){11}} \put(2,32){\vector(0,1){29}}

\put(3,39){\makebox(0,0)[bl]{\(b(G)\)}}

\put(015,55){\line(0,1){6}} \put(16,57){\makebox(0,0)[bl]{\(0\)}}
\put(010,55){\line(1,0){15}}

\put(010,44){\line(0,1){11}}
\put(011,50){\makebox(0,0)[bl]{\(1\)}} \put(09,44){\line(1,0){2}}

\put(025,55){\line(0,-1){7}}
\put(026,50){\makebox(0,0)[bl]{\(2\)}}
\put(015,48){\line(1,0){20}}

\put(035,55){\line(0,-1){4}}
\put(036,50){\makebox(0,0)[bl]{\(11\)}}
\put(025,55){\line(1,0){10}} \put(034,51){\line(1,0){02}}

\put(035,48){\line(0,-1){17}}
\put(032.4,43){\makebox(0,0)[bl]{\(5\)}}
\put(030,31){\line(1,0){10}}

\put(042,48){\line(0,-1){6}}
\put(043,43){\makebox(0,0)[bl]{\(12\)}}
\put(035,48){\line(1,0){07}} \put(041,42){\line(1,0){02}}

\put(025,48){\line(0,-1){7}}
\put(026,43){\makebox(0,0)[bl]{\(4\)}} \put(024,41){\line(1,0){2}}

\put(015,48){\line(0,-1){12}}
\put(016,43){\makebox(0,0)[bl]{\(3\)}}
\put(010,36){\line(1,0){10}}

\put(020,36){\line(0,-1){6}}
\put(021,31){\makebox(0,0)[bl]{\(7\)}} \put(019,30){\line(1,0){2}}

\put(010,36){\line(0,-1){8}}
\put(011,31){\makebox(0,0)[bl]{\(6\)}} \put(09,28){\line(1,0){2}}

\put(035,31){\line(0,-1){10}}
\put(036,28){\makebox(0,0)[bl]{\(9\)}} \put(034,21){\line(1,0){2}}

\put(030,31){\line(0,-1){5}}
\put(031,28){\makebox(0,0)[bl]{\(8\)}} \put(029,26){\line(1,0){2}}

\put(040,31){\line(0,-1){3}}
\put(041,28){\makebox(0,0)[bl]{\(10\)}}
\put(039,28){\line(1,0){2}}

\put(047,31){\line(0,-1){5}}
\put(048,28){\makebox(0,0)[bl]{\(13\)}}
\put(040,31){\line(1,0){07}} \put(046,26){\line(1,0){02}}

\end{picture}
\end{center}

\subsubsection{Transformation of Tree into Steiner Tree
 \cite{lev12hier}}

 Recently, two multicriteria problems for transformation
 of a tree into Steiner tree have been examined:
 (1) multicriteria problem for transformation of an initial
 tree into Steiner tree \cite{levnur11} and
  (2) multicriteria problem for transformation of an initial
 tree into Steiner tree  while taking into account  a cost of Steiner vertices
 \cite{levzam11}.
 Here, the transformation of a tree
 \(T = (A,E)\)
  into Steiner tree
 \(S = (A',E')\)
  is considered
 as addition of Steiner points into an initial tree
 (or a preliminary built spanning tree)
 while taking into account the following:
 ``cost'' (required resource) of each Steiner point,
 generalized ``profit'' of each Steiner point,
 total resource constraint (i.e., total ``cost'' of the selected Steiner
 points).
 The material is based on  \cite{lev12hier}.
 A simplest case is considered when Steiner points for triangles are only
 examined.
 Evidently, vector-like
``cost'' and ``profit'' can be used as well.
 The solving scheme is the following:

~~

 {\it Stage 1.} Identification
 (e.g., expert judgment, clustering)
  of \(m\) regions
 (clusters, groups of neighbor nodes)
 in the initial tree \(T\)
 for possible addition of Steiner points.

 {\it Stage 2.} Generation of possible Steiner points (candidates)
 and their attributes (i.e., cost of addition, ``profit'').

 {\it Stage 3.} Formulation of multiple choice problem for
 selection of the best additional Steiner points
 while taking into account resource constraint(s):
 \[ \max~ \sum_{i=1}^{m}  \sum_{i=1}^{q_{i}}    ~ c_{ij} x_{ij}
  ~~~ s.t. ~~ \sum_{i=1}^{m} \sum_{j=1}^{q_{i}}  a_{ij} x_{ij} \leq b,
 ~~ \sum_{j=1}^{q_{i}} x_{ij} =  1,
 ~~ x_{ij} \in \{0, 1\};\]
 where
  \(i\) is the index of region (\(i = \overline{1,m}\)),
 \(q_{i}\) is the number of versions for addition of Steiner
 points in region \(i=\overline{1,m}\),
  \(j\) is the index of version for addition of Steiner points in region
 (\( j = \overline{1,q_{i}} \)  ),
 \(x_{ij}\) is binary variable that equals \(1\) if
 version \(j\) in region \(i\) is selected,
 \(b\) is a total constraint for the required resources
 (i.e., a total ``cost'').

 {\it Stage 4.} Solving the  multiple choice problem
 to obtain the resultant Steiner tree \(S\).

~

 A numerical illustrative example illustrates the scheme.
 Initial tree is (Fig. 22):
 \(T = (A,E)\), \(A = \{ 1,2,3,4,5,6,7,8,9,10,11 \}\).
 Four regions are defined (Fig. 22):
 region \(1\):  \(\{ 1,2,3,4 \}\);
 region \(2\):  \(\{ 4,6,7 \}\);
 region \(3\):  \(\{ 4,5,6,9,11 \}\);
 and
 region \(4\):  \(\{ 7,8,10 \}\).
 The considered Steiner points are the following (Fig. 22):
 region \(1\):  \(s_{11},s_{12}\);
 region \(2\):  \(s_{21} \);
 region \(3\):  \(s_{31},s_{32}\);
 and
 region \(4\):  \(s_{41}\).
 Table 9 contains initial data  for multiple choice problem:
 binary variables and corresponding attributes
 (required resource as ``cost'', possible ``profit'').
 The corresponding multiple choice is:
 \[ \max~ \sum_{i=1}^{4}  \sum_{i=1}^{q_{i}}    ~ c_{ij} x_{ij}
 ~~~~ s.t. ~~ \sum_{i=1}^{4} \sum_{j=1}^{q_{i}}
 a_{ij} x_{ij} \leq b,
 ~~ \sum_{j=1}^{q_{i}} x_{ij} =  1,
 ~~ x_{ij} \in \{0, 1\}.\]
 An obtained solution (i.e., as set of additional Steiner
 points) is the following (a simple greedy heuristic was used)
 (Fig. 22):~~
  \(b_{1} = 2.9\);~
 \( \overline{x}_{b_{1}}\):~ \( x_{12}=1\), \(x_{21}=1\), \(x_{32}=1\), \(x_{41}=1\),
 Steiner points \(Z_{b_{1}} =  \{ s_{11},s_{31} \} \),
 total (additive) ``profit'' \( \overline{c} = 5.5 \).

\begin{center}
\begin{picture}(40,53)

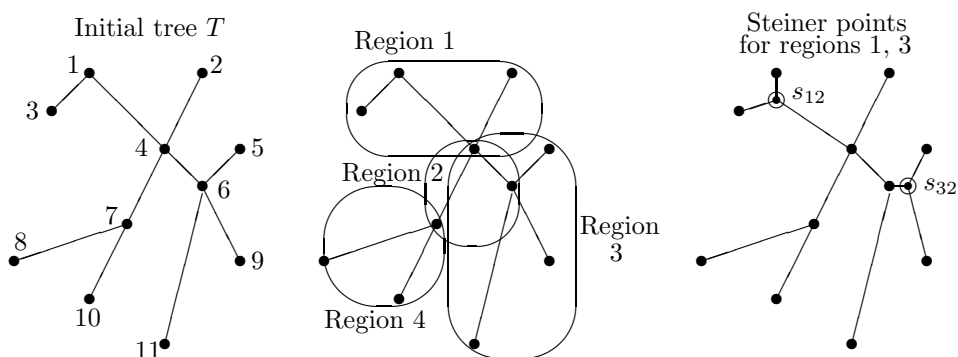
\put(022,00){\makebox(0,0)[bl]{Fig. 22. Initial tree,
 regions (clusters), Steiner tree}}

\put(08,48){\makebox(0,0)[bl]{Initial tree \(T\)}}

\put(01.5,37){\makebox(0,0)[bl]{\(3\)}} \put(05,38){\circle*{1.3}}

\put(05,38){\line(1,1){5}}



\put(10,43){\line(1,-1){10}}

\put(07,43){\makebox(0,0)[bl]{\(1\)}} \put(10,43){\circle*{1.3}}

\put(26,43){\makebox(0,0)[bl]{\(2\)}} \put(25,43){\circle*{1.3}}

\put(25,43){\line(-1,-2){5}}


\put(16,32){\makebox(0,0)[bl]{\(4\)}} \put(20,33){\circle*{1.3}}

\put(20,33){\line(-1,-2){5}}

\put(20,33){\line(1,-1){5}}

\put(31.5,17){\makebox(0,0)[bl]{\(9\)}} \put(30,18){\circle*{1.3}}

\put(31.5,32){\makebox(0,0)[bl]{\(5\)}} \put(30,33){\circle*{1.3}}

\put(30,33){\line(-1,-1){5}}

\put(25,28){\line(1,-2){5}}

\put(27,26.5){\makebox(0,0)[bl]{\(6\)}} \put(25,28){\circle*{1.3}}



\put(012,23){\makebox(0,0)[bl]{\(7\)}} \put(15,23){\circle*{1.3}}

\put(15,23){\line(-3,-1){15}}

\put(15,23){\line(-1,-2){5}}



\put(00,19.5){\makebox(0,0)[bl]{\(8\)}} \put(00,18){\circle*{1.3}}



\put(08,9.4){\makebox(0,0)[bl]{\(10\)}} \put(10,13){\circle*{1.3}}

\put(20,7){\line(1,4){5}}

\put(016,5){\makebox(0,0)[bl]{\(11\)}} \put(20,7){\circle*{1.3}}

\end{picture}
\begin{picture}(49,53)

\put(04,45.6){\makebox(0,0)[bl]{Region \(1\)}}
\put(16,38.3){\oval(26,12.6)}

\put(05,38){\circle*{1.3}}

\put(05,38){\line(1,1){5}}



\put(10,43){\line(1,-1){10}}

\put(10,43){\circle*{1.3}}

\put(25,43){\circle*{1.3}}

\put(25,43){\line(-1,-2){5}}


\put(02.4,28){\makebox(0,0)[bl]{Region \(2\)}}
\put(19.7,27){\oval(12.5,14)}

\put(20,33){\circle*{1.3}}

\put(20,33){\line(-1,-2){5}}

\put(20,33){\line(1,-1){5}}

\put(034,21){\makebox(0,0)[bl]{Region }}
\put(038,18){\makebox(0,0)[bl]{\(3\)}}
\put(25,20){\oval(17,30)}

\put(30,18){\circle*{1.3}}

\put(30,33){\circle*{1.3}}

\put(30,33){\line(-1,-1){5}}

\put(25,28){\line(1,-2){5}}

\put(25,28){\circle*{1.3}}



\put(15,23){\circle*{1.3}}

\put(15,23){\line(-3,-1){15}}

\put(15,23){\line(-1,-2){5}}



\put(00,18){\circle*{1.3}}



\put(00,8.5){\makebox(0,0)[bl]{Region \(4\)}}
\put(08,20){\oval(16,16)}

\put(10,13){\circle*{1.3}}

\put(20,7){\line(1,4){5}}

 \put(20,7){\circle*{1.3}}

\end{picture}
%
\begin{picture}(30,53)

\put(06,48){\makebox(0,0)[bl]{Steiner points}}
\put(05,45){\makebox(0,0)[bl]{for regions 1, 3}}

\put(05,38){\circle*{1.3}}





\put(10,43){\circle*{1.3}}

\put(25,43){\circle*{1.3}}

\put(10,39.5){\circle*{1.0}} \put(10,39.5){\circle{2.0}}

\put(12,39){\makebox(0,0)[bl]{\(s_{12}\)}}

\put(10,39.5){\line(-3,-1){5}} \put(10,39.5){\line(0,1){3.5}}
\put(10,39.5){\line(3,-2){10}}

\put(25,43){\line(-1,-2){5}}


\put(20,33){\circle*{1.3}}

\put(20,33){\line(-1,-2){5}}

\put(20,33){\line(1,-1){5}}

\put(30,18){\circle*{1.3}}

\put(30,33){\circle*{1.3}}



\put(27.5,28){\circle*{1.0}} \put(27.5,28){\circle{2.0}}

\put(29.5,27){\makebox(0,0)[bl]{\(s_{32}\)}}

\put(27.5,28){\line(-1,0){2.5}} \put(27.5,28){\line(1,2){2.5}}
\put(27.5,28){\line(1,-4){2.5}}

\put(25,28){\circle*{1.3}}

\put(15,23){\circle*{1.3}}

\put(15,23){\line(-3,-1){15}}

\put(15,23){\line(-1,-2){5}}

\put(00,18){\circle*{1.3}}



\put(10,13){\circle*{1.3}}

\put(20,7){\line(1,4){5}}

 \put(20,7){\circle*{1.3}}

\end{picture}
\end{center}

\begin{center}
\begin{picture}(68,67)
\put(1.5,62){\makebox(0,0)[bl]{Table 9. Data for multiple
 choice problem}}

\put(00,0){\line(1,0){68}} \put(00,49){\line(1,0){68}}
\put(00,60){\line(1,0){68}}


\put(00,00){\line(0,1){60}} \put(16,00){\line(0,1){60}}
\put(30,00){\line(0,1){60}} \put(44,00){\line(0,1){60}}
\put(55,00){\line(0,1){60}} \put(68,00){\line(0,1){60}}


\put(01,54.6){\makebox(0,0)[bl]{Region}}

\put(17,54.6){\makebox(0,0)[bl]{Binary}}
\put(17,51.4){\makebox(0,0)[bl]{variable}}

\put(32,55){\makebox(0,0)[bl]{Steiner}}
\put(32,51){\makebox(0,0)[bl]{point}}

\put(44.5,55){\makebox(0,0)[bl]{``Cost''}}
\put(47,51){\makebox(0,0)[bl]{\(c_{ij}\)}}

\put(55.5,55){\makebox(0,0)[bl]{``Profit''}}
\put(57,51){\makebox(0,0)[bl]{\(a_{ij}\)}}


\put(01,43.5){\makebox(0,0)[bl]{Region \(1\)}}

\put(21,44){\makebox(0,0)[bl]{\(x_{11}\)}}
\put(33,44){\makebox(0,0)[bl]{None}}
\put(47,44){\makebox(0,0)[bl]{\(0.0\)}}
\put(59,44){\makebox(0,0)[bl]{\(0.0\)}}

\put(21,40){\makebox(0,0)[bl]{\(x_{12}\)}}
\put(35,40){\makebox(0,0)[bl]{\(s_{11}\)}}
\put(47,40){\makebox(0,0)[bl]{\(3.1\)}}
\put(59,40){\makebox(0,0)[bl]{\(1.5\)}}

\put(21,36){\makebox(0,0)[bl]{\(x_{13}\)}}
\put(35,36){\makebox(0,0)[bl]{\(s_{12}\)}}
\put(47,36){\makebox(0,0)[bl]{\(1.2\)}}
\put(59,36){\makebox(0,0)[bl]{\(1.4\)}}



\put(00,34){\line(1,0){68}}

\put(01,29.5){\makebox(0,0)[bl]{Region \(2\)}}

\put(21,30){\makebox(0,0)[bl]{\(x_{21}\)}}
\put(33,30){\makebox(0,0)[bl]{None}}
\put(47,30){\makebox(0,0)[bl]{\(0.0\)}}
\put(59,30){\makebox(0,0)[bl]{\(0.0\)}}

\put(21,26){\makebox(0,0)[bl]{\(x_{22}\)}}
\put(35,26){\makebox(0,0)[bl]{\(s_{21}\)}}
\put(47,26){\makebox(0,0)[bl]{\(2.0\)}}
\put(59,26){\makebox(0,0)[bl]{\(1.3\)}}


\put(00,24){\line(1,0){68}}

\put(01,19.5){\makebox(0,0)[bl]{Region \(3\)}}

\put(21,20){\makebox(0,0)[bl]{\(x_{31}\)}}
\put(33,20){\makebox(0,0)[bl]{None}}
\put(47,20){\makebox(0,0)[bl]{\(0.0\)}}
\put(59,20){\makebox(0,0)[bl]{\(0.0\)}}

\put(21,16){\makebox(0,0)[bl]{\(x_{32}\)}}
\put(35,16){\makebox(0,0)[bl]{\(s_{31}\)}}
\put(47,16){\makebox(0,0)[bl]{\(2.4\)}}
\put(59,16){\makebox(0,0)[bl]{\(1.4\)}}

\put(21,12){\makebox(0,0)[bl]{\(x_{33}\)}}
\put(35,12){\makebox(0,0)[bl]{\(s_{32}\)}}
\put(47,12){\makebox(0,0)[bl]{\(1.8\)}}
\put(59,12){\makebox(0,0)[bl]{\(1.3\)}}



\put(00,10){\line(1,0){68}}

\put(01,05.5){\makebox(0,0)[bl]{Region \(4\)}}

\put(21,06){\makebox(0,0)[bl]{\(x_{41}\)}}
\put(33,6){\makebox(0,0)[bl]{None}}
\put(47,6){\makebox(0,0)[bl]{\(0.0\)}}
\put(59,6){\makebox(0,0)[bl]{\(0.0\)}}

\put(21,02){\makebox(0,0)[bl]{\(x_{42}\)}}
\put(35,02){\makebox(0,0)[bl]{\(s_{41}\)}}
\put(47,02){\makebox(0,0)[bl]{\(1.5\)}}
\put(59,02){\makebox(0,0)[bl]{\(1.2\)}}

\end{picture}
\end{center}

\subsection{Augmentation Problem}

 Graph augmentation problem is a generalization of
 ``hotlink assignment'' problem
  (e.g., \cite{esw76,khu97}).
 The goal is
 to modify an initial graph/network (e.g., by edges)
 such that the augmented graph will by satisfied
 some requirements (e.g., as increasing the connectivity).

\subsection{Transformation of Morphological Structure}

 Morphological system structure
 was suggested for the extended version of morphological analysis
 (HMMD) and contains the following parts
 (e.g., \cite{lev98,lev06,lev11agg,lev12morph,lev12hier}):
 (1) system hierarchy (e.g., a tree)
 (2) set of leaf nodes (i.e., system components),
 (3) sets of alternatives (DAs) for each system component;
 (4) estimates of the alternatives (ordinal as priority, interval,
 etc.),
 (5)  compatibility estimates for alternative pairs.
 A numerical example of the morphological system structure
 was presented in Fig. 16 and Table 4.
 In general,  the system morphological structure and its
 transformation are depicted in Fig. 23:
 ~\( S' \Rightarrow S'' \).
 Here,
 the following is transformed:

 (a) system hierarchy (tree) ~\( T' \Rightarrow T'' \),

 (a) basic system components as leaf nodes
   ~\( L'= \{1,...,i,...,m'\} \Rightarrow  L''= \{1,...,i,...,m''\} \).

 The numerical example of system structure transformation
 (as system reconfiguration)
 was described in section 3.1 (Fig. 13).

\begin{center}
\begin{picture}(60,64)
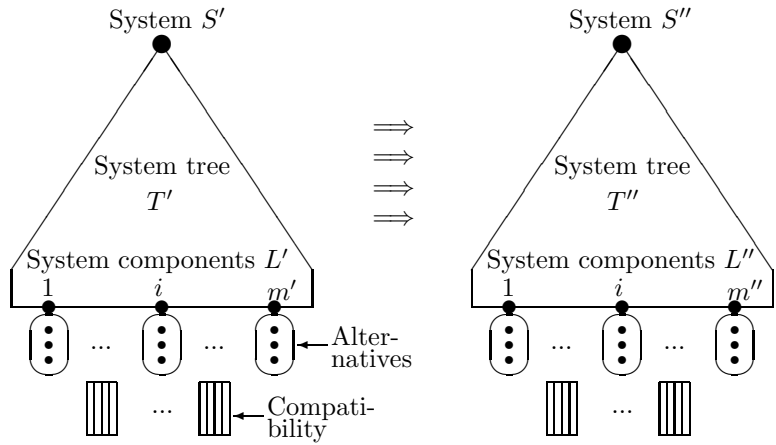
\put(05,00){\makebox(0,0)[bl]{Fig. 23. Morphological system
structure, its transformation}}

\put(13,59.5){\makebox(0,0)[bl]{System \(S'\)}}

\put(20,58){\circle*{2.5}}


\put(11,40){\makebox(0,0)[bl]{System tree}}
\put(18,36){\makebox(0,0)[bl]{\(T'\)}}

\put(00,28){\line(2,3){20}} \put(40,28){\line(-2,3){20}}
\put(00,23){\line(0,1){05}} \put(40,23){\line(0,1){05}}
\put(00,23){\line(1,0){40}}


\put(48,46){\makebox(0,0)[bl]{\(\Longrightarrow \)}}
\put(48,42){\makebox(0,0)[bl]{\(\Longrightarrow \)}}
\put(48,38){\makebox(0,0)[bl]{\(\Longrightarrow \)}}
\put(48,34){\makebox(0,0)[bl]{\(\Longrightarrow \)}}


\put(02,27.5){\makebox(0,0)[bl]{System components \(L'\)}}

\put(04,24.5){\makebox(0,0)[bl]{\(1\)}}

\put(05,23){\circle*{1.8}} \put(05,18){\oval(5,8)}

\put(05,20){\circle*{1}} \put(05,18){\circle*{1}}
\put(05,16){\circle*{1}}

\put(19,24.5){\makebox(0,0)[bl]{\(i\)}}

\put(20,23){\circle*{1.8}} \put(20,18){\oval(5,8)}

\put(20,20){\circle*{1}} \put(20,18){\circle*{1}}
\put(20,16){\circle*{1}}

\put(34,24){\makebox(0,0)[bl]{\(m'\)}}

\put(35,23){\circle*{1.8}} \put(35,18){\oval(5,8)}

\put(35,20){\circle*{1}} \put(35,18){\circle*{1}}
\put(35,16){\circle*{1}}


\put(42,18){\vector(-1,0){4}}

\put(42.5,18){\makebox(0,0)[bl]{Alter-}}
\put(42.5,15){\makebox(0,0)[bl]{natives}}


\put(10,06){\line(1,0){04}} \put(10,13){\line(1,0){04}}
\put(10,06){\line(0,1){7}} \put(14,06){\line(0,1){7}}

\put(11,06){\line(0,1){7}} \put(12,06){\line(0,1){7}}
\put(13,06){\line(0,1){7}}

\put(25,06){\line(1,0){04}} \put(25,13){\line(1,0){04}}
\put(25,06){\line(0,1){7}} \put(29,06){\line(0,1){7}}

\put(26,06){\line(0,1){7}} \put(27,06){\line(0,1){7}}
\put(28,06){\line(0,1){7}}


\put(33.5,08.5){\vector(-1,0){4}}

\put(34,08){\makebox(0,0)[bl]{Compati-}}
\put(34,05){\makebox(0,0)[bl]{bility}}


\put(10.5,17.5){\makebox(0,0)[bl]{...}}
\put(25.5,17.5){\makebox(0,0)[bl]{...}}

\put(18.6,9){\makebox(0,0)[bl]{...}}

\end{picture}
%
\begin{picture}(40,64)

\put(13,59.5){\makebox(0,0)[bl]{System \(S''\)}}

\put(20,58){\circle*{2.5}}


\put(11,40){\makebox(0,0)[bl]{System tree}}
\put(18,36){\makebox(0,0)[bl]{\(T''\)}}

\put(00,28){\line(2,3){20}} \put(40,28){\line(-2,3){20}}
\put(00,23){\line(0,1){05}} \put(40,23){\line(0,1){05}}
\put(00,23){\line(1,0){40}}


\put(02,27.5){\makebox(0,0)[bl]{System components \(L''\)}}

\put(04,24.5){\makebox(0,0)[bl]{\(1\)}}

\put(05,23){\circle*{1.8}} \put(05,18){\oval(5,8)}

\put(05,20){\circle*{1}} \put(05,18){\circle*{1}}
\put(05,16){\circle*{1}}

\put(19,24.5){\makebox(0,0)[bl]{\(i\)}}

\put(20,23){\circle*{1.8}} \put(20,18){\oval(5,8)}

\put(20,20){\circle*{1}} \put(20,18){\circle*{1}}
\put(20,16){\circle*{1}}

\put(34,24){\makebox(0,0)[bl]{\(m''\)}}

\put(35,23){\circle*{1.8}} \put(35,18){\oval(5,8)}

\put(35,20){\circle*{1}} \put(35,18){\circle*{1}}
\put(35,16){\circle*{1}}


\put(10,06){\line(1,0){04}} \put(10,13){\line(1,0){04}}
\put(10,06){\line(0,1){7}} \put(14,06){\line(0,1){7}}

\put(11,06){\line(0,1){7}} \put(12,06){\line(0,1){7}}
\put(13,06){\line(0,1){7}}

\put(25,06){\line(1,0){04}} \put(25,13){\line(1,0){04}}
\put(25,06){\line(0,1){7}} \put(29,06){\line(0,1){7}}

\put(26,06){\line(0,1){7}} \put(27,06){\line(0,1){7}}
\put(28,06){\line(0,1){7}}


\put(10.5,17.5){\makebox(0,0)[bl]{...}}
\put(25.5,17.5){\makebox(0,0)[bl]{...}}

\put(18.6,9){\makebox(0,0)[bl]{...}}

\end{picture}
\end{center}

\subsection{Transformation of Layered Structure}

 The multilayer system structure is examined as follows:
 (a) set of layers,
 (b) for each layer: set of nodes, topology over the nodes,
 (c) connection of nodes for neighbor layer.

 Thus, the modification process of the layered structure can
 involve the following:

 {\it 1.} Modification of the layer-structure:
 (1.1) Addition of a layer,
 (1.2) Deletion of a layer.

 {\it 2.} Modification of a layer:
   (2.1) addition of layer nodes,
  (2.2) deletion of layer nodes
  (2.3) modification for the layer topology
 (a network over the layer nodes).

 {\it 3.} Modification of inter-layer connections
 (i.e., connection for nodes of neighbor layers):
  reassignment.

 Fig. 24 depicts the modification process
 ~\(S' \Rightarrow S''\).

 A numerical example for two-layered network will be presented
 later (extension of layers, reassignment).

\begin{center}
\begin{picture}(69,54)

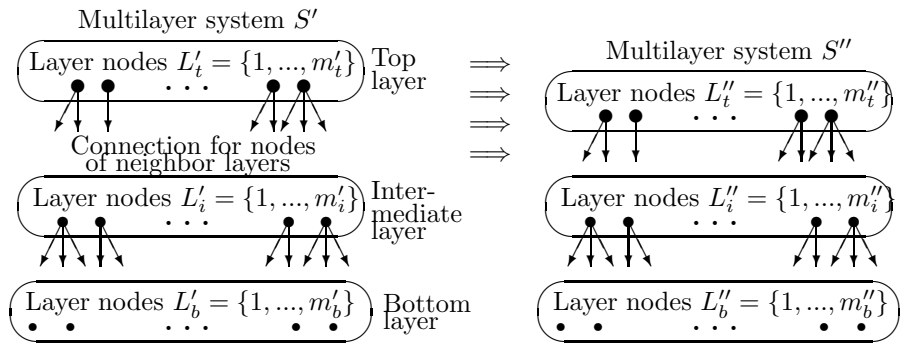
\put(21.7,00){\makebox(0,0)[bl]{Fig. 24. Transformation of
 multilayer structure}}

\put(09,47){\makebox(0,0)[bl]{Multilayer system \(S'\) }}


\put(48,42){\makebox(0,0)[bl]{Top}}
\put(48,39){\makebox(0,0)[bl]{layer}}

\put(02.5,41.2){\makebox(0,0)[bl]{Layer nodes \(L'_{t} =
\{1,...,m'_{t}\}\)}}

\put(24,42){\oval(46,8)}

\put(09,40){\circle*{1.8}} \put(13,40){\circle*{1.8}}
\put(20.5,39.5){\makebox(0,0)[bl]{{\bf .~.~.}}}

\put(35,40){\circle*{1.8}}  \put(39,40){\circle*{1.8}}


\put(09,40){\vector(0,-1){06}} \put(09,40){\vector(-1,-2){03}}
\put(13,40){\vector(0,-1){06}}

\put(35,40){\vector(0,-1){06}} \put(35,40){\vector(-1,-2){03}}

\put(39,40){\vector(0,-1){06}} \put(39,40){\vector(-1,-2){03}}
\put(39,40){\vector(1,-2){03}}

\put(8,31){\makebox(0,0)[bl]{Connection for nodes}}
\put(10,28){\makebox(0,0)[bl]{of neighbor layers}}


\put(48,25){\makebox(0,0)[bl]{Inter-}}
\put(48,22){\makebox(0,0)[bl]{mediate}}
\put(48,19){\makebox(0,0)[bl]{layer}}

\put(03,23.2){\makebox(0,0)[bl]{Layer nodes \(L'_{i} =
\{1,...,m'_{i}\}\)}}

\put(24,24){\oval(46,8)}

\put(07,22){\circle*{1.5}} \put(12,22){\circle*{1.5}}
\put(20.5,21.5){\makebox(0,0)[bl]{{\bf .~.~.}}}

\put(37,22){\circle*{1.5}}  \put(42,22){\circle*{1.5}}


\put(07,22){\vector(0,-1){06}} \put(07,22){\vector(-1,-2){03}}
\put(07,22){\vector(1,-2){03}}

\put(12,22){\vector(0,-1){06}} \put(12,22){\vector(1,-2){03}}

\put(37,22){\vector(0,-1){06}} \put(37,22){\vector(-1,-2){03}}

\put(42,22){\vector(0,-1){06}} \put(42,22){\vector(-1,-2){03}}
\put(42,22){\vector(1,-2){03}}


\put(61,42){\makebox(0,0)[bl]{\(\Longrightarrow \)}}
\put(61,38){\makebox(0,0)[bl]{\(\Longrightarrow \)}}
\put(61,34){\makebox(0,0)[bl]{\(\Longrightarrow \)}}
\put(61,30){\makebox(0,0)[bl]{\(\Longrightarrow \)}}



\put(49.6,10){\makebox(0,0)[bl]{Bottom}}
\put(49.6,07){\makebox(0,0)[bl]{layer}}

\put(24,10){\oval(48,8)}

\put(02,09.2){\makebox(0,0)[bl]{Layer nodes \(L'_{b} =
\{1,...,m'_{b}\}\)}}

\put(03,08){\circle*{1}} \put(08,08){\circle*{1}}
\put(20.5,07.5){\makebox(0,0)[bl]{{\bf .~.~.}}}

\put(38,08){\circle*{1}}  \put(43,08){\circle*{1}}

\end{picture}
%
\begin{picture}(48,45)



\put(09,43){\makebox(0,0)[bl]{Multilayer system \(S''\) }}

\put(02.5,37.2){\makebox(0,0)[bl]{Layer nodes \(L''_{t} =
\{1,...,m''_{t}\}\)}}

\put(24,38){\oval(46,8)}

\put(09,36){\circle*{1.8}} \put(13,36){\circle*{1.8}}
\put(20.5,35.5){\makebox(0,0)[bl]{{\bf .~.~.}}}

\put(35,36){\circle*{1.8}}  \put(39,36){\circle*{1.8}}


\put(09,36){\vector(0,-1){06}} \put(09,36){\vector(-1,-2){03}}
\put(13,36){\vector(0,-1){06}}

\put(35,36){\vector(0,-1){06}} \put(35,36){\vector(-1,-2){03}}

\put(39,36){\vector(0,-1){06}} \put(39,36){\vector(-1,-2){03}}
\put(39,36){\vector(1,-2){03}}



\put(03,23.2){\makebox(0,0)[bl]{Layer nodes \(L''_{i} =
\{1,...,m''_{i}\}\)}}

\put(24,24){\oval(46,8)}

\put(07,22){\circle*{1.5}} \put(12,22){\circle*{1.5}}
\put(20.5,21.5){\makebox(0,0)[bl]{{\bf .~.~.}}}

\put(37,22){\circle*{1.5}}  \put(42,22){\circle*{1.5}}


\put(07,22){\vector(0,-1){06}} \put(07,22){\vector(-1,-2){03}}
\put(07,22){\vector(1,-2){03}}

\put(12,22){\vector(0,-1){06}} \put(12,22){\vector(1,-2){03}}

\put(37,22){\vector(0,-1){06}} \put(37,22){\vector(-1,-2){03}}

\put(42,22){\vector(0,-1){06}} \put(42,22){\vector(-1,-2){03}}
\put(42,22){\vector(1,-2){03}}



\put(24,10){\oval(48,8)}

\put(02,09.2){\makebox(0,0)[bl]{Layer nodes \(L''_{b} =
\{1,...,m''_{b}\}\)}}

\put(03,08){\circle*{1}} \put(08,08){\circle*{1}}
\put(20.5,07.5){\makebox(0,0)[bl]{{\bf .~.~.}}}

\put(38,08){\circle*{1}}  \put(43,08){\circle*{1}}

\end{picture}
\end{center}

\section{Examples for Network Improvement/Extension \cite{levsib10} }

 \subsection{Network Hierarchy}

 A traditional network hierarchy can be considered as follows
 (e.g.,  \cite{kuz05,mur99,tanen02}):
 ~(a) international (multi-country, continent) network  GAN;
 ~(b) metropolitan network MN;
 ~(c) wide area network WAN; and
 ~(d) local area network LAN.
 From the ``engineering'' viewpoint, hierarchical layers involve the following:
 (i) backbone network;
 (ii) global network as a set of
 interconnected network segments
 (including additional centers, cross-connections, and bridges);
 (iii) access network/network segment (cluster):
 bi-connected topology (about 20 nodes); and
 (iv) distributed network (a simple hard topology, e.g., bus, tree, ring).

 A simplified example of
  a three-layer network hierarchy is the following:
 (a) ``center'' systems (e.g., hubs),
 (b) access points, and
  (c) distributed networks.

\subsection{Requirements/Criteria}

  Contextual classification of the requirements
 to communication networks is considered as follows
 (e.g., \cite{kuz05,levsib10}):
 (1) ``user'' requirements:
 cost,
 time characteristics,
 quality (information errors, reliability of connection);
  (2) system requirements:
%
  cost,
%
 reliability (or stability, redundancy, k-connectivity),
%
 manageability,
%
 maintainability,
%
 testability,
%
 modularity,
%
 adaptability,
%
  safety,
  and
%
 flexibility (e.g., reconfigurability);
 (3) mobility requirements; and
 (4) system evolution/development requirements:
%
 possibility for re-design (upgradeability).

 On the other hand, it is possible to consider
 a correspondence of the requirements to network hierarchical
 layers, for example:
 (1) top layer:
 cost,
%
 safety (stability, reliability, redundancy,
 survivability),
%
  manageability, adaptability, flexibility,
%
 upgradeability;
 ~(2) medium layer:
%
  basic quality,
%
  reliability, and
%
  survivability;
 ~(3) bottom layer:
%
 basic quality (time, cost, etc.) and
%
  reliability.

\subsection{Network Design/Development: Basic Problems}

 In Table 10, some basic problems are pointed out from the viewpoint
 of network layers and two types of activities:
  network design and network improvement/extension.

 Thus,
 underlaying combinatorial optimization problems
 (e.g.,
 minimum spanning tree,
 minimum Steiner tree,
 covering,
  design of a k-connected topology,
 location/placement of network nodes,
 selection of some additional links as additional network edges,
 selection and location of additional network nodes)
 can be considered.
 In the numerical examples,
 the following combinatorial problems are used:
 (1) multiple criteria ranking
 (e.g., \cite{lev98,lev12b,roy96}),
 here our modification of outranking technique
 (Electre-like method) is used
  \cite{lev12b};
 (2) clustering
 (e.g.,  \cite{jain99}),
  here our modification of agglomerative algorithm is used
 \cite{lev07clust};
 (3) assignment/allocation problem
  (e.g., \cite{gar79}),
 here our heuristic is used \cite{levpet10a}; and
 (4) multicriteria multiple choice problem,
 (e.g., \cite{gar79,keller04},
 here our heuristic is used \cite{levsaf10a}.

\begin{center}
\begin{picture}(112,110)

\put(09.5,106){\makebox(0,0)[bl]{Table 10. Network layers and
design/improvement problems}}

\put(00,0){\line(1,0){112}} \put(00,94){\line(1,0){112}}
\put(00,104){\line(1,0){112}}

\put(00,0){\line(0,1){104}} \put(26,0){\line(0,1){104}}
\put(66,0){\line(0,1){104}} \put(112,0){\line(0,1){104}}


\put(0.5,100){\makebox(0,0)[bl]{Network layer}}

\put(26.5,100){\makebox(0,0)[bl]{Some basic structural}}
\put(26.5,96){\makebox(0,0)[bl]{design problems}}

\put(66.5,99.5){\makebox(0,0)[bl]{Basic network improvement/}}
\put(66.5,96){\makebox(0,0)[bl]{extension design problems}}

\put(01,89){\makebox(0,0)[bl]{System of hubs}}
\put(01,85){\makebox(0,0)[bl]{(centers)}}

\put(26.5,89){\makebox(0,0)[bl]{Network topology design:}}
\put(26.5,85){\makebox(0,0)[bl]{(a) clustering}}
\put(26.5,81){\makebox(0,0)[bl]{(b)
  spanning tree/forest}}

\put(26.5,77){\makebox(0,0)[bl]{(c)
 Steiner tree}}

\put(26.5,73){\makebox(0,0)[bl]{(d) k-connected network}}

\put(26.5,69){\makebox(0,0)[bl]{(e) covering problem}}

\put(66.5,89){\makebox(0,0)[bl]{(1) addition of hub(s)}}
\put(72,85){\makebox(0,0)[bl]{(center(s))}}

\put(66.5,81){\makebox(0,0)[bl]{(2) addition of links}}
\put(72,77){\makebox(0,0)[bl]{(e.g., bridge(s))}}

\put(66.5,73){\makebox(0,0)[bl]{(3) redesign of network}}
\put(72,69){\makebox(0,0)[bl]{topology}}

\put(0.5,64){\makebox(0,0)[bl]{Network over}}
\put(0.5,60){\makebox(0,0)[bl]{gateways}}

\put(26.5,64){\makebox(0,0)[bl]{Network topology design:}}
\put(26.5,60){\makebox(0,0)[bl]{(a) clustering}}

\put(26.5,56){\makebox(0,0)[bl]{(b)
 spanning tree/forest}}

\put(26.5,52){\makebox(0,0)[bl]{(c)
 Steiner tree}}

\put(26.5,48){\makebox(0,0)[bl]{(d) k-connected network}}
\put(26.5,44){\makebox(0,0)[bl]{(e) covering problem}}

\put(66.5,64){\makebox(0,0)[bl]{(1) addition of access}}
\put(72,60){\makebox(0,0)[bl]{point(s)}}

\put(66.5,56){\makebox(0,0)[bl]{(2) addition of links }}
\put(72,52){\makebox(0,0)[bl]{(e.g., bridge(s))}}

\put(66.5,48){\makebox(0,0)[bl]{(3) redesign of network}}
\put(72,44){\makebox(0,0)[bl]{topology}}

\put(0.5,39){\makebox(0,0)[bl]{Access network}}

\put(26.5,39){\makebox(0,0)[bl]{Network topology design:}}

\put(26.5,35){\makebox(0,0)[bl]{(a) clustering}}

\put(26.5,31){\makebox(0,0)[bl]{(b)
 spanning tree/forest}}

\put(26.5,27){\makebox(0,0)[bl]{(c)
 Steiner tree}}

\put(26.5,23){\makebox(0,0)[bl]{(d) 2-connected network}}

\put(26.5,19){\makebox(0,0)[bl]{(e) covering problem}}

\put(66.5,39){\makebox(0,0)[bl]{(1) addition of access}}
\put(72,35){\makebox(0,0)[bl]{point(s)}}

\put(66.5,31){\makebox(0,0)[bl]{(2) addition of links }}
\put(72,27){\makebox(0,0)[bl]{(e.g., bridge(s))}}

\put(66.5,23){\makebox(0,0)[bl]{(3) redesign of network }}
\put(72,19){\makebox(0,0)[bl]{topology}}


\put(0.5,14){\makebox(0,0)[bl]{Distributed}}
\put(0.5,10){\makebox(0,0)[bl]{network}}

\put(26.5,14){\makebox(0,0)[bl]{Last mile problem:}}
\put(26.5,10){\makebox(0,0)[bl]{(a) choice of topology }}
\put(26.5,06){\makebox(0,0)[bl]{(b) choice of connection}}
\put(26.5,02){\makebox(0,0)[bl]{(c) choice of access point}}

\put(66.5,14){\makebox(0,0)[bl]{(1) addition of user(s)}}
\put(66.5,10){\makebox(0,0)[bl]{(2) addition of distributed}}
\put(72,06){\makebox(0,0)[bl]{network(s)}}

\end{picture}
\end{center}

\subsection{Network Improvement}

 Here, a simplified example for improvement
  of Moscow phone network (at a macro level) is examined
 \cite{levsib10}.
 Table 11 contains the considered Moscow regions
 ~\(\{ A_{1},...,A_{9} \}\)~
 and their estimates upon parameters
 (expert judgment, ordinal scale [1,10]):~
 population \(P_{1}\),
 level of an existing communication infrastructure \(P_{2}\), and
 volume of the region array \(P_{3}\).

\begin{center}
\begin{picture}(51,56)

\put(00,52){\makebox(0,0)[bl]{Table 11. Regions,  their
estimates}}

\put(00,0){\line(1,0){51}} \put(00,38){\line(1,0){51}}

\put(30,44){\line(1,0){21}}

\put(00,50){\line(1,0){51}}

\put(00,0){\line(0,1){50}}
 \put(10,0){\line(0,1){50}}

 \put(30,0){\line(0,1){50}}
\put(51,0){\line(0,1){50}}


\put(37,38){\line(0,1){06}} \put(44,38){\line(0,1){06}}

\put(11,46){\makebox(0,0)[bl]{Region}}

\put(31.7,46){\makebox(0,0)[bl]{Parameters}}

\put(31.5,39){\makebox(0,0)[bl]{\(P_{1}\)}}
\put(38.5,39){\makebox(0,0)[bl]{\(P_{2}\)}}
\put(45.5,39){\makebox(0,0)[bl]{\(P_{3}\)}}


\put(1,34){\makebox(0,0)[bl]{\(A_{1}\)}}
\put(1,30){\makebox(0,0)[bl]{\(A_{2}\)}}
\put(1,26){\makebox(0,0)[bl]{\(A_{3}\)}}
\put(1,22){\makebox(0,0)[bl]{\(A_{4}\)}}
\put(1,18){\makebox(0,0)[bl]{\(A_{5}\)}}
\put(1,14){\makebox(0,0)[bl]{\(A_{6}\)}}
\put(1,10){\makebox(0,0)[bl]{\(A_{7}\)}}
\put(1,06){\makebox(0,0)[bl]{\(A_{8}\)}}
\put(1,02){\makebox(0,0)[bl]{\(A_{9}\)}}

\put(11,34){\makebox(0,0)[bl]{Central}}
\put(11,30){\makebox(0,0)[bl]{South}}
\put(11,26){\makebox(0,0)[bl]{South-west}}
\put(11,22){\makebox(0,0)[bl]{South-east}}
\put(11,18){\makebox(0,0)[bl]{North}}
\put(11,14){\makebox(0,0)[bl]{North-east}}
\put(11,10){\makebox(0,0)[bl]{North-west}}
\put(11,06){\makebox(0,0)[bl]{West}}
\put(11,02){\makebox(0,0)[bl]{East}}


\put(31,34){\makebox(0,0)[bl]{\(10\)}}
\put(38,34){\makebox(0,0)[bl]{\(10\)}}
\put(46,34){\makebox(0,0)[bl]{\(9\)}}



\put(32,30){\makebox(0,0)[bl]{\(8\)}}
\put(39,30){\makebox(0,0)[bl]{\(9\)}}
\put(45,30){\makebox(0,0)[bl]{\(10\)}}



\put(32,26){\makebox(0,0)[bl]{\(6\)}}
\put(39,26){\makebox(0,0)[bl]{\(7\)}}
\put(46,26){\makebox(0,0)[bl]{\(4\)}}



\put(32,22){\makebox(0,0)[bl]{\(3\)}}
\put(39,22){\makebox(0,0)[bl]{\(8\)}}
\put(46,22){\makebox(0,0)[bl]{\(8\)}}



\put(32,18){\makebox(0,0)[bl]{\(4\)}}
\put(39,18){\makebox(0,0)[bl]{\(4\)}}
\put(46,18){\makebox(0,0)[bl]{\(5\)}}



\put(32,14){\makebox(0,0)[bl]{\(1\)}}
\put(39,14){\makebox(0,0)[bl]{\(1\)}}
\put(46,14){\makebox(0,0)[bl]{\(9\)}}



\put(32,10){\makebox(0,0)[bl]{\(4\)}}
\put(39,10){\makebox(0,0)[bl]{\(5\)}}
\put(46,10){\makebox(0,0)[bl]{\(4\)}}



\put(32,06){\makebox(0,0)[bl]{\(5\)}}
\put(39,06){\makebox(0,0)[bl]{\(6\)}}
\put(46,06){\makebox(0,0)[bl]{\(4\)}}



\put(32,02){\makebox(0,0)[bl]{\(2\)}}
\put(39,02){\makebox(0,0)[bl]{\(4\)}}
\put(46,02){\makebox(0,0)[bl]{\(5\)}}


\end{picture}
\end{center}
 After clustering (hierarchical clustering is used
 \cite{lev07clust}),
 the following clusters are obtained:
 ~\( G^{1} = \{ A_{1} \} \),
 ~\( G^{2} = \{ A_{2} \} \),
 ~\( G^{3} = \{ A_{3},A_{8} \} \),
 ~\( G^{4} = \{ A_{4} \} \),
 ~\( G^{5} = \{ A_{5},A_{7},A_{9} \} \),
 and
 ~\( G^{6} = \{ A_{6} \} \).
 Table 12 contains five development/improvement actions at the region
 level
 ~\(\{D_{1},D_{2},D_{3},D_{4},D_{5}\}\)~
 (\(D_{1}\) corresponds to a case when activities are absent).
 The actions are evaluated upon criteria
 (ordinal scale [0.5], expert judgment):~
 generalized profit
 ~(\(C_{1}\)),
 complexity ~(\(C_{2}\)),
 perspective profit ~(\(C_{3}\)), and
 expenditure ~(\(C_{4}\)).

\begin{center}
\begin{picture}(78,44)
\put(04,40){\makebox(0,0)[bl]{Table 12. Development/improvement
actions}}

\put(00,0){\line(1,0){78}} \put(00,26){\line(1,0){78}}

\put(50,32){\line(1,0){28}}

\put(00,38){\line(1,0){78}}

\put(00,0){\line(0,1){38}} \put(10,0){\line(0,1){38}}
\put(50,0){\line(0,1){38}} \put(78,0){\line(0,1){38}}


\put(57,26){\line(0,1){06}} \put(64,26){\line(0,1){06}}
\put(71,26){\line(0,1){06}}

\put(11,33.6){\makebox(0,0)[bl]{Development action}}

\put(58,34){\makebox(0,0)[bl]{Criteria}}

\put(51.5,27.6){\makebox(0,0)[bl]{\(C_{1}\)}}
\put(58.5,27.6){\makebox(0,0)[bl]{\(C_{2}\)}}
\put(65.5,27.6){\makebox(0,0)[bl]{\(C_{3}\)}}
\put(72.5,27.6){\makebox(0,0)[bl]{\(C_{4}\)}}

\put(1,21.5){\makebox(0,0)[bl]{\(D_{1}\)}}
\put(1,17.5){\makebox(0,0)[bl]{\(D_{2}\)}}
\put(1,13.5){\makebox(0,0)[bl]{\(D_{3}\)}}
\put(1,09.5){\makebox(0,0)[bl]{\(D_{4}\)}}
\put(1,01.5){\makebox(0,0)[bl]{\(D_{5}\)}}

\put(11,22){\makebox(0,0)[bl]{None}}
\put(11,18){\makebox(0,0)[bl]{New links}}
\put(11,14){\makebox(0,0)[bl]{Reparation of links}}
\put(11,10){\makebox(0,0)[bl]{Extension (new links}}
\put(11,06){\makebox(0,0)[bl]{and devices)}}
\put(11,02){\makebox(0,0)[bl]{Deletion of old links}}



\put(52,22){\makebox(0,0)[bl]{\(0\)}}
\put(59,22){\makebox(0,0)[bl]{\(0\)}}
\put(66,22){\makebox(0,0)[bl]{\(0\)}}
\put(73,22){\makebox(0,0)[bl]{\(0\)}}


\put(52,18){\makebox(0,0)[bl]{\(5\)}}
\put(59,18){\makebox(0,0)[bl]{\(5\)}}
\put(66,18){\makebox(0,0)[bl]{\(5\)}}
\put(73,18){\makebox(0,0)[bl]{\(5\)}}


\put(52,14){\makebox(0,0)[bl]{\(2\)}}
\put(59,14){\makebox(0,0)[bl]{\(2\)}}
\put(66,14){\makebox(0,0)[bl]{\(2\)}}
\put(73,14){\makebox(0,0)[bl]{\(2\)}}


\put(52,10){\makebox(0,0)[bl]{\(3\)}}
\put(59,10){\makebox(0,0)[bl]{\(3\)}}
\put(66,10){\makebox(0,0)[bl]{\(3\)}}
\put(73,10){\makebox(0,0)[bl]{\(3\)}}


\put(52,02){\makebox(0,0)[bl]{\(1\)}}
\put(59,02){\makebox(0,0)[bl]{\(1\)}}
\put(66,02){\makebox(0,0)[bl]{\(1\)}}
\put(73,02){\makebox(0,0)[bl]{\(1\)}}

\end{picture}
\end{center}

 Fig. 25 depicts a two-stage solving scheme.
 A composite development plan is based on
   the following multcriteria multiple choice problem
  (a simplest greedy-like heuristic is used):
 \[\max \sum_{i=1}^{6} \sum_{j=1}^{5} c^{1}_{ij} x_{ij},
 ~~\min \sum_{i=1}^{6} \sum_{j=1}^{5} c^{2}_{ij} x_{ij},
 ~~\max \sum_{i=1}^{6} \sum_{j=1}^{5} c^{3}_{ij} x_{ij},
 ~~\min \sum_{i=1}^{6} \sum_{j=1}^{5} c^{4}_{ij} x_{ij}\]
 \[s.t. ~~ \sum_{i=1}^{6} \sum_{j=1}^{5} a_{ij} x_{ij} \leq b,
 ~~\sum_{j=1}^{5} x_{ij}=1 ~ i=\overline{1,6},
 ~~ x_{ij} \in \{0,1\}.\]
  As a result, the following solution is obtained:
  ~\(S_{1} = < D^{1}_{4} \star D^{2}_{3} \star D^{3}_{5} \star D^{4}_{1}
 \star D^{5}_{1} \star D^{6}_{1} > \).
 In the case of  interconnections of solutions for
 neighbor regions,
  it is necessary to use combinatorial synthesis based on
   HMMD \cite{lev98,lev06,lev12morph,lev12a}.


\begin{center}
\begin{picture}(82,61)
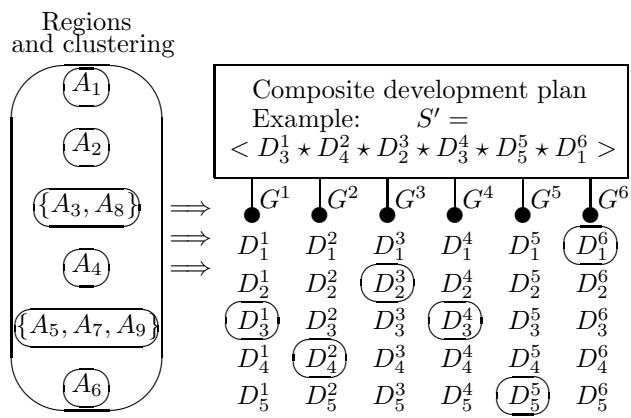
\put(06,0){\makebox(0,0)[bl]{Fig. 25. Illustration for two-stage
 scheme \cite{levsib10}}}


\put(04,57){\makebox(0,0)[bl]{Regions }}

\put(00,54){\makebox(0,0)[bl]{and clustering}}

\put(10,30){\oval(20,46)}

\put(10,50){\oval(6,5)} \put(8,49){\makebox(0,0)[bl]{\(A_{1}\)}}

\put(10,42){\oval(6,5)} \put(8,41){\makebox(0,0)[bl]{\(A_{2}\)}}

\put(10,34){\oval(14,5)}
\put(3.5,32.5){\makebox(0,0)[bl]{\(\{A_{3},A_{8}\}\)}}

\put(10,26){\oval(6,5)} \put(8,25){\makebox(0,0)[bl]{\(A_{4}\)}}

\put(10,18){\oval(19,5)}
\put(0.6,16.5){\makebox(0,0)[bl]{\(\{A_{5},A_{7},A_{9}\}\)}}

\put(10,10){\oval(6,5)} \put(8,9){\makebox(0,0)[bl]{\(A_{6}\)}}


\put(21,33){\makebox(0,0)[bl]{\(\Longrightarrow\)}}
\put(21,29){\makebox(0,0)[bl]{\(\Longrightarrow\)}}
\put(21,25){\makebox(0,0)[bl]{\(\Longrightarrow\)}}

\put(32,48){\makebox(0,0)[bl]{Composite development plan}}
\put(32,44){\makebox(0,0)[bl]{Example:  ~~~~~\(S' = \)}}

\put(29,40){\makebox(0,0)[bl]{\(<D^{1}_{3}\star D^{2}_{4}\star
D^{3}_{2}\star D^{4}_{3}\star D^{5}_{5}\star D^{6}_{1}>\)}}

\put(27,38){\line(1,0){55}} \put(27,53){\line(1,0){55}}
\put(27,38){\line(0,1){15}} \put(82,38){\line(0,1){15}}

\put(32,33){\line(0,1){5}} \put(32,33){\circle*{2}}
\put(33,34){\makebox(0,0)[bl]{\(G^{1}\)}}

\put(30,27){\makebox(0,0)[bl]{\(D^{1}_{1}\)}}
\put(30,22){\makebox(0,0)[bl]{\(D^{1}_{2}\)}}
\put(30,17){\makebox(0,0)[bl]{\(D^{1}_{3}\)}}
\put(30,12){\makebox(0,0)[bl]{\(D^{1}_{4}\)}}
\put(30,07){\makebox(0,0)[bl]{\(D^{1}_{5}\)}}

\put(32,19){\oval(7,5)}

\put(41,33){\line(0,1){5}} \put(41,33){\circle*{2}}
\put(42,34){\makebox(0,0)[bl]{\(G^{2}\)}}

\put(39,27){\makebox(0,0)[bl]{\(D^{2}_{1}\)}}
\put(39,22){\makebox(0,0)[bl]{\(D^{2}_{2}\)}}
\put(39,17){\makebox(0,0)[bl]{\(D^{2}_{3}\)}}
\put(39,12){\makebox(0,0)[bl]{\(D^{2}_{4}\)}}
\put(39,07){\makebox(0,0)[bl]{\(D^{2}_{5}\)}}

\put(41,14){\oval(7,5)}

\put(50,33){\line(0,1){5}} \put(50,33){\circle*{2}}
\put(51,34){\makebox(0,0)[bl]{\(G^{3}\)}}

\put(48,27){\makebox(0,0)[bl]{\(D^{3}_{1}\)}}
\put(48,22){\makebox(0,0)[bl]{\(D^{3}_{2}\)}}
\put(48,17){\makebox(0,0)[bl]{\(D^{3}_{3}\)}}
\put(48,12){\makebox(0,0)[bl]{\(D^{3}_{4}\)}}
\put(48,07){\makebox(0,0)[bl]{\(D^{3}_{5}\)}}

\put(50,24){\oval(7,5)}

\put(59,33){\line(0,1){5}} \put(59,33){\circle*{2}}
\put(60,34){\makebox(0,0)[bl]{\(G^{4}\)}}

\put(57,27){\makebox(0,0)[bl]{\(D^{4}_{1}\)}}
\put(57,22){\makebox(0,0)[bl]{\(D^{4}_{2}\)}}
\put(57,17){\makebox(0,0)[bl]{\(D^{4}_{3}\)}}
\put(57,12){\makebox(0,0)[bl]{\(D^{4}_{4}\)}}
\put(57,07){\makebox(0,0)[bl]{\(D^{4}_{5}\)}}

\put(59,19){\oval(7,5)}

\put(68,33){\line(0,1){5}} \put(68,33){\circle*{2}}
\put(69,34){\makebox(0,0)[bl]{\(G^{5}\)}}

\put(66,27){\makebox(0,0)[bl]{\(D^{5}_{1}\)}}
\put(66,22){\makebox(0,0)[bl]{\(D^{5}_{2}\)}}
\put(66,17){\makebox(0,0)[bl]{\(D^{5}_{3}\)}}
\put(66,12){\makebox(0,0)[bl]{\(D^{5}_{4}\)}}
\put(66,07){\makebox(0,0)[bl]{\(D^{5}_{5}\)}}

\put(68,09){\oval(7,5)}

\put(77,33){\line(0,1){5}} \put(77,33){\circle*{2}}
\put(78,34){\makebox(0,0)[bl]{\(G^{6}\)}}

\put(75,27){\makebox(0,0)[bl]{\(D^{6}_{1}\)}}
\put(75,22){\makebox(0,0)[bl]{\(D^{6}_{2}\)}}
\put(75,17){\makebox(0,0)[bl]{\(D^{6}_{3}\)}}
\put(75,12){\makebox(0,0)[bl]{\(D^{6}_{4}\)}}
\put(75,07){\makebox(0,0)[bl]{\(D^{6}_{5}\)}}

\put(77,29){\oval(7,5)}

\end{picture}
\end{center}

\subsection{Network Extension}

 Here,
  an applied example for
 a regional communication network is considered:
 ~(i) there exists a communication network for a region,
 ~(ii) it is needed to design an additional communication network
 for a neighbor region.
 Three extension strategies may be considered:

 {\it Strategy I.} Designing the additional communication network
 (i.e., definition of possible positions for
 communication facilities,
 location of communication devices,
 definition of system modes, etc.)
 and synthesis of the obtained two networks.

 {\it Strategy II.}
 Designing a new ``generalized'' communication network for
 an integrated region
 (i.e., the previous region and the additional regions, integrated design).

 {\it Strategy III.}
 Designing the additional communication network
  for the neighbor region and modification of the communication
  network for the previous region as coordination
  between the network for previous region and neighbor regions
 (e.g., reconfiguration, replacement of communication nodes, re-linking).

 Note, the network extension approaches can be used for various hierarchical
 layers of a communication network.
%

 Now,
 two extension strategies above
 at the layer of connection between users and access points
 (i.e., assignment of users to access points)
 are examined.
 Two regions (including users and access point) are considered:
 (i) initial regions
 (17 users and 3 access points,
 Table 13 and Table 14, Fig. 26) and
 (ii) additional regions
 (11 users and 3 access points,
 Table 13 and Table 14, Fig. 27).
 Two extension design strategies
 are considered:
 {\it strategy I} (separated design for initial region and for additional region) and
 {\it strategy II} (integrated design).

 For the initial region,
 the following parameters are used:
 set of users \( \Psi = \{1,...,i,...,n  \} \) (\(n=17\)),
 set of access points  \( \Theta = \{1,...,i,...,m  \} \)
 (\(m=3\)).
 Each user is described by  parameter vector
 \((x_{i},y_{i},z_{i},f_{i},p_{i} )\),
 where vector components are as follows
 (Table 4):
 coordinates of user \((x_{i},y_{i},z_{i}\),
 required frequency bandwidth \(f_{i}\)
 (scale: 1 Mbit/s ... 10 Mbit/s),
 priority \(p_{i}\)
 (ordinal scale [1,2,3], all user requirements are satisfied in case
 \(p_{i} = 1\)),
 required reliability \(r_{i}\)
  (ordinal scale [1,10],
 \(10\) corresponds to maximum reliability).
 Analogically parameters of access points are considered
 (by index \(j\), Table 5) including parameter \(n_{j}\)
 (maximal possible number of users under service).
\begin{center}
\begin{picture}(48,71)

\put(21.5,67){\makebox(0,0)[bl]{Table 13. Data for initial
region}}

\put(15,62.5){\makebox(0,0)[bl]{(a) users}}

\put(00,00){\line(1,0){48}} \put(00,54){\line(1,0){48}}
\put(00,61){\line(1,0){48}}

\put(00,00){\line(0,1){61}} \put(06,00){\line(0,1){61}}
\put(13,00){\line(0,1){61}} \put(20,00){\line(0,1){61}}
\put(27,00){\line(0,1){61}} \put(34,00){\line(0,1){61}}
\put(41,00){\line(0,1){61}} \put(48,00){\line(0,1){61}}


\put(02,56){\makebox(0,0)[bl]{\(i\)}}

\put(08,56){\makebox(0,0)[bl]{\(x_{i}\)}}
\put(15,56){\makebox(0,0)[bl]{\(y_{i}\)}}
\put(22,56){\makebox(0,0)[bl]{\(z_{i}\)}}

\put(29,56){\makebox(0,0)[bl]{\(f_{i}\)}}

\put(36,56){\makebox(0,0)[bl]{\(p_{i}\)}}
\put(43,56){\makebox(0,0)[bl]{\(r_{i}\)}}

\put(02,50){\makebox(0,0)[bl]{\(1\)}}
\put(08,50){\makebox(0,0)[bl]{\(30\)}}
\put(14,50){\makebox(0,0)[bl]{\(165\)}}
\put(22.5,50){\makebox(0,0)[bl]{\(5\)}}

\put(28.5,50){\makebox(0,0)[bl]{\(10\)}}


\put(36.5,50){\makebox(0,0)[bl]{\(2\)}}
\put(43.5,50){\makebox(0,0)[bl]{\(5\)}}

\put(02,46){\makebox(0,0)[bl]{\(2\)}}
\put(08,46){\makebox(0,0)[bl]{\(58\)}}
\put(14,46){\makebox(0,0)[bl]{\(174\)}}
\put(22.5,46){\makebox(0,0)[bl]{\(5\)}}

\put(29.5,46){\makebox(0,0)[bl]{\(5\)}}


\put(36.5,46){\makebox(0,0)[bl]{\(1\)}}
\put(43.5,46){\makebox(0,0)[bl]{\(9\)}}

\put(02,42){\makebox(0,0)[bl]{\(3\)}}
\put(08,42){\makebox(0,0)[bl]{\(95\)}}
\put(14,42){\makebox(0,0)[bl]{\(156\)}}
\put(22.5,42){\makebox(0,0)[bl]{\(0\)}}
\put(29.5,42){\makebox(0,0)[bl]{\(6\)}}


\put(36.5,42){\makebox(0,0)[bl]{\(1\)}}
\put(43.5,42){\makebox(0,0)[bl]{\(6\)}}

\put(02,38){\makebox(0,0)[bl]{\(4\)}}
\put(08,38){\makebox(0,0)[bl]{\(52\)}}
\put(14,38){\makebox(0,0)[bl]{\(134\)}}
\put(22.5,38){\makebox(0,0)[bl]{\(5\)}}
\put(29.5,38){\makebox(0,0)[bl]{\(6\)}}


\put(36.5,38){\makebox(0,0)[bl]{\(1\)}}
\put(43.5,38){\makebox(0,0)[bl]{\(8\)}}

\put(02,34){\makebox(0,0)[bl]{\(5\)}}
\put(08,34){\makebox(0,0)[bl]{\(85\)}}
\put(14,34){\makebox(0,0)[bl]{\(134\)}}
\put(22.5,34){\makebox(0,0)[bl]{\(3\)}}
\put(29.5,34){\makebox(0,0)[bl]{\(6\)}}


\put(36.5,34){\makebox(0,0)[bl]{\(1\)}}
\put(43.5,34){\makebox(0,0)[bl]{\(7\)}}

\put(02,30){\makebox(0,0)[bl]{\(6\)}}
\put(08,30){\makebox(0,0)[bl]{\(27\)}}
\put(14,30){\makebox(0,0)[bl]{\(109\)}}
\put(22.5,30){\makebox(0,0)[bl]{\(7\)}}
\put(29.5,30){\makebox(0,0)[bl]{\(8\)}}


\put(36.5,30){\makebox(0,0)[bl]{\(3\)}}
\put(43.5,30){\makebox(0,0)[bl]{\(5\)}}

\put(02,26){\makebox(0,0)[bl]{\(7\)}}
\put(08,26){\makebox(0,0)[bl]{\(55\)}}
\put(14,26){\makebox(0,0)[bl]{\(105\)}}
\put(22.5,26){\makebox(0,0)[bl]{\(2\)}}
\put(29.5,26){\makebox(0,0)[bl]{\(7\)}}


\put(36.5,26){\makebox(0,0)[bl]{\(2\)}}
\put(42.5,26){\makebox(0,0)[bl]{\(10\)}}

\put(02,22){\makebox(0,0)[bl]{\(8\)}}
\put(08,22){\makebox(0,0)[bl]{\(98\)}}
\put(15,22){\makebox(0,0)[bl]{\(89\)}}
\put(22.5,22){\makebox(0,0)[bl]{\(3\)}}
\put(28.5,22){\makebox(0,0)[bl]{\(10\)}}


\put(36.5,22){\makebox(0,0)[bl]{\(1\)}}
\put(42.5,22){\makebox(0,0)[bl]{\(10\)}}

\put(02,18){\makebox(0,0)[bl]{\(9\)}}
\put(08,18){\makebox(0,0)[bl]{\(25\)}}
\put(15,18){\makebox(0,0)[bl]{\(65\)}}
\put(22.5,18){\makebox(0,0)[bl]{\(2\)}}
\put(29.5,18){\makebox(0,0)[bl]{\(7\)}}


\put(36.5,18){\makebox(0,0)[bl]{\(3\)}}
\put(43.5,18){\makebox(0,0)[bl]{\(5\)}}

\put(01,14){\makebox(0,0)[bl]{\(10\)}}
\put(08,14){\makebox(0,0)[bl]{\(52\)}}
\put(15,14){\makebox(0,0)[bl]{\(81\)}}
\put(22.5,14){\makebox(0,0)[bl]{\(1\)}}
\put(28.5,14){\makebox(0,0)[bl]{\(10\)}}


\put(36.5,14){\makebox(0,0)[bl]{\(1\)}}
\put(43.5,14){\makebox(0,0)[bl]{\(8\)}}

\put(01,10){\makebox(0,0)[bl]{\(11\)}}
\put(08,10){\makebox(0,0)[bl]{\(65\)}}
\put(15,10){\makebox(0,0)[bl]{\(25\)}}
\put(22.5,10){\makebox(0,0)[bl]{\(7\)}}
\put(29.5,10){\makebox(0,0)[bl]{\(6\)}}


\put(36.5,10){\makebox(0,0)[bl]{\(2\)}}
\put(43.5,10){\makebox(0,0)[bl]{\(9\)}}

\put(01,06){\makebox(0,0)[bl]{\(12\)}}
\put(08,06){\makebox(0,0)[bl]{\(93\)}}
\put(15,06){\makebox(0,0)[bl]{\(39\)}}
\put(22.5,06){\makebox(0,0)[bl]{\(1\)}}
\put(28.5,06){\makebox(0,0)[bl]{\(10\)}}


\put(36.5,06){\makebox(0,0)[bl]{\(1\)}}
\put(42.5,06){\makebox(0,0)[bl]{\(10\)}}

\put(01,02){\makebox(0,0)[bl]{\(13\)}}
\put(07,02){\makebox(0,0)[bl]{\(172\)}}
\put(15,02){\makebox(0,0)[bl]{\(26\)}}
\put(22.5,02){\makebox(0,0)[bl]{\(2\)}}
\put(28.5,02){\makebox(0,0)[bl]{\(10\)}}


\put(36.5,02){\makebox(0,0)[bl]{\(2\)}}
\put(43.5,02){\makebox(0,0)[bl]{\(7\)}}


\end{picture}
%
\begin{picture}(42,71)

\put(08,62.5){\makebox(0,0)[bl]{(b) access points
}}

\put(00,40){\line(1,0){42}} \put(00,54){\line(1,0){42}}
\put(00,61){\line(1,0){42}}

\put(00,40){\line(0,1){21}} \put(04,40){\line(0,1){21}}
\put(11,40){\line(0,1){21}} \put(18,40){\line(0,1){21}}
\put(24,40){\line(0,1){21}} \put(30,40){\line(0,1){21}}
\put(36,40){\line(0,1){21}} \put(42,40){\line(0,1){21}}

\put(01,56){\makebox(0,0)[bl]{\(j\)}}
\put(05.5,56){\makebox(0,0)[bl]{\(x_{j}\)}}
\put(12.5,56){\makebox(0,0)[bl]{\(y_{j}\)}}
\put(19,56){\makebox(0,0)[bl]{\(z_{j}\)}}
\put(25,56){\makebox(0,0)[bl]{\(f_{j}\)}}
\put(31,56){\makebox(0,0)[bl]{\(n_{j}\)}}
\put(37,56){\makebox(0,0)[bl]{\(r_{j}\)}}


\put(01,50){\makebox(0,0)[bl]{\(1\)}}
\put(05.5,50){\makebox(0,0)[bl]{\(50\)}}
\put(11.5,50){\makebox(0,0)[bl]{\(157\)}}
\put(19,50){\makebox(0,0)[bl]{\(10\)}}
\put(25,50){\makebox(0,0)[bl]{\(30\)}}
\put(32,50){\makebox(0,0)[bl]{\(4\)}}
\put(37,50){\makebox(0,0)[bl]{\(10\)}}


\put(01,46){\makebox(0,0)[bl]{\(2\)}}
\put(05.5,46){\makebox(0,0)[bl]{\(72\)}}
\put(11.5,46){\makebox(0,0)[bl]{\(102\)}}
\put(19,46){\makebox(0,0)[bl]{\(10\)}}
\put(25,46){\makebox(0,0)[bl]{\(42\)}}
\put(32,46){\makebox(0,0)[bl]{\(6\)}}
\put(37,46){\makebox(0,0)[bl]{\(10\)}}


\put(01,42){\makebox(0,0)[bl]{\(3\)}}
\put(05.5,42){\makebox(0,0)[bl]{\(45\)}}
\put(12.5,42){\makebox(0,0)[bl]{\(52\)}}
\put(19,42){\makebox(0,0)[bl]{\(10\)}}
\put(25,42){\makebox(0,0)[bl]{\(45\)}}
\put(31,42){\makebox(0,0)[bl]{\(10\)}}
\put(37,42){\makebox(0,0)[bl]{\(10\)}}


\end{picture}
\end{center}

 Thus,
 each pair ``user-access point'' (i.e.,
 \((i,j),  i \in \Psi, j \in \Theta \)) can be described:
 (1) reliability ~\(r_{ij} = min  \{r_{i},r_{j}\}\),
 (2) distance ~\(l_{ij}\),
 (3) priority ~\(p_{ij} = p_{i}\), and
 (4) required bandwidth  ~\(f_{ij} = f_{i}\).
 In addition,
 a ``connectivity'' parameter is considered:~
   \(\beta_{ij}\) equals \(1\) if \(l_{ij} \leq l\) and \(0\)
   otherwise (\(L\) corresponds to distance constraint).
 This parameter defines
 \( \forall i \in \Psi \)
 a subset of possible access points
 \(\Theta_{i} \subseteq \Theta\).
%
%
 The assignment of user \( i\) to access point \( j\)
 is defined by Boolean variable
 \(x_{ij}\) (\(x_{ij} =1\) in the case of assignment
  \( i\) to  \( j\)  and   \(x_{ij} =0\) otherwise).
 Thus,
 the assignment solution
  ~(\( \Psi \) \(\Rightarrow \)  ~\( \Theta \))~
 is defined by Boolean matrix
 ~\(X = || x_{ij} ||, ~ i=\overline{1,n}, ~j=\overline{1,m}\).
%
%
 Finally, the problem formulation is the following:
 \[\max \sum_{i=1}^{n}   \sum_{j \in \Theta_{i}}  r_{ij} x_{ij},
%
%
 ~~\max \sum_{i=1}^{n}   \sum_{j \in \Theta_{i}}  f_{ij}
 x_{ij},
%
 ~~\max \sum_{i=1}^{n}   \sum_{j \in \Theta_{i}}  p_{ij}
 x_{ij} \]
%
%
%
%
 \[s.t. ~~~ \sum_{i=1}^{n}  f_{ij} x_{ij}  \leq  f_{j}
 ~ \forall j \in \Theta,
%
   ~~ \sum_{i=1}^{n}  x_{ij}  \leq  n_{j}
 ~ \forall j \in \Theta,
%
 ~~ \sum_{j \in \Theta_{i}} x_{ij}\leq 1 ~\forall i \in \Psi ,\]
%
%
  \[ x_{ij} \in \{0,1\}, ~\forall ~ i=\overline{1,n},  ~\forall
  ~j=\overline{1,m},
%
   ~~ x_{ij} = 0, ~~\forall ~i=\overline{1,n},   ~j \in  \{\Theta \setminus \Theta_{i}
  \}. \]
%
 Here, a simplified two-stage heuristic used:
 (i) transformation of vector estimate for each pair \((i,j)\)
 into an ordinal estimate (by multicriteria ranking, ELECTRE-like
 technique),
 (ii) solving the obtained one-criterion assignment problem
 (by greedy algorithm).
 Thus, Fig. 26 depicts  the obtained solution: an assignment of users to access
 points.

 Fig. 27 depicts two regions: initial region, additional region,
 and corresponding assignment of users to access points.
 Here the assignment problems are solved separately
 for the initial region (assignment from Fig. 26) and for
 the additional region (i.e., {\it strategy I}).

\begin{center}
\begin{picture}(40,78)
\put(01,00){\makebox(0,0)[bl]{Fig. 26. Initial region
  \cite{levsib10}}}

\put(02,73.5){\makebox(0,0)[bl]{Access}}
\put(02,69.5){\makebox(0,0)[bl]{point}}
\put(08,69.4){\vector(1,-2){2.5}}

\put(22.5,73.5){\makebox(0,0)[bl]{Users}}
\put(22,74){\vector(-1,0){4}} \put(25,72){\vector(1,-2){2}}


\put(14,32){\oval(3.2,4)} \put(14,32){\circle*{2}}
\put(12,27){\makebox(0,0)[bl]{\(10\)}}

\put(15,42){\oval(3.2,4)} \put(15,42){\circle*{2}}
\put(13,37){\makebox(0,0)[bl]{\(7\)}}

\put(19,39){\line(1,0){6}} \put(19,39){\line(1,2){3}}
\put(25,39){\line(-1,2){3}} \put(22,45){\line(0,1){3}}
\put(22,48){\circle*{1}} \put(21,39.5){\makebox(0,0)[bl]{\(2\)}}

\put(22,48){\circle{2.5}} \put(22,48){\circle{3.5}}
\put(24.5,40){\line(2,-1){6}}
\put(23,42.3){\line(1,2){3}} \put(26,48.3){\line(0,1){4.8}}
\put(20.5,42){\line(-1,0){4.5}}

\put(05,43){\oval(3.2,4)} \put(05,43){\circle*{2}}
\put(03,38){\makebox(0,0)[bl]{\(6\)}}

\put(11,60){\line(-1,-3){6}}

\put(05,67){\oval(3.2,4)} \put(05,67){\circle*{2}}
\put(03.8,62){\makebox(0,0)[bl]{\(1\)}}

\put(15,74){\oval(3.2,4)} \put(15,74){\circle*{2}}
\put(13.8,69){\makebox(0,0)[bl]{\(2\)}}

\put(09,60){\line(1,0){6}} \put(09,60){\line(1,2){3}}
\put(15,60){\line(-1,2){3}} \put(12,66){\line(0,1){3}}
\put(12,69){\circle*{1}} \put(11,60.5){\makebox(0,0)[bl]{\(1\)}}

\put(12,69){\circle{2.5}} \put(12,69){\circle{3.5}}
\put(10,62){\line(-1,1){5}}
\put(13.5,60){\line(0,-1){6}}
\put(14,62.3){\line(1,2){2}} \put(16,66.3){\line(0,1){5.8}}


\put(14,53){\oval(3.2,4)} \put(14,53){\circle*{2}}
\put(12.8,48){\makebox(0,0)[bl]{\(4\)}}

\put(04,27){\oval(3.2,4)} \put(04,27){\circle*{2}}
\put(02,22){\makebox(0,0)[bl]{\(9\)}}

\put(18,10){\oval(3.2,4)} \put(18,10){\circle*{2}}
\put(12,09){\makebox(0,0)[bl]{\(11\)}}

\put(28,14){\oval(3.2,4)} \put(28,14){\circle*{2}}
\put(26,9){\makebox(0,0)[bl]{\(12\)}}

\put(30,38){\oval(3.2,4)} \put(30,38){\circle*{2}}
\put(28,33){\makebox(0,0)[bl]{\(8\)}}

\put(25,55){\oval(3.2,4)} \put(25,55){\circle*{2}}
\put(24,50){\makebox(0,0)[bl]{\(5\)}}

\put(29,66){\oval(3.2,4)} \put(29,66){\circle*{2}}
\put(28,61){\makebox(0,0)[bl]{\(3\)}}
\put(28.5,66){\line(-3,-1){14}}

\put(09,16){\line(1,0){6}} \put(09,16){\line(1,2){3}}
\put(15,16){\line(-1,2){3}} \put(12,22){\line(0,1){3}}
\put(12,25){\circle*{1}} \put(11,16.5){\makebox(0,0)[bl]{\(3\)}}

\put(12,25){\circle{2.5}} \put(12,25){\circle{3.5}}
\put(10,18){\line(-2,3){6.5}}
\put(13.5,18.5){\line(1,2){05}} \put(18.5,28.5){\line(-1,1){04}}
\put(14,18){\line(4,-1){14}}
\put(14,16){\line(1,-2){3}}


\put(36,19){\oval(3.2,4)} \put(36,19){\circle*{2}}
\put(34,13){\makebox(0,0)[bl]{\(13\)}}
\put(34.4,18.6){\line(-1,0){20.6}}

\put(39,10){\line(0,1){6}} \put(39.1,10){\line(0,1){6}}
\put(39,20){\line(0,1){6}} \put(39.1,20){\line(0,1){6}}
\put(39,30){\line(0,1){6}} \put(39.1,30){\line(0,1){6}}
\put(39,40){\line(0,1){6}} \put(39.1,40){\line(0,1){6}}
\put(32,48){\line(0,1){6}} \put(32.1,48){\line(0,1){6}}
\put(32,58){\line(0,1){6}} \put(32.1,58){\line(0,1){6}}
\put(32,68){\line(0,1){6}} \put(32.1,68){\line(0,1){6}}

\end{picture}
\end{center}

\begin{center}
\begin{picture}(48,67)

\put(18.5,63){\makebox(0,0)[bl]{Table 14. Data for additional
 region}}

\put(15,58.5){\makebox(0,0)[bl]{(a) users}}

\put(00,00){\line(1,0){48}} \put(00,50){\line(1,0){48}}
\put(00,57){\line(1,0){48}}

\put(00,00){\line(0,1){57}} \put(06,00){\line(0,1){57}}
\put(13,00){\line(0,1){57}} \put(20,00){\line(0,1){57}}
\put(27,00){\line(0,1){57}} \put(34,00){\line(0,1){57}}
\put(41,00){\line(0,1){57}} \put(48,00){\line(0,1){57}}

\put(02,52){\makebox(0,0)[bl]{\(i\)}}

\put(08,52){\makebox(0,0)[bl]{\(x_{i}\)}}
\put(15,52){\makebox(0,0)[bl]{\(y_{i}\)}}
\put(22,52){\makebox(0,0)[bl]{\(z_{i}\)}}

\put(29,52){\makebox(0,0)[bl]{\(f_{i}\)}}

\put(36,52){\makebox(0,0)[bl]{\(p_{i}\)}}
\put(43,52){\makebox(0,0)[bl]{\(r_{i}\)}}

\put(01,46){\makebox(0,0)[bl]{\(14\)}}
\put(07,46){\makebox(0,0)[bl]{\(110\)}}
\put(14,46){\makebox(0,0)[bl]{\(169\)}}
\put(22.5,46){\makebox(0,0)[bl]{\(5\)}}
\put(29.5,46){\makebox(0,0)[bl]{\(7\)}}


\put(36.5,46){\makebox(0,0)[bl]{\(2\)}}
\put(43.5,46){\makebox(0,0)[bl]{\(5\)}}

\put(01,42){\makebox(0,0)[bl]{\(15\)}}
\put(07,42){\makebox(0,0)[bl]{\(145\)}}
\put(14,42){\makebox(0,0)[bl]{\(181\)}}
\put(22.5,42){\makebox(0,0)[bl]{\(3\)}}
\put(29.5,42){\makebox(0,0)[bl]{\(5\)}}


\put(36.5,42){\makebox(0,0)[bl]{\(2\)}}
\put(43.5,42){\makebox(0,0)[bl]{\(4\)}}

\put(01,38){\makebox(0,0)[bl]{\(16\)}}
\put(07,38){\makebox(0,0)[bl]{\(170\)}}
\put(14,38){\makebox(0,0)[bl]{\(161\)}}
\put(22.5,38){\makebox(0,0)[bl]{\(5\)}}
\put(29.5,38){\makebox(0,0)[bl]{\(7\)}}


\put(36.5,38){\makebox(0,0)[bl]{\(2\)}}
\put(43.5,38){\makebox(0,0)[bl]{\(4\)}}

\put(01,34){\makebox(0,0)[bl]{\(17\)}}
\put(07,34){\makebox(0,0)[bl]{\(120\)}}
\put(14,34){\makebox(0,0)[bl]{\(140\)}}
\put(22.5,34){\makebox(0,0)[bl]{\(6\)}}
\put(29.5,34){\makebox(0,0)[bl]{\(4\)}}


\put(36.5,34){\makebox(0,0)[bl]{\(2\)}}
\put(43.5,34){\makebox(0,0)[bl]{\(6\)}}

\put(01,30){\makebox(0,0)[bl]{\(18\)}}
\put(07,30){\makebox(0,0)[bl]{\(150\)}}
\put(14,30){\makebox(0,0)[bl]{\(136\)}}
\put(22.5,30){\makebox(0,0)[bl]{\(3\)}}
\put(29.5,30){\makebox(0,0)[bl]{\(6\)}}


\put(36.5,30){\makebox(0,0)[bl]{\(2\)}}
\put(43.5,30){\makebox(0,0)[bl]{\(7\)}}

\put(01,26){\makebox(0,0)[bl]{\(19\)}}
\put(07,26){\makebox(0,0)[bl]{\(175\)}}
\put(14,26){\makebox(0,0)[bl]{\(125\)}}
\put(22.5,26){\makebox(0,0)[bl]{\(1\)}}
\put(29.5,26){\makebox(0,0)[bl]{\(8\)}}


\put(36.5,26){\makebox(0,0)[bl]{\(3\)}}
\put(43.5,26){\makebox(0,0)[bl]{\(5\)}}

\put(01,22){\makebox(0,0)[bl]{\(20\)}}
\put(07,22){\makebox(0,0)[bl]{\(183\)}}
\put(15,22){\makebox(0,0)[bl]{\(91\)}}
\put(22.5,22){\makebox(0,0)[bl]{\(4\)}}
\put(29.5,22){\makebox(0,0)[bl]{\(4\)}}


\put(36.5,22){\makebox(0,0)[bl]{\(3\)}}
\put(43.5,22){\makebox(0,0)[bl]{\(5\)}}

\put(01,18){\makebox(0,0)[bl]{\(21\)}}
\put(07,18){\makebox(0,0)[bl]{\(135\)}}
\put(15,18){\makebox(0,0)[bl]{\(59\)}}
\put(22.5,18){\makebox(0,0)[bl]{\(4\)}}
\put(28.5,18){\makebox(0,0)[bl]{\(13\)}}


\put(36.5,18){\makebox(0,0)[bl]{\(3\)}}
\put(43.5,18){\makebox(0,0)[bl]{\(4\)}}

\put(01,14){\makebox(0,0)[bl]{\(22\)}}
\put(07,14){\makebox(0,0)[bl]{\(147\)}}
\put(15,14){\makebox(0,0)[bl]{\(79\)}}
\put(22.5,14){\makebox(0,0)[bl]{\(5\)}}
\put(29.5,14){\makebox(0,0)[bl]{\(7\)}}


\put(36.5,14){\makebox(0,0)[bl]{\(3\)}}
\put(42.5,14){\makebox(0,0)[bl]{\(16\)}}

\put(01,10){\makebox(0,0)[bl]{\(23\)}}
\put(07,10){\makebox(0,0)[bl]{\(172\)}}
\put(15,10){\makebox(0,0)[bl]{\(26\)}}
\put(22.5,10){\makebox(0,0)[bl]{\(2\)}}
\put(28.5,10){\makebox(0,0)[bl]{\(10\)}}


\put(36.5,10){\makebox(0,0)[bl]{\(2\)}}
\put(43.5,10){\makebox(0,0)[bl]{\(7\)}}

\put(01,06){\makebox(0,0)[bl]{\(24\)}}
\put(07,06){\makebox(0,0)[bl]{\(165\)}}
\put(15,06){\makebox(0,0)[bl]{\(50\)}}
\put(22.5,06){\makebox(0,0)[bl]{\(3\)}}
\put(28.5,06){\makebox(0,0)[bl]{\(7\)}}


\put(36.5,06){\makebox(0,0)[bl]{\(3\)}}
\put(43.5,06){\makebox(0,0)[bl]{\(3\)}}

\put(01,02){\makebox(0,0)[bl]{\(25\)}}
\put(07,02){\makebox(0,0)[bl]{\(127\)}}
\put(15,02){\makebox(0,0)[bl]{\(95\)}}
\put(22.5,02){\makebox(0,0)[bl]{\(5\)}}
\put(28.5,02){\makebox(0,0)[bl]{\(7\)}}


\put(36.5,02){\makebox(0,0)[bl]{\(2\)}}
\put(43.5,02){\makebox(0,0)[bl]{\(5\)}}


\end{picture}
%
\begin{picture}(42,67)

\put(08,58.5){\makebox(0,0)[bl]{(b) access points}}

\put(00,36){\line(1,0){42}} \put(00,50){\line(1,0){42}}
\put(00,57){\line(1,0){42}}

\put(00,36){\line(0,1){21}} \put(04,36){\line(0,1){21}}
\put(11,36){\line(0,1){21}} \put(18,36){\line(0,1){21}}
\put(24,36){\line(0,1){21}} \put(30,36){\line(0,1){21}}
\put(36,36){\line(0,1){21}} \put(42,36){\line(0,1){21}}

\put(01,52){\makebox(0,0)[bl]{\(j\)}}
\put(05.5,52){\makebox(0,0)[bl]{\(x_{j}\)}}
\put(12.5,52){\makebox(0,0)[bl]{\(y_{j}\)}}
\put(19,52){\makebox(0,0)[bl]{\(z_{j}\)}}
\put(25,52){\makebox(0,0)[bl]{\(f_{j}\)}}
\put(31,52){\makebox(0,0)[bl]{\(n_{j}\)}}
\put(37,52){\makebox(0,0)[bl]{\(r_{j}\)}}


\put(01,46){\makebox(0,0)[bl]{\(4\)}}
\put(04.5,46){\makebox(0,0)[bl]{\(150\)}}
\put(11.5,46){\makebox(0,0)[bl]{\(165\)}}
\put(19,46){\makebox(0,0)[bl]{\(10\)}}
\put(25,46){\makebox(0,0)[bl]{\(30\)}}
\put(32,46){\makebox(0,0)[bl]{\(5\)}}
\put(37,46){\makebox(0,0)[bl]{\(15\)}}


\put(01,42){\makebox(0,0)[bl]{\(5\)}}
\put(04.5,42){\makebox(0,0)[bl]{\(140\)}}
\put(11.5,42){\makebox(0,0)[bl]{\(112\)}}
\put(19,42){\makebox(0,0)[bl]{\(10\)}}
\put(25,42){\makebox(0,0)[bl]{\(32\)}}
\put(32,42){\makebox(0,0)[bl]{\(5\)}}
\put(38,42){\makebox(0,0)[bl]{\(8\)}}


\put(01,38){\makebox(0,0)[bl]{\(6\)}}
\put(04.5,38){\makebox(0,0)[bl]{\(147\)}}
\put(12.5,38){\makebox(0,0)[bl]{\(47\)}}
\put(19,38){\makebox(0,0)[bl]{\(10\)}}
\put(25,38){\makebox(0,0)[bl]{\(30\)}}
\put(32,38){\makebox(0,0)[bl]{\(5\)}}
\put(37,38){\makebox(0,0)[bl]{\(15\)}}

\end{picture}
\end{center}

\begin{center}
\begin{picture}(68,80)

\put(00,00){\makebox(0,0)[bl]{Fig. 27.
 Two regions: separated assignment \cite{levsib10}}}


\put(14,32){\oval(3.2,4)} \put(14,32){\circle*{2}}
\put(12,27){\makebox(0,0)[bl]{\(10\)}}

\put(15,42){\oval(3.2,4)} \put(15,42){\circle*{2}}
\put(13,37){\makebox(0,0)[bl]{\(7\)}}

\put(19,39){\line(1,0){6}} \put(19,39){\line(1,2){3}}
\put(25,39){\line(-1,2){3}} \put(22,45){\line(0,1){3}}
\put(22,48){\circle*{1}}

\put(21,39.5){\makebox(0,0)[bl]{\(2\)}}

\put(22,48){\circle{2.5}} \put(22,48){\circle{3.5}}
\put(24.5,40){\line(2,-1){6}}
\put(23,42.3){\line(1,2){3}} \put(26,48.3){\line(0,1){4.8}}
\put(20.5,42){\line(-1,0){4.5}}

\put(05,43){\oval(3.2,4)} \put(05,43){\circle*{2}}
\put(03,38){\makebox(0,0)[bl]{\(6\)}}

\put(11,60){\line(-1,-3){6}}

\put(05,67){\oval(3.2,4)} \put(05,67){\circle*{2}}
\put(03.8,62){\makebox(0,0)[bl]{\(1\)}}

\put(15,74){\oval(3.2,4)} \put(15,74){\circle*{2}}
\put(17.3,73){\makebox(0,0)[bl]{\(2\)}}

\put(09,60){\line(1,0){6}} \put(09,60){\line(1,2){3}}
\put(15,60){\line(-1,2){3}} \put(12,66){\line(0,1){3}}
\put(12,69){\circle*{1}} \put(11,60.5){\makebox(0,0)[bl]{\(1\)}}

\put(12,69){\circle{2.5}} \put(12,69){\circle{3.5}}
\put(10,62){\line(-1,1){5}}
\put(13.5,60){\line(0,-1){6}}
\put(14,62.3){\line(1,2){2}} \put(16,66.3){\line(0,1){5.8}}

\put(14,53){\oval(3.2,4)} \put(14,53){\circle*{2}}
\put(12.8,48){\makebox(0,0)[bl]{\(4\)}}

\put(04,27){\oval(3.2,4)} \put(04,27){\circle*{2}}
\put(02,22){\makebox(0,0)[bl]{\(9\)}}


\put(18,10){\oval(3.2,4)} \put(18,10){\circle*{2}}

\put(12,09){\makebox(0,0)[bl]{\(11\)}}

\put(28,14){\oval(3.2,4)} \put(28,14){\circle*{2}}
\put(26,9){\makebox(0,0)[bl]{\(12\)}}

\put(30,38){\oval(3.2,4)} \put(30,38){\circle*{2}}
\put(28,33){\makebox(0,0)[bl]{\(8\)}}

\put(38,60){\oval(3.2,4)} \put(38,60){\circle*{2}}
\put(36,55){\makebox(0,0)[bl]{\(17\)}}

\put(50,58){\oval(3.2,4)} \put(50,58){\circle*{2}}
\put(49,53){\makebox(0,0)[bl]{\(18\)}}

\put(42,46){\line(1,0){6}} \put(42,46){\line(1,2){3}}
\put(48,46){\line(-1,2){3}} \put(45,52){\line(0,1){3}}
\put(45,55){\circle*{1}} \put(44,46.5){\makebox(0,0)[bl]{\(5\)}}

\put(45,55){\circle{2.5}} \put(45,55){\circle{3.5}}

\put(43,48){\line(-1,3){4}}

\put(46,49.4){\line(1,2){4.5}}
\put(47,48){\line(3,1){13}}

\put(60,52){\oval(3.2,4)} \put(60,52){\circle*{2}}
\put(58,47){\makebox(0,0)[bl]{\(19\)}}

\put(63,35){\oval(3.2,4)} \put(63,35){\circle*{2}}
\put(61,30){\makebox(0,0)[bl]{\(20\)}}

\put(49,30){\oval(3.2,4)} \put(49,30){\circle*{2}}
\put(47,25){\makebox(0,0)[bl]{\(22\)}}

\put(44,20){\oval(3.2,4)} \put(44,20){\circle*{2}}
\put(42,15){\makebox(0,0)[bl]{\(21\)}}

\put(61,10){\oval(3.2,4)} \put(61,10){\circle*{2}}
\put(55,09){\makebox(0,0)[bl]{\(23\)}}

\put(61,20){\oval(3.2,4)} \put(61,20){\circle*{2}}
\put(59,15){\makebox(0,0)[bl]{\(24\)}}

\put(42,37){\oval(3.2,4)} \put(42,37){\circle*{2}}
\put(40,32){\makebox(0,0)[bl]{\(25\)}}

\put(42,37){\line(1,3){3}}

\put(47,13){\line(1,0){6}} \put(47,13){\line(1,2){3}}
\put(53,13){\line(-1,2){3}} \put(50,19){\line(0,1){3}}
\put(50,22){\circle*{1}} \put(49,13.5){\makebox(0,0)[bl]{\(6\)}}

\put(50,22){\circle{2.5}} \put(50,22){\circle{3.5}}
\put(61,20){\line(-2,-1){9}}
\put(48,15.5){\line(-1,1){4.5}}
\put(48.7,17){\line(-1,2){4.2}} \put(44.5,25.4){\line(1,1){4}}
\put(51,16){\line(1,2){07.5}} \put(58.5,31){\line(1,1){04}}
\put(52,15){\line(2,-1){8.6}}

\put(25,55){\oval(3.2,4)} \put(25,55){\circle*{2}}
\put(24,50){\makebox(0,0)[bl]{\(5\)}}
\put(29,66){\oval(3.2,4)} \put(29,66){\circle*{2}}
\put(28,61){\makebox(0,0)[bl]{\(3\)}}
\put(28.5,66){\line(-3,-1){14}}

\put(35,72){\oval(3.2,4)} \put(35,72){\circle*{2}}
\put(34,67){\makebox(0,0)[bl]{\(14\)}}

\put(45,77){\oval(3.2,4)} \put(45,77){\circle*{2}}
\put(43,72){\makebox(0,0)[bl]{\(15\)}}

\put(58,67){\oval(3.2,4)} \put(58,67){\circle*{2}}
\put(57,62){\makebox(0,0)[bl]{\(16\)}}

\put(47,68){\line(1,0){6}} \put(47,68){\line(1,2){3}}
\put(53,68){\line(-1,2){3}} \put(50,74){\line(0,1){3}}
\put(50,77){\circle*{1}} \put(49,68.5){\makebox(0,0)[bl]{\(4\)}}

\put(50,77){\circle{2.5}} \put(50,77){\circle{3.5}}

\put(48,70){\line(-1,0){09}} \put(39,70){\line(-2,1){5}}
\put(48.4,71.2){\line(-1,2){3}}
\put(52,70){\line(2,-1){5}}

\put(09,16){\line(1,0){6}} \put(09,16){\line(1,2){3}}
\put(15,16){\line(-1,2){3}} \put(12,22){\line(0,1){3}}
\put(12,25){\circle*{1}} \put(11,16.5){\makebox(0,0)[bl]{\(3\)}}

\put(12,25){\circle{2.5}} \put(12,25){\circle{3.5}}
\put(10,18){\line(-2,3){6.5}}
\put(13.5,18.5){\line(1,2){05}} \put(18.5,28.5){\line(-1,1){04}}
\put(14,18){\line(4,-1){14}}
\put(14,16){\line(1,-2){3}}


\put(36,19){\oval(3.2,4)} \put(36,19){\circle*{2}}
\put(34,13){\makebox(0,0)[bl]{\(13\)}}
\put(34.4,18.6){\line(-1,0){20.6}}

\put(39,10){\line(0,1){6}} \put(39.1,10){\line(0,1){6}}
\put(39,20){\line(0,1){6}} \put(39.1,20){\line(0,1){6}}
\put(39,30){\line(0,1){6}} \put(39.1,30){\line(0,1){6}}
\put(39,40){\line(0,1){6}} \put(39.1,40){\line(0,1){6}}
\put(32,48){\line(0,1){6}} \put(32.1,48){\line(0,1){6}}
\put(32,58){\line(0,1){6}} \put(32.1,58){\line(0,1){6}}
\put(32,68){\line(0,1){6}} \put(32.1,68){\line(0,1){6}}

\end{picture}
\end{center}

 Fig. 28 depicts the results
 of integrated (joint) design {\it strategy II}).
 Note,
 the following users are re-assigned:
 \(3\), \(13\), and \(25\).
 Generally, it may be reasonable
 (to decrease the dimensions of the problems under the solving process)
 to consider the following approach:
 (a) design (assignment) for the initial region,
 (b) design (assignment) for the additional region,
 (c) re-design (re-assignment) for users
 which belong to a border subregion,
 e.g., in the example (Fig. 27, Fig. 28)
 the user set involves the following users:~
 \( \{ 3, 5, 8, 13, 14, 17, 21, 25  \}\).

\begin{center}
\begin{picture}(68,80)
\put(01,00){\makebox(0,0)[bl]{Fig. 28.
 Two regions: joint assignment
  \cite{levsib10}}}


\put(14,32){\oval(3.2,4)} \put(14,32){\circle*{2}}
\put(12,27){\makebox(0,0)[bl]{\(10\)}}

\put(15,42){\oval(3.2,4)} \put(15,42){\circle*{2}}
\put(13,37){\makebox(0,0)[bl]{\(7\)}}

\put(19,39){\line(1,0){6}} \put(19,39){\line(1,2){3}}
\put(25,39){\line(-1,2){3}} \put(22,45){\line(0,1){3}}
\put(22,48){\circle*{1}} \put(21,39.5){\makebox(0,0)[bl]{\(2\)}}

\put(22,48){\circle{2.5}} \put(22,48){\circle{3.5}}
\put(24.5,40){\line(2,-1){6}}
\put(23,42.3){\line(1,2){3}} \put(26,48.3){\line(0,1){4.8}}
\put(20.5,42){\line(-1,0){4.5}}

\put(05,43){\oval(3.2,4)} \put(05,43){\circle*{2}}
\put(03,38){\makebox(0,0)[bl]{\(6\)}}

\put(11,60){\line(-1,-3){6}}

\put(05,67){\oval(3.2,4)} \put(05,67){\circle*{2}}
\put(03.8,62){\makebox(0,0)[bl]{\(1\)}}

\put(15,74){\oval(3.2,4)} \put(15,74){\circle*{2}}
\put(17.3,73){\makebox(0,0)[bl]{\(2\)}}

\put(09,60){\line(1,0){6}} \put(09,60){\line(1,2){3}}
\put(15,60){\line(-1,2){3}} \put(12,66){\line(0,1){3}}
\put(12,69){\circle*{1}} \put(11,60.5){\makebox(0,0)[bl]{\(1\)}}

\put(12,69){\circle{2.5}} \put(12,69){\circle{3.5}}
\put(10,62){\line(-1,1){5}}
\put(13.5,60){\line(0,-1){6}}
\put(14,62.3){\line(1,2){2}} \put(16,66.3){\line(0,1){5.8}}

\put(14,53){\oval(3.2,4)} \put(14,53){\circle*{2}}
\put(12.8,48){\makebox(0,0)[bl]{\(4\)}}

\put(04,27){\oval(3.2,4)} \put(04,27){\circle*{2}}
\put(02,22){\makebox(0,0)[bl]{\(9\)}}

\put(18,10){\oval(3.2,4)} \put(18,10){\circle*{2}}
\put(12,09){\makebox(0,0)[bl]{\(11\)}}

\put(28,14){\oval(3.2,4)} \put(28,14){\circle*{2}}
\put(26,9){\makebox(0,0)[bl]{\(12\)}}

\put(30,38){\oval(3.2,4)} \put(30,38){\circle*{2}}
\put(28,33){\makebox(0,0)[bl]{\(8\)}}

\put(38,60){\oval(3.2,4)} \put(38,60){\circle*{2}}
\put(36,55){\makebox(0,0)[bl]{\(17\)}}

\put(50,58){\oval(3.2,4)} \put(50,58){\circle*{2}}
\put(49,53){\makebox(0,0)[bl]{\(18\)}}

\put(42,46){\line(1,0){6}} \put(42,46){\line(1,2){3}}
\put(48,46){\line(-1,2){3}} \put(45,52){\line(0,1){3}}
\put(45,55){\circle*{1}} \put(44,46.5){\makebox(0,0)[bl]{\(5\)}}

\put(45,55){\circle{2.5}} \put(45,55){\circle{3.5}}

\put(43,48){\line(-1,3){4}}

\put(46,49.4){\line(1,2){4.5}}
\put(47,48){\line(3,1){13}}

\put(60,52){\oval(3.2,4)} \put(60,52){\circle*{2}}
\put(58,47){\makebox(0,0)[bl]{\(19\)}}

\put(63,35){\oval(3.2,4)} \put(63,35){\circle*{2}}
\put(61,30){\makebox(0,0)[bl]{\(20\)}}

\put(49,30){\oval(3.2,4)} \put(49,30){\circle*{2}}
\put(47,25){\makebox(0,0)[bl]{\(22\)}}

\put(44,20){\oval(3.2,4)} \put(44,20){\circle*{2}}
\put(42,15){\makebox(0,0)[bl]{\(21\)}}

\put(61,10){\oval(3.2,4)} \put(61,10){\circle*{2}}
\put(55,09){\makebox(0,0)[bl]{\(23\)}}


\put(61,20){\oval(3.2,4)} \put(61,20){\circle*{2}}
\put(59,15){\makebox(0,0)[bl]{\(24\)}}

\put(42,37){\oval(3.2,4)} \put(42,37){\circle*{2}}
\put(40,32){\makebox(0,0)[bl]{\(25\)}}


\put(42,37){\line(-1,0){6}} \put(36,37){\line(-1,1){4}}
\put(32,41){\line(-1,0){8}}

\put(47,13){\line(1,0){6}} \put(47,13){\line(1,2){3}}
\put(53,13){\line(-1,2){3}} \put(50,19){\line(0,1){3}}
\put(50,22){\circle*{1}} \put(49,13.5){\makebox(0,0)[bl]{\(6\)}}

\put(50,22){\circle{2.5}} \put(50,22){\circle{3.5}}
\put(61,20){\line(-2,-1){9}}
\put(48,15.5){\line(-1,1){4.5}}
\put(48.7,17){\line(-1,2){4.2}} \put(44.5,25.4){\line(1,1){4}}
\put(51,16){\line(1,2){07.5}} \put(58.5,31){\line(1,1){04}}
\put(52,15){\line(2,-1){8.6}}

\put(25,55){\oval(3.2,4)} \put(25,55){\circle*{2}}
\put(24,50){\makebox(0,0)[bl]{\(5\)}}
\put(29,66){\oval(3.2,4)} \put(29,66){\circle*{2}}
\put(28,61){\makebox(0,0)[bl]{\(3\)}}
\put(29,66){\line(1,0){15}} \put(44,66){\line(2,1){4}}

\put(35,72){\oval(3.2,4)} \put(35,72){\circle*{2}}
\put(34,67){\makebox(0,0)[bl]{\(14\)}}

\put(45,77){\oval(3.2,4)} \put(45,77){\circle*{2}}
\put(43,72){\makebox(0,0)[bl]{\(15\)}}

\put(58,67){\oval(3.2,4)} \put(58,67){\circle*{2}}
\put(57,62){\makebox(0,0)[bl]{\(16\)}}

\put(47,68){\line(1,0){6}} \put(47,68){\line(1,2){3}}
\put(53,68){\line(-1,2){3}} \put(50,74){\line(0,1){3}}
\put(50,77){\circle*{1}} \put(49,68.5){\makebox(0,0)[bl]{\(4\)}}

\put(50,77){\circle{2.5}} \put(50,77){\circle{3.5}}
\put(48,70){\line(-1,0){09}} \put(39,70){\line(-2,1){5}}
\put(48.4,71.2){\line(-1,2){3}}
\put(52,70){\line(2,-1){5}}

\put(09,16){\line(1,0){6}} \put(09,16){\line(1,2){3}}
\put(15,16){\line(-1,2){3}} \put(12,22){\line(0,1){3}}
\put(12,25){\circle*{1}} \put(11,16.5){\makebox(0,0)[bl]{\(3\)}}

\put(12,25){\circle{2.5}} \put(12,25){\circle{3.5}}
\put(10,18){\line(-2,3){6.5}}
\put(13.5,18.5){\line(1,2){05}} \put(18.5,28.5){\line(-1,1){04}}
\put(14,18){\line(4,-1){14}}
\put(14,16){\line(1,-2){3}}


\put(36,19){\oval(3.2,4)} \put(36,19){\circle*{2}}
\put(34,13){\makebox(0,0)[bl]{\(13\)}}

\put(36,19){\line(1,0){3}} \put(39,19){\line(1,-2){2.5}}
\put(41.5,14){\line(1,0){6}}


\put(39,10){\line(0,1){6}} \put(39.1,10){\line(0,1){6}}
\put(39,20){\line(0,1){6}} \put(39.1,20){\line(0,1){6}}
\put(39,30){\line(0,1){6}} \put(39.1,30){\line(0,1){6}}
\put(39,40){\line(0,1){6}} \put(39.1,40){\line(0,1){6}}
\put(32,48){\line(0,1){6}} \put(32.1,48){\line(0,1){6}}
\put(32,58){\line(0,1){6}} \put(32.1,58){\line(0,1){6}}
\put(32,68){\line(0,1){6}} \put(32.1,68){\line(0,1){6}}

\end{picture}
\end{center}

\section{Conclusion}

 The paper describes combinatorial approaches to system
 improvement/extension of modular systems.
 Basic system improvement/extension actions are described
 (e.g., improvement/replacement of system components, improvement/modification
  of system structure)
 and composite
 improvement/extension strategies.
 The described methods are
 based on combinatorial optimization problems
 (knapsack problem, multiple choice problem, assignment/allocation problem,
 spanning problems, hotlink assignment problem, etc.).
 Realistic numerical examples illustrate the described strategies.

 In the future, it may be  reasonable to consider
 the following research directions:
 (1) examination of various  real-world applications,
 (2) examination of multi-stage system improvement/extension
 strategies,
 (3) special study of improvement/extention strategies for problem
 solving frameworks
 (e.g., \cite{lev12b}),
 (4) taking into account uncertainty,
%
 (5) special study of reoptimization/restructuring approaches,
 and
 (6) usage of the described system approaches in
 education (computer science, engineering, applied mathematics).

\end{document}